\documentclass[pdflatex,sn-mathphys-num]{sn-jnl}

\usepackage{graphicx}%
\usepackage{multirow}%
\usepackage{amsmath,amssymb,amsfonts}%
\usepackage{amsthm}%
\usepackage{mathrsfs}%
\usepackage{mathrsfs}
\DeclareFontFamily{U}{rsfs}{\skewchar\font127}
\DeclareFontShape{U}{rsfs}{m}{n}{%
   <-6> rsfs5
   <6-8> rsfs7
   <8-> rsfs10
}{}
\usepackage[title]{appendix}%
\usepackage{xcolor}%
\usepackage{textcomp}%
\usepackage{manyfoot}%
\usepackage{booktabs}%
\usepackage{algorithm}%
\usepackage{algorithmicx}%
\usepackage{algpseudocode}%
\usepackage{listings}%
\usepackage[utf8]{inputenc}
\usepackage[shortlabels]{enumitem}
\usepackage{pdflscape}
\usepackage{longtable} 
\usepackage{adjustbox}
\usepackage{url}
\usepackage{threeparttable}
\usepackage{float}
\usepackage{array}
\usepackage{placeins}
\usepackage{multirow}

\graphicspath{ {./}}
\newcounter{fignum}
\newcommand{\includecroppedfigure}[1][]{%
    \includegraphics[page=\thefignum,#1]{Figure_cropped.pdf}%
    \stepcounter{fignum}%
}

\theoremstyle{thmstyleone}%
\theoremstyle{thmstyletwo}%

\theoremstyle{thmstylethree}%

\begin{document}

\title{A Decade of Generative Adversarial Networks for Porous Material Reconstruction}

\author[1]{\fnm{Ali} \sur{Sadeghkhani}}

\author[1]{\fnm{Brandon} \sur{Bennett}}

\author[2]{\fnm{Masoud} \sur{Babaei}}

\author*[1]{\fnm{Arash} \sur{Rabbani}}\email{a.rabbani@leeds.ac.uk}

\affil*[1]{\orgdiv{School of Computer Science}, \orgname{University of Leeds}, \orgaddress{\city{Leeds}, \country{UK}}}

\affil[2]{\orgdiv{School of Chemical Engineering}, \orgname{The University of Manchester}, \orgaddress{\city{Manchester}, \country{UK}}}


\abstract{Digital reconstruction of porous materials has become increasingly critical for applications ranging from geological reservoir characterization to tissue engineering and electrochemical device design. While traditional methods such as micro-computed tomography and statistical reconstruction approaches have established foundations in this field, the emergence of deep learning techniques, particularly Generative Adversarial Networks (GANs), has revolutionized porous media reconstruction capabilities. This comprehensive review systematically analyzes 96 peer-reviewed articles published from 2017 to early 2026, examining the evolution and applications of GAN-based approaches for porous material image reconstruction.

We categorize GAN architectures into six distinct classes, namely Vanilla GANs, Multi-Scale GANs, Conditional GANs, Attention-Enhanced GANs, Style-based GANs, and Hybrid Architecture GANs. Each category addresses specific challenges in porous media reconstruction, from capturing multi-scale hierarchical features to enabling precise property control and improving computational efficiency. Our analysis reveals substantial progress in reconstruction capabilities, including improvements in porosity accuracy (within 1\% of original samples), permeability prediction (up to 79\% reduction in mean relative errors), and achievable reconstruction volumes (from initial $64^3$ to current $2{,}200^3$ voxels).

The review demonstrates that while fundamental GANs established baseline capabilities, specialized architectures have introduced critical innovations including progressive growing for multi-scale feature capture, conditional generation for targeted property control, attention mechanisms for long-range dependency preservation, and hybrid approaches for enhanced training stability with limited data. Despite these advances, persistent challenges remain in computational efficiency, memory constraints for large-scale reconstruction, and maintaining structural continuity in 2D-to-3D transformations. This systematic analysis provides researchers and practitioners with a comprehensive framework for selecting appropriate GAN architectures based on specific application requirements and identifies crucial directions for future development in this rapidly evolving field.}

\maketitle

\section{Introduction}

Digital Reconstruction of porous media plays a crucial role in various scientific and engineering applications, including tissue engineering, bone scaffold design, metal foam and rock reconstruction. This technique allows researchers to create digital representations of these materials, enabling detailed analysis of their internal structures. This technology offers several advantages over traditional methods, providing valuable insights into aspects like porosity, permeability, porous media structure, and fluid flow behavior within porous structure.

Traditional methods for 3D porous medium reconstruction include micro-Computed Tomography (micro-CT) and focused ion beam scanning electron microscopy (FIB-SEM) \citep{Hemes2015Multi-scaleTomography, Li2018DirectQuantification, Tahmasebi2015Three-DimensionalImages, Gupta2019Open-sourcePrinter, Krohn1986FractalImages}, optimization-based approaches \citep{Joshi1974AMedia, Adler1990FlowMedia, Quiblier1984AMedia, Roberts1997StatisticalImages, Hazlett1997Statistical1, Yeong1998ReconstructingMedia}, and multi-point statistics (MPS) \citep{Strebelle2002Conditional1, Okabe2004PredictionStatistics, Tahmasebi2012Multiple-pointFunctions}. While these methods have established the foundations of the field, deep learning has emerged as a more computationally efficient alternative, capable of capturing complex structural characteristics of porous media directly from data \citep{Mosser2017ReconstructionNetworks, Cai2020PredictionForest}. Among deep learning architectures, Convolutional Neural Networks (CNNs) were the first to be widely applied to porous material modelling, demonstrating effectiveness in image processing, segmentation, and property prediction tasks \citep{Baduge2022ArtificialApplications, Li2019PredictingLearning, Lingxin2022ADetection, Shin2019DigitalNetwork, Wang2021EstablishmentPenalty}. Building on these advances, Generative Adversarial Networks (GANs) \citep{Goodfellow2014GenerativeNetworks} have gained significant traction in pore-scale imaging tasks including image segmentation, resolution enhancement, and reconstruction. By learning the probability distribution of training data, GANs can generate synthetic rock representations that capture the inherent variability of real porous media structures.

However, GAN training presents inherent challenges including mode collapse, vanishing gradients, and exploding gradients \citep{Cao2019RecentVision}. These difficulties, combined with the growing demand for generating data with specific features, have driven the development of various specialized GAN models, each tailored to address specific limitations and requirements across different application domains.

In recent years, there have been several researches and reviewing articles focusing on imaging techniques and digital representations for porous media analysis. In terms of traditional methods, Berg et al. \citep{Berg2017IndustrialTechnology} provided a valuable resource by outlining imaging and image analysis techniques alongside methods for calculating properties from digital representations of porous media. Bostanabad et al. \citep{Bostanabad2018ComputationalTechniques} and Sahimi et al. \citep{Sahimi2021ReconstructionApplications} built upon this foundation by reviewing microstructure characterization and reconstruction methods, with a focus on statistically-based approaches. Wang et al. \citep{Wang2023NumericalReview} presented a comprehensive review of pore-scale reservoir modeling via numerical methods.

Tahmasebi et al. \citep{Tahmasebi2020MachineScale} investigated the potential of machine learning and deep learning algorithms for applications in porous media and geoscience. Rabbani et al. \citep{Rabbani2021ReviewTechniques} further investigated machine learning for porous material analysis, leveraging dimension reduction and clustering techniques. Wang et al. \citep{Wang2021DeepModeling} conducted a thorough review of deep learning tools used for pore-scale imaging and modeling, even discussing research into deep learning for image generation in porous media. A comprehensive review and analysis of different types of data-driven additive manufacturing (AM) modeling, and the potential of physics-informed machine learning in future data-driven AM modelling was presented by Wang et al. \citep{Wang2022Data-drivenDirections}. Ferreira et al. \citep{Ferreira2022GAN-basedTaxonomy} provided a comprehensive review of the use of GANs for the generation of volumetric data, covering healthcare and medical applications.

More recently, Li et al.\ \citep{Li2023AdvancesTechnology} reviewed deep learning applications in digital rock analysis, highlighting its use in reconstruction, image enhancement, segmentation, and property prediction. This trend continued with Mirzaee et al.\ \citep{Mirzaee2023MinireviewOutlook} reviewing machine and deep learning approaches for porous microstructure reconstruction, comparing existing techniques. Yang et al.\ \citep{Yang2023RecentProduction} demonstrated the potential of multiscale digital rock techniques by integrating imaging and fluid flow modeling for shale gas production. Beyond geological and materials applications, Alajaji et al.\ \citep{Alajaji2024GenerativeDirections} explored the application and challenges of GANs in digital histopathology, showcasing the broader applicability of these techniques across biological imaging domains. Most recently, Elrahmani et al.\ \citep{Elrahmani2026SyntheticReview} provided a domain-specific synthesis of deep generative models for synthetic porous media generation, systematically evaluating VAEs, diffusion models, and multimodal 3D pipelines alongside hybrid frameworks coupling grain-scale synthesis with discrete element method-based packing.

A critical review of the existing literature reveals a gap in knowledge regarding the specific applicability of various Generative Adversarial Network (GAN) models for porous media reconstruction. While numerous studies since 2017 have explored GAN-based reconstruction of porous media, a comprehensive analysis comparing the effectiveness of different GAN architectures appears to be lacking. This identified knowledge gap motivates the present review article, which aims to delve deeper into this specific aspect.

The scope and focus of this review can be visualised through a text-analysis-based word cloud (Figure \ref{fig:wordcloud}) generated from the titles and abstracts of all identified relevant publications. The dominance of terms such as ``reconstruction,'' ``microstructure,'' and ``generative adversarial'' confirms the central position of GAN-based generation within the broader digital rock literature. Notably, the frequent co-occurrence of terms related to both structural characterisation (``pore,'' ``structure,'' ``feature'') and computational methodology (``network,'' ``deep learning,'' ``training'') reflects the inherently interdisciplinary nature of this field, bridging materials science and machine learning.

\begin{figure}[H]
    \centering
    \includecroppedfigure[width=0.9\textwidth]
    \caption{Word cloud visualization generated from titles and abstracts of reviewed publications in GAN-based porous media reconstruction. Font size corresponds to term frequency, revealing key research themes and methodological approaches.}
    \label{fig:wordcloud}
\end{figure}

To ensure a systematic approach in identifying relevant articles for this review, a structured search was performed across several academic databases including Google Scholar, PubMed, IEEE Xplore, Web of Science, and Science Direct. The search was guided by a set of carefully chosen keywords, which included "Generative Adversarial Networks" (GANs), "Porous Media Reconstruction", "3D Reconstruction", and "Digital Rock Modeling". Additional specific terms such as "Super-resolution GAN", "Stochastic Reconstruction", "Conditional GAN", "Wasserstein GAN", and "Multi-scale GAN" were also employed to capture the full breadth of relevant literature.

The initial search returned 185 articles related to these keywords. After applying a more refined screening process focused specifically on "Porous Medium Reconstruction by GAN Models", the list was narrowed down to 96 articles. These articles were selected for their direct relevance to the application of GANs in the reconstruction of porous media, covering a range of techniques, from traditional GAN models to more advanced approaches incorporating multi-scale and conditional GANs, attention mechanisms, and super-resolution methods. This systematic approach ensured a comprehensive collection of studies, which were categorized into the following GAN-based methods:
\begin{enumerate}
    \item Vanilla GAN
    \item Multi-Scale GAN
    \item Conditional GAN
    \item Attention-Enhanced GAN
    \item Style-based GAN
    \item Hybrid Architecture GAN
\end{enumerate}

\begin{figure}[H]
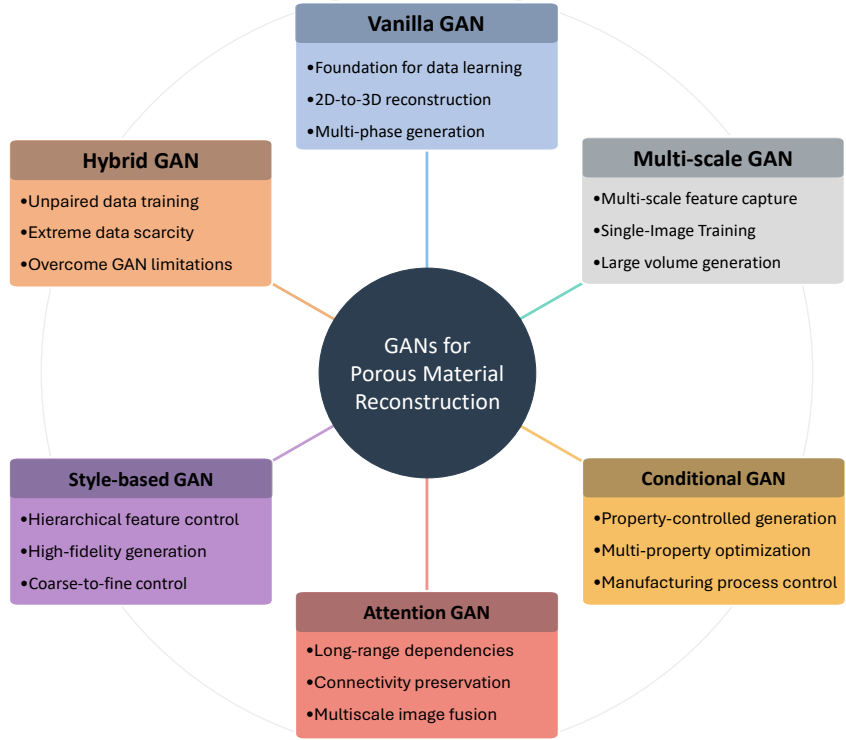

    \centering
    \includecroppedfigure[width=0.9\textwidth]
    \caption{Overview of GAN architectures for porous media reconstruction. The classification encompasses six distinct architectural paradigms: (1) Vanilla GAN establishing foundational capabilities for 3D reconstruction and multi-phase generation, (2) Multi-Scale GAN enabling hierarchical feature capture and large volume generation through progressive or concurrent training, (3) Conditional GAN providing precise property-controlled generation and manufacturing process optimization, (4) Attention-Enhanced GAN preserving long-range dependencies and structural connectivity through selective feature weighting, (5) Style-based GAN offering hierarchical control from coarse-to-fine features with high-fidelity generation, and (6) Hybrid Architecture GAN combining multiple paradigms to address training stability, data scarcity, and architectural limitations.}
    \label{fig:taxonomy}
\end{figure}

The evolution of these GAN-based approaches can be systematically organized based on architectural innovations that address specific challenges in porous media reconstruction (Figure \ref{fig:taxonomy}). The classification reflects a chronological and conceptual progression from 2017 to 2026, with each successive paradigm building upon the foundations of earlier implementations to address key challenges including multi-scale feature representation, property control, computational efficiency, and training stability. The following sections examine each category in detail.

\pagebreak

\section{Vanilla GAN}
\label{Vanilla GAN}

Generative Adversarial Networks (GANs), introduced by Goodfellow et al. \citep{Goodfellow2014GenerativeNetworks} in 2014, established a new framework in generative modeling that has since become widely adopted in the field of artificial intelligence. The framework implements an adversarial training process between two neural networks, a generator network (G) that creates synthetic samples by learning the underlying data distribution, and a discriminator network (D) that functions as a classifier to distinguish between real and generated samples.

\begin{figure}[H]
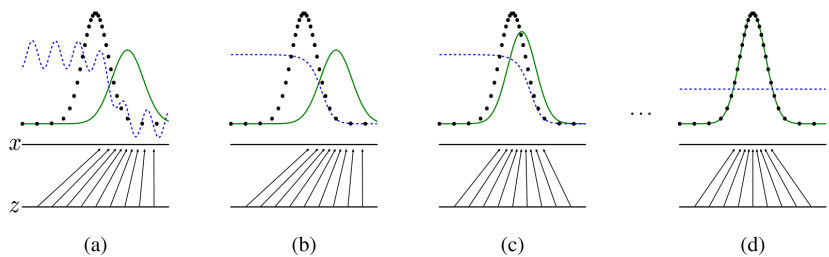

    \centering
    \includecroppedfigure[width=0.9\textwidth]
    \caption{Generative adversarial nets training process. Blue dashed line represents the discriminative distribution (D), black dotted line shows the data generating distribution, and green solid line indicates the generative distribution (G). The four stages illustrate progression from (a) initial partially-trained networks through (b,c) alternating discriminator and generator updates to (d) Nash equilibrium where $p_g = p_{data}$. (Figure adapted from Goodfellow et al. \citep{Goodfellow2014GenerativeNetworks})}
    \label{fig:GAN}
\end{figure}

The unique aspect of GANs lies in their adversarial training dynamics, where both networks continuously improve through competition. The generator learns to produce increasingly realistic samples while the discriminator enhances its ability to distinguish real from synthetic data. This process evolves through distinct phases, from initial partially-trained networks to a final equilibrium where the generator perfectly matches the data distribution. Figure \ref{fig:GAN} visualizes this evolution, showing how the discriminative distribution (blue dashed line), data generating distribution (black dotted line), and generative distribution (green solid line) change throughout training.

The adversarial optimization process illustrated in Figure \ref{fig:GAN} is mathematically formulated as:

\begin{equation}\label{eq:adversarial_loss}
L_{\text{adv}}(\theta, \phi) = \mathbb{E}_{x \sim p_{\text{data}}(x)}\left[\log(D_{\theta}(x))\right] + \mathbb{E}_{z \sim p_{z}(z)}\left[\log(1 - D_{\theta}(G_{\phi}(z)))\right]
\end{equation}

In this formulation, the generator network transforms random noise $z$ (sampled from the distribution shown on the lower horizontal line in Figure.~\ref{fig:GAN}) into synthetic samples $G_{\phi}(z)$, while the discriminator network produces probability values between 0 and 1 to classify samples as real or synthetic. The mapping $x = G(z)$ imposes a non-uniform distribution $p_g$ on the transformed samples, with the generator contracting in regions of high density and expanding in regions of low density, as indicated by the upward arrows in Figure~\ref{fig:GAN}.

\begin{figure}[H]
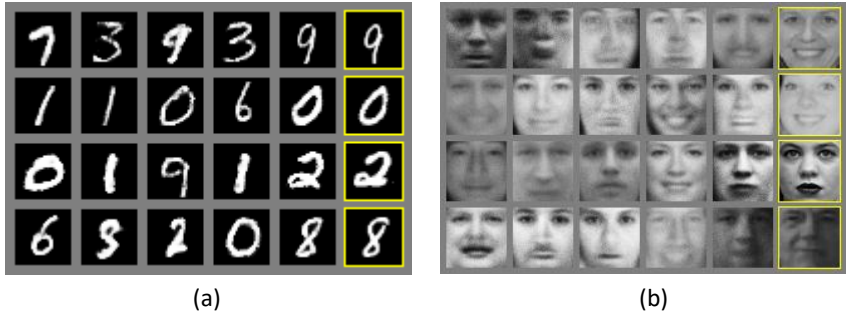

    \centering
    \includecroppedfigure[width=0.9\textwidth]
    \caption{Examples of samples generated by GAN model. (a) MNIST digit samples generated by GAN, with the yellow-highlighted column showing the closest training example to demonstrate the model isn't simply memorizing the training data. (b) Face samples generated from the Toronto Face Database (TFD), with the yellow-highlighted column again showing the nearest training examples. These examples demonstrate the GAN's ability to generate realistic synthetic data across different domains. (Figure adapted from Goodfellow et al. \citep{Goodfellow2014GenerativeNetworks})}
    \label{fig:PrimGAN}
\end{figure}

Examples of MNIST digits and TFD faces generated by GANs are shown in Figure \ref{fig:PrimGAN}. Left panel shows generated handwritten digits, while right panel displays synthetic face images with various expressions. The rightmost column in each panel (highlighted in yellow) shows the nearest training example to each generated sample, demonstrating that the model is creating novel samples rather than memorizing training data.

The implementation of GANs has evolved into two primary architectural approaches. The first approach utilizes fully connected networks, incorporating dense layers where each input node connects to every node in subsequent layers. This architecture proves effective for simpler datasets. The second approach, known as Deep Convolutional GANs (DCGANs), replaces Multi-Layer Perceptrons with Convolutional Neural Networks in both generator and discriminator components\cite{Goodfellow2014GenerativeNetworks, Radford2015UnsupervisedNetworks}. 

The application of GANs, particularly DCGANs, to microstructure reconstruction represents a significant advancement in materials science and engineering. DCGANs demonstrate enhanced performance in image generation tasks, particularly in digital rock reconstruction. In the context of porous media reconstruction, vanilla GAN architectures have evolved through four distinct methodological paradigms, each addressing specific challenges and introducing novel solutions to reconstruction problems.

\subsection{Foundational 3D DCGAN Architectures}

\begin{figure}[H]
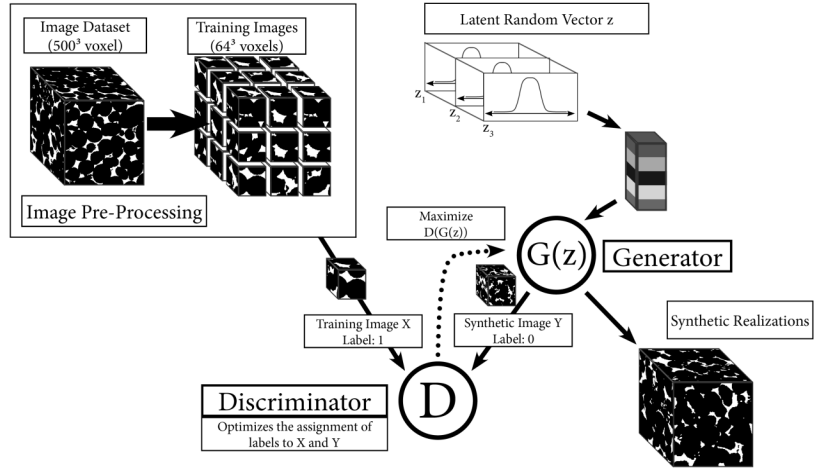

    \centering
    \includecroppedfigure[width=0.9\textwidth]
    \caption{Schematic representation of the GAN architecture for 3D porous media reconstruction. The generator network (left) transforms random noise z into synthetic 3D structures through sequential transpose convolution layers, while the discriminator network (right) evaluates whether samples are real or generated through volumetric convolution layers. (Figure adapted from Mosser et al. \citep{Mosser2017ReconstructionNetworks})}
    \label{fig:mosser1}
\end{figure}

\begin{figure}[H]
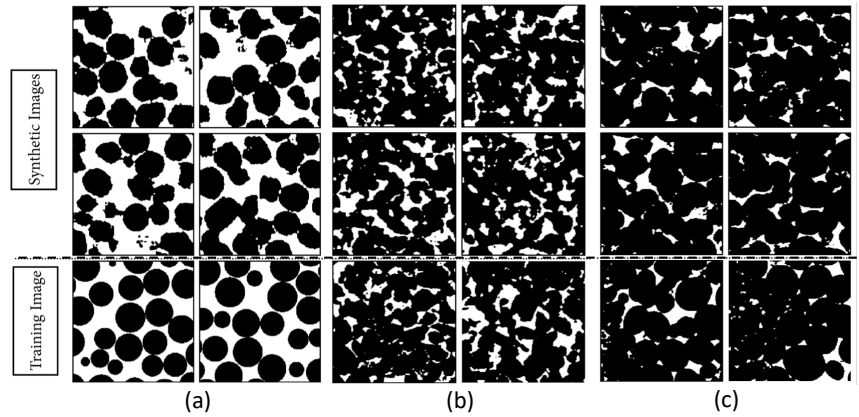

    \centering
    \includecroppedfigure[width=0.9\textwidth]
    \caption{Comparison of original (top) and GAN-generated (bottom) 3D porous media samples: (a,d) Beadpack showing uniform spherical particles, (b,e) Berea sandstone displaying complex intergranular pore structures, (c,f) Ketton limestone exhibiting intricate pore networks. The generated samples demonstrate visual similarity to the training data while maintaining key morphological and statistical properties. (Figure adapted from Mosser et al. \citep{Mosser2017ReconstructionNetworks})}
    \label{fig:mosser2}
\end{figure}

The work by Mosser et al. \cite{Mosser2017ReconstructionNetworks} introduced the first application of GANs for three-dimensional porous media reconstruction (Fig. \ref{fig:mosser1}). Their architecture employed fully convolutional neural networks for both generator and discriminator, with the generator using transposed convolutions and batch normalization, while the discriminator used volumetric convolution layers. They demonstrated successful reconstruction of beadpack (128$^3$ voxels), Berea sandstone (64$^3$ voxels), and Ketton limestone (64$^3$ voxels) samples (Fig. \ref{fig:mosser2}). Their method accurately captured key properties, achieving porosity differences within 1\% of original samples, matching two-point correlation functions, and reproducing similar permeability distributions, while maintaining computational efficiency for generating large samples.

Building upon this foundational work, Mosser et al. \cite{Mosser2018StochasticNetworks} extended their approach to unsegmented grayscale micro-CT data, reconstructing oolitic Ketton limestone samples. This advancement demonstrated that GANs could work directly with continuous grayscale intensity values rather than requiring binary segmented images, achieving successful reconstruction up to 450$^3$ voxels. The method preserved morphological characteristics through Minkowski functionals evaluation, showing that GANs could capture complex microstructural features in grayscale porous media.

Following these pioneering works, Liu et al. \cite{Liu2019ANetworks} investigated the application of modified DCGAN architectures to both homogeneous (Berea sandstone) and heterogeneous (Estaillades carbonate) reservoir samples. Their key innovation involved implementing data augmentation through 3D rotation of training samples, which improved the network's ability to capture diverse structural features. They successfully generated large-scale reconstructions up to 528$^3$ voxels from training data of 64$^3$ and 128$^3$ voxels, demonstrating the scalability of the approach for different rock types.

Valsecchi et al. \cite{Valsecchi2020StochasticGANs} presented an improved 3D-to-3D GAN reconstruction method that refined the architectural design for generating porous media including beadpack, Berea sandstone, and Ketton limestone. Their architecture employed upsampling layers combined with 3D convolutions in the generator, while the discriminator utilized multiple convolutional blocks with batch normalization and dropout for enhanced training stability. The implementation demonstrated improved reconstruction quality compared to earlier methods while maintaining computational efficiency.

Application-specific implementations of foundational DCGAN architectures emerged across diverse material domains. Zhao et al. \cite{Zhao20213DLearning} developed a specialized DCGAN implementation for tight sandstone reconstruction, achieving approximately 4.5\% porosity matching with original samples while successfully preserving pore connectivity and geometry. Their architecture employed deconvolution layers in the generator with kernel size of 2$\times$2$\times$2 and stride of 2, enabling progressive upsampling from initial noise vectors to 32$^3$ voxel output volumes. Samaei et al. \cite{Samaei2021MechanicalFEM} applied DCGAN to generate porous microstructures of zirconia-silica bilayer coatings on aluminum alloys, combining the GAN framework with mechanics of structure genome for multiscale modeling of mechanical performance.

Two significant training innovations emerged to address computational efficiency and stability challenges. Zhang Q. et al. \cite{Zhang2022DigitalNetworks} introduced a pretraining strategy for discriminators that substantially reduced computational requirements, achieving comparable reconstruction quality with only 800 training iterations compared to thousands traditionally required. Their approach involved pretraining the discriminator on a larger dataset before joint adversarial training, enabling the network to learn discriminative features more efficiently. The method maintained porosity accuracy within 0.15\% of ground truth for Berea sandstone samples while reducing training time by over 60\%.

Zhang T. et al. \cite{Zhang2022AWGAN-GP} introduced an improved WGAN-GP architecture addressing mode collapse and gradient instability in porous media reconstruction. Their key innovations included replacing the original multi-layer perceptron with convolutional neural networks and incorporating BatchNorm3d in both generator and discriminator to enhance training stability. The generator employed four ConvTranspose3d layers (kernel size 4, stride 2) while the discriminator mirrored this structure with Conv3d layers. The method achieved porosity accuracy within 0.1-0.3\% of reference values for shale and Berea sandstone samples, demonstrating superior reconstruction quality compared to traditional multiple-point statistics methods.

Addressing the field-of-view versus resolution trade-off through super-resolution, Zhang T. et al. \cite{Zhang2023ANetwork} developed 3DRGAN combining residual networks with a dual discriminator architecture for shale reconstruction. The generator employs 12 residual blocks with Conv3D layers and trilinear upsampling achieving $2\times$ resolution enhancement ($40^3$ to $80^3$ voxels). The architecture utilizes two discriminators, an adversarial discriminator employing WGAN-GP loss and a feature discriminator extracting perceptual features. Validation on shale samples demonstrated superior reconstruction quality compared to DCGAN and SNESIM across pore space characteristics, multi-point connectivity, and permeability matching, while maintaining improved diversity compared to SRGAN.

\begin{figure}[H]
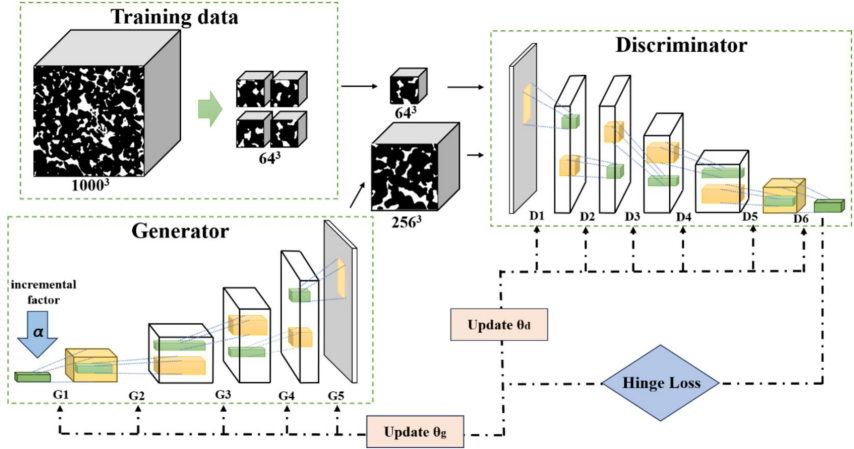

    \centering
    \includecroppedfigure[width=0.9\textwidth]
    \caption{Schematic of three-dimensional porous media image generation using a DCGAN algorithm. The framework consists of three main components: (1) Training data preparation where a 1000$^3$ voxel image is divided into 64$^3$ subvolumes, (2) A generator network (G1-G5) that processes data with an incremental factor $\alpha$ to enable upscaling, and (3) A discriminator network (D1-D6) that evaluates image quality through multiple stages. The network parameters $\theta_g$ and $\theta_d$ are updated iteratively using a Hinge Loss function to optimize generation quality. (Figure adapted from Zhang et al. \citep{Zhang2025PredictionModelling})}
    \label{fig:Zhang2025PredictionModelling1}
\end{figure}

Zhang et al. \cite{Zhang2025PredictionModelling} introduced a significant architectural advancement addressing volume upscaling limitations in DCGAN for CO$_2$ storage efficiency prediction (Fig. \ref{fig:Zhang2025PredictionModelling1}). Their novel enlargement factor approach enabled generation of 512$^3$ voxel images from 64$^3$ voxel training samples, representing an 8$^3$ volume upscaling factor. The enlargement factor was mathematically formulated as $\alpha = (\lambda - 64)/16 + 1$, where $\lambda$ represents the desired output size. The architecture incorporated an intermediate generation step where 256$^3$ voxel images were fed to the discriminator during training. The key innovation was utilization of Hinge Loss function, which facilitated training by handling size disparity between training data and generated images through a weight-averaging mechanism and constrained optimization margin. The method achieved relative differences less than 10\% in morphological properties (porosity, specific surface area, Euler characteristic) and residual CO$_2$ saturation predictions (0.354 $\pm$ 0.014) closely matching experimental values (0.390 $\pm$ 0.016). Integration with pore network modeling enabled comprehensive flow property analysis including relative permeability and capillary pressure predictions. The approach generated an ensemble of 1000 realizations (512$^3$ voxels each) for uncertainty quantification, determining minimum number of simulations needed for representative statistics.

These foundational 3D DCGAN architectures established the core principles for direct 3D-to-3D reconstruction of porous media, demonstrating that GANs could effectively learn complex three-dimensional microstructural features from training data and generate statistically similar synthetic samples. The progression from binary to grayscale reconstruction, incorporation of training stability improvements through WGAN-GP and Hinge Loss, successful application across diverse material systems, and development of volume upscaling capabilities validated the versatility of this approach. Integration with physics-based modeling frameworks further extended their utility for property prediction and uncertainty quantification. However, these methods required access to 3D training data obtained through expensive imaging techniques such as micro-CT or FIB-SEM tomography, motivating the development of alternative approaches that could leverage more accessible 2D imaging data.

\subsection{SliceGAN and 2D Discriminator Architectures}

This paradigm generates three-dimensional structures from readily available two-dimensional images by employing 2D discriminators that evaluate cross-sectional slices of GAN-generated 3D volumes, thereby maintaining reconstruction quality without requiring volumetric training data.

Valsecchi et al. \cite{Valsecchi2020StochasticNetworks} demonstrated the first successful implementation of 2D-to-3D reconstruction for porous media through a GAN architecture that evaluated 2D cross-sections while generating 3D volumes. Their key innovation was designing a discriminator that operated on 2D cross-sections while the generator produced complete 3D structures. This approach leveraged the insight that examining cross-sections along all three axes could effectively assess the quality of reconstructed 3D structures. The architecture employed a 3D generator with upsampling layers and Conv3D operations, while the 2D discriminator processed 64$\times$64 pixel slices through multiple convolutional blocks with batch normalization and dropout. The method successfully generated beadpack, Berea sandstone, and Ketton limestone structures while maintaining accurate two-point correlation functions and requiring only 2D training images.

Coiffier et al. \cite{Coiffier20203DNetworks} advanced this paradigm through their DiAGAN (Dimension Augmenter GAN) framework for geological materials reconstruction. DiAGAN implemented a random cut sampling mechanism between generator and discriminator networks, where the 3D generator output was randomly sliced along arbitrary planes before being evaluated by the 2D discriminator. The generator architecture employed four Conv3D layers with instance normalization and 2$\times$ upscaling, while the discriminator processed cuts using five Conv2D layers with global max pooling. This approach enabled generation of images of any size without retraining by providing latent noise vectors of different dimensions. The method utilized WGAN-GP loss function for training stability and successfully reconstructed various geological materials including packed spheres, folded layers, sand grains, channels, and sedimentary deposits.

For electrochemical materials, Sciazko et al. \cite{Sciazko2021UnsupervisedImage} developed a GAN architecture combining a 3D generator with a 2D discriminator for Ni-GDC SOFC anode reconstruction. The generator employed five transposed convolution blocks with softmax activation for three-phase generation (64$^3$ voxels), while the 2D discriminator processed 64$\times$64 pixel cross-sections. The key innovation was validating cross-sections in all spatial directions during training, ensuring 3D structural coherence from 2D SEM images alone.

\begin{figure}[H]
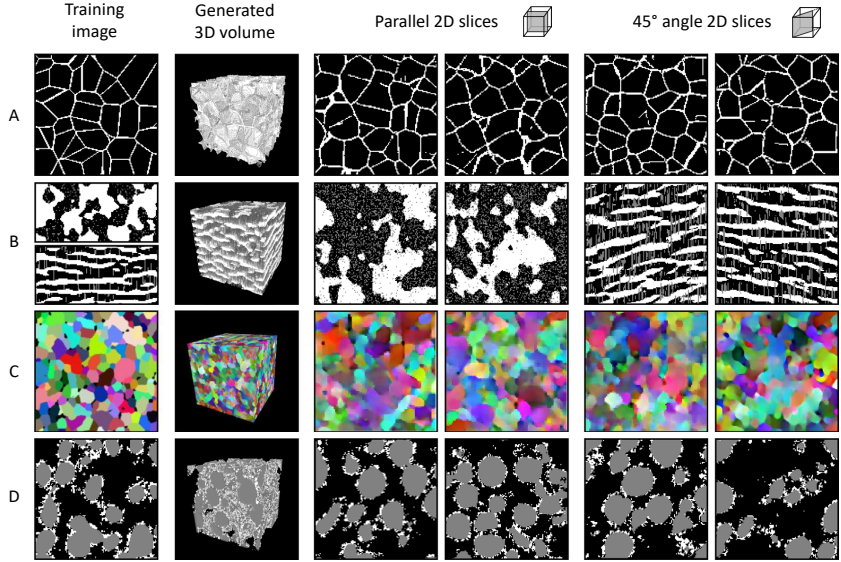

    \centering
    \includecroppedfigure[width=0.9\textwidth]
    \caption{Representative results from SliceGAN showing microstructure generation across different material systems. From top to bottom: (A) synthetic polycrystalline grains, (B) anisotropic battery separator material, (C) polycrystalline steel with orientation coloring, and (D) NMC battery cathode. For each row: training image (left), generated 3D volume, parallel 2D slices showing consistency with training data, and 45° angle slices demonstrating 3D structural coherence. (Figure adapted from Kench et al. \citep{Kench2021GeneratingExpansion})}
    \label{fig:Kench2021GeneratingExpansion1}
\end{figure}

The SliceGAN framework, introduced by Kench et al. \cite{Kench2021GeneratingExpansion}, marked a significant architectural advancement by implementing orthogonal 2D discriminators to evaluate generated 3D structures through cross-sectional views (Fig. \ref{fig:Kench2021GeneratingExpansion1}). The framework employed a slicing operation that extracted sections along x, y, and z directions at 1-voxel increments before feeding them to the 2D discriminator. The generator consisted of five 3D transpose convolution layers (4$\times$4$\times$4$\times$64 $\to$ 64$\times$64$\times$64$\times$3, kernel size 4, stride 2), while the discriminator utilized five Conv2D layers processing 64$\times$64 pixel slices. A critical technical contribution was their concept of uniform information density, which ensured consistent quality throughout the generated volume through careful selection of transpose convolutional parameters. The method demonstrated effectiveness across diverse materials including polycrystalline grains, battery separators, steel, and battery cathodes, achieving statistical similarity with real 3D datasets while requiring only 2D training data.

The SliceGAN framework inspired several subsequent implementations addressing specific application domains. Sugiura et al. \cite{Sugiura2022HourlySliceGAN} applied SliceGAN to dual-phase steel visualization, demonstrating that 3D microstructural reconstruction could be reduced from a daily task (via serial sectioning) to an hourly process, with generated structures showing volume fractions of martensite in good agreement with experimentally reconstructed 3D images. Li et al. \cite{Li2022DigitalPenalty} implemented WGAN-GP for digital rock reconstruction from 2D images of Berea sandstone and Ketton limestone. The use of Wasserstein distance provided stable training performance, with generated image quality correlating with discriminator loss. Their integrated framework automated the entire workflow from training set generation to synthetic rock validation, demonstrating accurate reconstruction in terms of two-point correlation and morphological properties.

Recent implementations have emphasized rigorous validation frameworks combining GAN generation with statistical controllers and experimental measurements. Li et al. \cite{Li20243DLearning} adapted SliceGAN for SEM image reconstruction of porous ceramic materials (SOFC electrode materials and ceramic atomizer cores), developing a systematic workflow requiring only single 2D cross-sectional images. Their method combined GAN generation with multi-criteria statistical controllers using four two-point descriptors, enabling validation against mercury intrusion porosimetry measurements for porosity, pore size distribution, and tortuosity factor. Xu et al. \cite{Xu2024Multi-criteriaProperties} developed a similar multi-criteria guided approach for cementitious materials, integrating DCGAN with statistical controllers that evaluated generated 64$^3$ voxel structures using four two-point descriptors (two-point probability function, two-point cluster function, lineal-path function, and chord-length distribution) from single 2D images. Training utilized WGAN-GP loss function with up to 60,000 iterations for convergence, successfully predicting transport properties (diffusion coefficients, permeability) aligned with experimental data for hardened cement paste and cellular concrete, demonstrating physics-informed 2D-to-3D reconstruction with property preservation. Amiri et al. \cite{Amiri2024NewGANs} enhanced the SliceGAN architecture for heterogeneous media reconstruction by developing a workflow relying exclusively on 2D images from three orthogonal sections. Their approach combined high-resolution electron microscopy and optical imaging of Berea sandstone samples with modified SliceGAN architecture. The method achieved close agreement with $\mu$CT measurements in terms of porosity (within 0.6\%), pore connectivity, and spatial correlation functions, demonstrating potential for assessing variability in heterogeneous rocks.

Recent applications have extended SliceGAN to specialized geological formations with distinct microstructural characteristics. Zhou et al. \citep{Zhou2025DigitalSliceGAN} applied SliceGAN to tight carbonate rocks from the Ordos Basin, integrating MAPS and QEMSCAN imaging to create three-phase segmented training data distinguishing pores, dolomite, and calcite. This multi-modal imaging approach addressed the challenge of accurately representing heterogeneous mineral distributions alongside pore structures. The implementation achieved porosity matching within 1\% (generated: 7-8\% vs. original: 8\%) with strong consistency in auto-correlation functions and Minkowski functionals, though formation factor calculations revealed variability (775-5894 across three axes), indicating inherent stochasticity in single-image reconstruction.

Addressing organic matter (OM) characterization in shales, Liu et al. \citep{Liu2025Three-dimensionalShales} developed a workflow combining SliceGAN with high-resolution MAPS imaging (16 nm resolution) for REV-size ($\approx$100 $\mu$m) OM reconstruction. Their innovative approach classified individual OM watersheds into three types based on surface porosity: Type A ($>$20\%), Type B (10--20\%), and Type C ($<$10\%). The method processed 196$\times$196 $\mu$m$^2$ images through watershed segmentation, substantial coarsening (20$\times$ resolution reduction), and SliceGAN reconstruction validated against FIB-SEM data with porosity error $<$5\%. Applied to Longmaxi Formation shales, the approach revealed that Type C OM (low porosity) exhibited superior connectivity at REV scale despite lower individual watershed permeability, demonstrating that 2D-disconnected OM can achieve 3D connectivity essential for shale gas flow modeling while reducing imaging costs through reconstruction from 614$\times$614 pixel 2D images.

Distinct methodological innovations extended the SliceGAN paradigm for specialized challenges. Liu et al. \cite{Liu20223DLearning} developed an innovative approach for fuel cell catalyst layer reconstruction by combining DCGAN with spatial interpolation in latent space. Their method generated spatially-continuous microstructure slices through interpolation between different latent vectors, enabling reconstruction of 512$\times$512$\times$600 voxel volumes from 2D training images with phase ratio accuracy within 2\% of real samples and demonstrated structural similarity in porosity, particle size distribution, and tortuosity.

Addressing multi-modal imaging integration, Chi et al. \cite{Chi2024DigitalFramework} developed a 2D-3D fusion framework combining a lightweight 3D super-resolution generator with residual architecture and a 2D discriminator for integrating low-resolution CT with high-resolution SEM images of tight sandstone. The generator employs Conv3D layers (kernel $3\times3\times3$) with upsampling modules achieving $4\times$ to $8\times$ resolution enhancement (up to $512^3$ voxels), trained using WGAN-GP loss combined with MSE reconstruction loss. Validation demonstrated accurate preservation of both macro-scale CT structures and micro-scale SEM features, with close agreement in porosity, pore connectivity, and permeability measured through Lattice Boltzmann simulation.

Extending 2D-to-3D reconstruction to large-scale anisotropic materials, Chi et al. \cite{Chi2024ReconstructionNetwork} developed a GAN-based method for reconstructing anisotropic 3D digital rocks (up to 1024$^3$ voxels) using only 2D shale images. The key innovation was their overlapping splitting method, where input random noise was divided with overlap to generate small-scale digital rocks (64$^3$ voxels) which were then concatenated into larger-scale 3D structures. Training utilized multiple discriminators (one for each principal direction) to preserve anisotropy. The method's effectiveness was validated through permeability simulations, demonstrating preservation of directional properties with values of 5.32$\times$10$^{-4}$ mD, 8.55$\times$10$^{-4}$ mD, and 1.65$\times$10$^{-6}$ mD in x, y, z directions respectively, accurately capturing the anisotropic nature of shale.

The 2D discriminator architecture paradigm represents a fundamental shift from 3D-to-3D reconstruction approaches, offering several key advantages: (1) elimination of expensive 3D imaging requirements, (2) ability to work with readily available 2D microscopy images, (3) maintenance of 3D structural coherence through orthogonal cross-section evaluation, and (4) scalability to large reconstruction volumes. The widespread adoption of WGAN-GP loss function across these implementations (appearing in 6 out of 10 papers in this category) indicates convergence toward stable training methodologies. These architectures have successfully demonstrated applicability across diverse material systems including geological samples, electrochemical materials, ceramics, and metallic alloys, validating the versatility and robustness of the 2D discriminator approach for porous media reconstruction.

\subsection{Multi-Phase and Complex Material Architectures}

The reconstruction of multi-phase materials and topologically complex microstructures presents unique challenges beyond binary porous media, requiring architectural modifications to handle multiple constituent phases, intricate connectivity patterns, and preservation of spatial coherence. Three significant implementations have advanced GAN capabilities in this domain through novel architectural designs and specialized loss functions.

\begin{figure}[H]
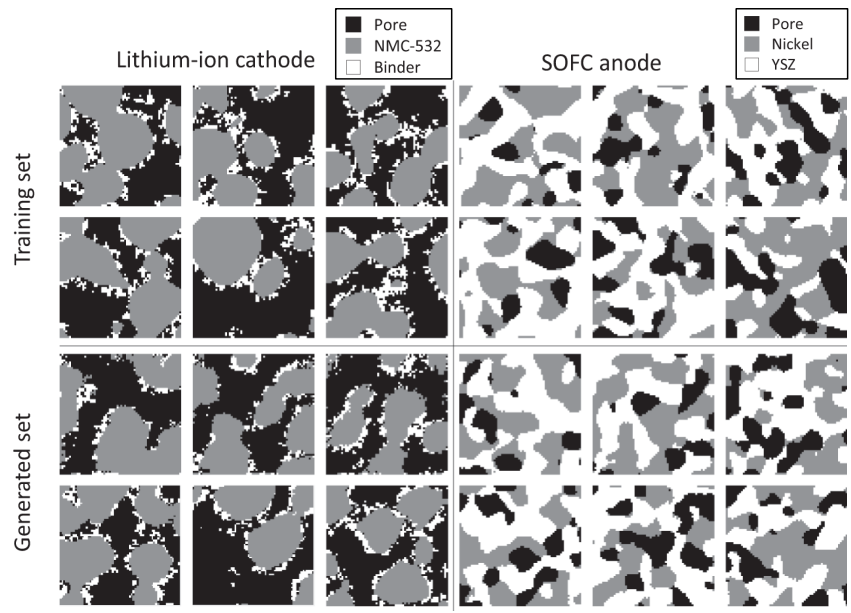

    \centering
    \includecroppedfigure[width=0.9\textwidth]
    \caption{Comparison between training and generated microstructures for lithium-ion battery cathode (left) and solid oxide fuel cell (SOFC) anode (right). Top row: Original training set images obtained from experimental data. Bottom row: Synthetic microstructures generated by GAN model. The generated structures demonstrate visual similarity to the training data while maintaining distinct features for each material system. (Figure adapted from Gayon-Lombardo et al. \cite{Gayon-Lombardo2020PoresBoundaries})}
    \label{fig:Gayon-Lombardo2020PoresBoundaries1}
\end{figure}

The first successful multi-phase 3D microstructure generation for electrochemical applications was achieved by Gayon-Lombardo et al. \cite{Gayon-Lombardo2020PoresBoundaries}, who developed architectures for lithium-ion battery cathodes (containing NMC particles, carbon-binder domain, and pore space) and SOFC anodes (containing Ni, YSZ, and pore phases) (Fig. \ref{fig:Gayon-Lombardo2020PoresBoundaries1}). Their generator employed five ConvTranspose3d layers with softmax activation in the final layer to produce probabilistic phase assignments across three channels, while the discriminator processed the three-phase input through five Conv3d layers. Training employed binary cross-entropy loss with label smoothing to improve stability. The method achieved volume fraction accuracy typically within 1-2\% error across all phases and successfully implemented periodic boundary conditions for seamless tiling, enabling generation of arbitrarily large volumes while maintaining phase connectivity and morphological characteristics. Building upon this foundation, Hsu et al. \cite{Hsu2021MicrostructureMaterials} advanced multi-phase GAN capabilities to practical scales for SOFC anodes at 96$^3$ voxel resolution. Their implementation employed spectral normalization in discriminator layers (ConvSN3D) to enhance training stability and utilized WGAN loss function. A critical achievement was accurate reproduction of triple-phase boundary (TPB) density (3.06$\pm$0.23 versus 3.16$\pm$0.20 $\mu$m/$\mu$m$^3$ experimental), which is essential for electrochemical performance. The method demonstrated superior performance compared to DREAM.3D algorithm in visual, statistical, and electrochemical property fidelity, establishing feasibility for practical-scale integrated computational materials engineering (ICME) applications.

\begin{figure}[H]
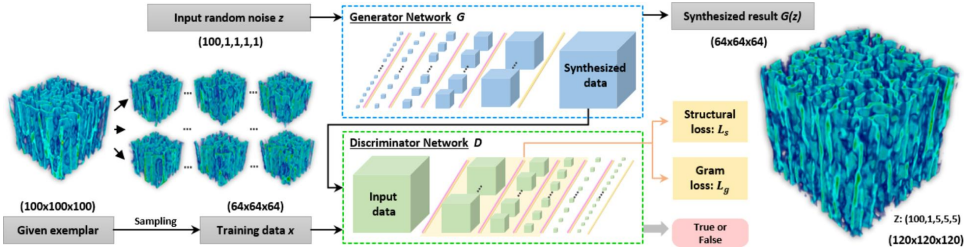

    \centering
    \includecroppedfigure[width=1.0\textwidth]
    \caption{Schematic overview of the ScaffoldGAN framework showing the generator network G that takes random noise z as input, discriminator network D, and the combined loss functions ($L_{adv}$, $L_g$, $L_s$) used for training. The framework generates synthesized 3D scaffold structures while maintaining spatial coherence through structural and gram losses. (Figure adapted from Zhang et al. \citep{Zhang2021ScaffoldGAN:Networks}, with permission)}
    \label{fig:Zhang2021ScaffoldGAN:Networks1}
\end{figure}

For biomedical scaffold materials, Zhang et al. \cite{Zhang2021ScaffoldGAN:Networks} introduced ScaffoldGAN, incorporating novel structural loss terms to enforce spatial coherence in complex porous structures (Fig. \ref{fig:Zhang2021ScaffoldGAN:Networks1}). Their architecture addressed the specific challenge that biological scaffolds require not only appropriate porosity and pore size distribution but also strong connectivity and mechanical integrity. The key innovation was a combined loss function:

\begin{equation}
L_{total} = L_{adv} + \lambda_gL_g + \lambda_sL_s
\end{equation}

where $L_{adv}$ represents standard adversarial loss, $L_g$ denotes gram loss for visual appearance preservation, and $L_s$ represents structural loss for spatial coherence, with hyperparameters $\lambda_g = 0.1$ and $\lambda_s = 1.0$. The gram loss computed correlations between feature maps across different layers, ensuring consistent visual patterns throughout the volume, while the structural loss enforced connectivity by penalizing disconnected regions and ensuring trabecular-like patterns characteristic of bone scaffolds. The method successfully generated scaffold structures for vertebral bone samples and metal foams (Ni/Cu/Al), achieving porosity errors averaging around 0.6\% across different bone samples while maintaining strong spatial coherence and connectivity metrics. The fully convolutional architecture enabled generation of structures up to 420$^3$ voxels, substantially larger than training data.

These implementations demonstrate distinct approaches to multi-phase and complex material reconstruction. Approaches include softmax-based probabilistic phase assignment with periodic boundaries for electrochemical materials, spectral normalization for practical-scale stability, and explicit structural constraints through multi-component loss functions for biomedical applications. Common across all implementations was the recognition that multi-phase materials require architectural considerations beyond binary porous media, including multi-channel outputs for phase representation, validation of phase-specific properties (volume fractions, TPB density, connectivity metrics), and careful balance between visual similarity and structural integrity. The successful application across diverse domains validates the versatility of these architectural innovations while highlighting the importance of domain-specific validation metrics for ensuring physically realistic multi-phase reconstructions.

\subsection{Current Limitations and Future Directions}

Training instability and mode collapse were the earliest documented challenges in vanilla GAN architectures for porous media. Mosser et al. \citep{Mosser2017ReconstructionNetworks, Mosser2018StochasticNetworks} identified these issues when using traditional binary cross-entropy loss functions, and Zhang T. et al. \citep{Zhang2022AWGAN-GP} confirmed that vanilla GANs suffer from mode collapse and gradient instability, necessitating architectures incorporating BatchNorm3d to normalise data distribution. These stability challenges have driven the widespread adoption of WGAN-GP variants across recent implementations \citep{Li2022DigitalPenalty, Xu2024Multi-criteriaProperties, Li20243DLearning, Chi2024ReconstructionNetwork, Amiri2024NewGANs}.

Resolution and scale limitations further constrain current implementations. Chi P. et al. \citep{Chi2024ReconstructionNetwork} documented challenges in reconstructing 3D pore structures at the nanoscale, which ``limit the study of rock physical properties in reservoirs,'' and developed an overlapping splitting method to concatenate small-scale digital rocks into larger-scale 3D structures up to $1024^3$ voxels. Zhang Y. et al. \citep{Zhang2025PredictionModelling} achieved an enlargement factor of $8^3$ relative to training data dimensions ($64^3$ voxels), but noted this required specialised DCGAN algorithms. The trade-off between sample size and resolution is particularly challenging for heterogeneous rocks, as Amiri H. et al. \citep{Amiri2024NewGANs} noted that ``limitations arise from the trade-off between sample size and resolution, particularly in heterogeneous rocks with multi-scale features where both high resolution and a large field of view are essential.''

Several architecture-specific future directions have been identified. Chi P. et al. \citep{Chi2024ReconstructionNetwork} demonstrated the potential of overlapping reconstruction for generating large-scale anisotropic 3D digital rocks. Zhang Y. et al. \citep{Zhang2025PredictionModelling} proposed combining GANs with pore network modelling to predict properties at larger scales, as evidenced by their successful integration of DCGAN with pore network modelling to predict $\text{CO}_2$ storage efficiency. The systematic workflows developed by Amiri H. et al. \citep{Amiri2024NewGANs} for purely 2D-based training and by Li X. et al. \citep{Li20243DLearning} for single cross-sectional image reconstruction represent important alternatives that reduce dependence on expensive 3D imaging data.

\pagebreak

\section{Multi-Scale GAN}
\label{Multi-Scale GAN}

The reconstruction of porous media presents unique challenges due to their inherent multi-scale nature, where features range from nanometer-scale pores to millimeter-scale structures, and complex structural characteristics. While conventional GANs have shown promise in this domain, they often struggle with computational efficiency and the requirement for extensive training data when dealing with larger datasets \citep{Xia2022Multi-scaleNetworks, Zhang2023StochasticMechanisms}. Multi-scale GANs (MS-GANs) have emerged as a solution to address these limitations, offering improved efficiency and capability in capturing hierarchical features across different scales while requiring minimal training data.

MS-GANs employ hierarchical architectures that process images at multiple resolutions simultaneously or progressively. The fundamental principle involves decomposing the generation task into scale-specific sub-problems, where each scale captures different levels of detail. This approach originated from progressive growing techniques in computer vision \citep{Karras2017ProgressiveVariation}, later adapted for porous media through innovations like pyramid generators and single-image learning frameworks. The mathematical formulation extends traditional GANs by incorporating scale-dependent generators $G_s$ and discriminators $D_s$:

\begin{equation}
\min_{G_s} \max_{D_s} V(D_s, G_s) = \mathbb{E}_{x \sim p_{data}^s(x)}[\log D_s(x)] + \mathbb{E}_{z \sim p_z(z)}[\log(1-D_s(G_s(z)))]
\end{equation}

where $s$ denotes the scale level, and each generator-discriminator pair operates at a specific resolution, learning scale-specific features while maintaining coherence across scales.

The evolution of multi-scale GANs for porous media can be traced through two main architectural paradigms, each addressing specific aspects of the multi-scale reconstruction challenge.

\subsection{Progressive Growing and Pyramid-based Approaches}

\begin{figure}[H]
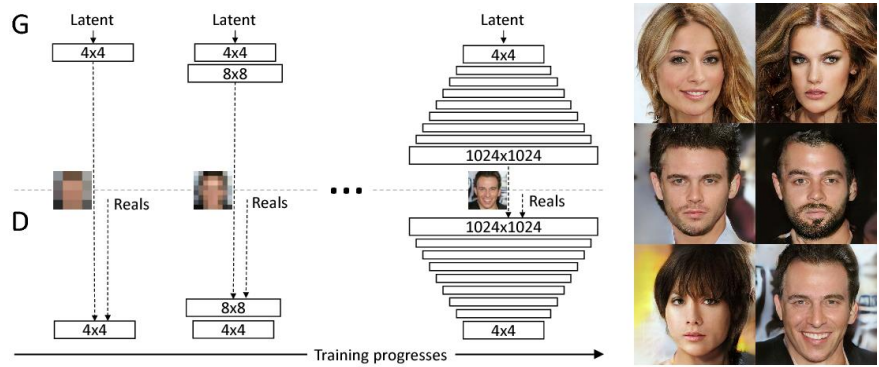

    \centering
    \includecroppedfigure[width=0.9\textwidth]
    \caption{Schematic illustration of the progressive growing training strategy for GANs. The methodology initiates with low-resolution networks (4×4 spatial resolution) for both generator (G) and discriminator (D), with subsequent incremental addition of convolutional layers to increase spatial resolution during training. All network layers remain trainable throughout the progressive growth process, enabling stable high-resolution synthesis. The right panel demonstrates representative outputs generated at 1024×1024 resolution, showcasing the framework's capability for high-quality image synthesis. (Figure adapted from Karras et al. \citep{Karras2017ProgressiveVariation}).}
    \label{fig:Karras2017ProgressiveVariation1}
\end{figure}

Karras et al. \citep{Karras2017ProgressiveVariation} introduced the progressive growing framework for GANs, establishing the foundational methodology of gradually increasing resolution during training by adding new layers to both generator and discriminator networks. As illustrated in Figure \ref{fig:Karras2017ProgressiveVariation1}, this incremental growth strategy begins with low-resolution networks (4×4 pixels) and progressively incorporates additional layers while maintaining trainability of all existing layers, thereby enabling stable convergence at resolutions up to 1024×1024 pixels. This approach stabilized training and enabled generation of high-quality images at unprecedented resolutions, setting the stage for applications in porous media reconstruction.

\begin{figure}[H]
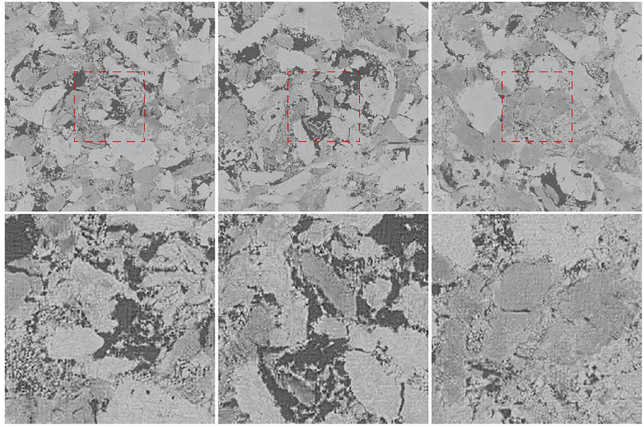

    \centering
    \includecroppedfigure[width=0.7\textwidth]
    \caption{Representative carbonate microstructure samples generated by the progressive growing GAN framework. Top row displays three independently synthesized 2D cross-sections at 1024×1024 resolution, with red dashed boxes indicating regions of interest. Bottom row presents magnified views of the corresponding marked regions, revealing the method's capability to preserve fine-scale pore structure details and phase boundary characteristics. The generated microstructures exhibit visual fidelity comparable to experimental carbonate images, demonstrating diverse micro-scale structural features 
    without resolution degradation. (Figure adapted from You et al. \citep{You20213DGAN}).}
    \label{fig:You20213DGAN1}
\end{figure}

You et al. \citep{You20213DGAN} pioneered the application of progressive growing GANs to carbonate digital rock reconstruction. Their implementation achieved a nine-fold speedup in imaging processes and over 4,500 times compression of image data. The architecture utilized linear interpolation of inverted latent vectors for 3D reconstruction, demonstrating that progressive growth maintains structural coherence across different scales of carbonate samples. The method proved particularly effective for capturing the complex pore networks characteristic of carbonate rocks, where dissolution features create multi-scale heterogeneity (Figure \ref{fig:You20213DGAN1}).

\begin{figure}[H]
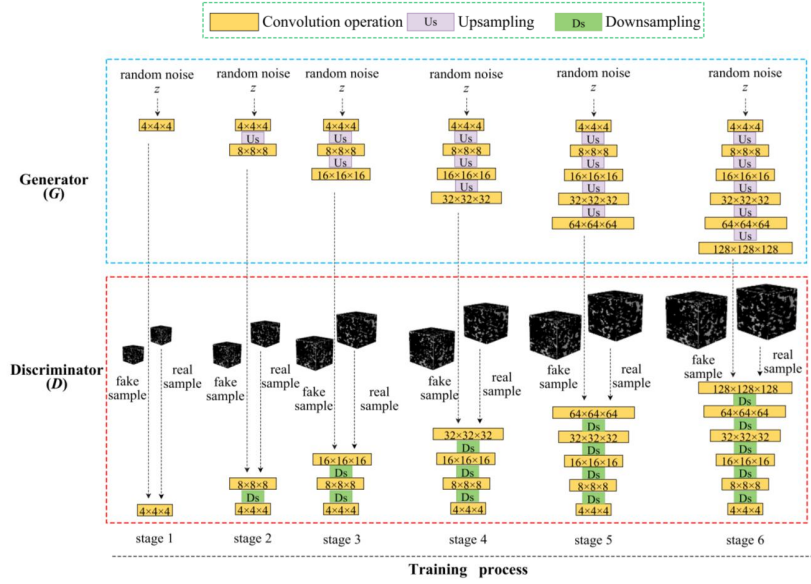

    \centering
    \includecroppedfigure[width=0.9\textwidth]
    \caption{The progressive growing architecture of MS-GAN for multi-scale porous media reconstruction. The network grows from 4×4×4 to 128×128×128 resolution through six stages, with generator (G) and discriminator (D) networks expanding via upsampling (Us) and downsampling (Ds) operations respectively. Random noise inputs (z) are processed through increasingly complex network paths across stages. (Figue adapted from Xia et al. \citep{Xia2022Multi-scaleNetworks}).}
    \label{fig:Xia1}
\end{figure}

Xia et al. \citep{Xia2022Multi-scaleNetworks} developed the MS-GAN framework with synchronized growth of generator and discriminator networks, processing volumes from $4 \times 4 \times 4$ to $128 \times 128 \times 128$ voxels through six progressive stages. Their architecture, illustrated in Figure \ref{fig:Xia1}, employed upsampling operations in the generator pathway and corresponding downsampling in the discriminator, maintaining balanced feature learning throughout the growth process. This approach significantly improved training stability compared to single-scale implementations, achieving consistent reconstruction quality across all resolution levels.

\begin{figure}[H]
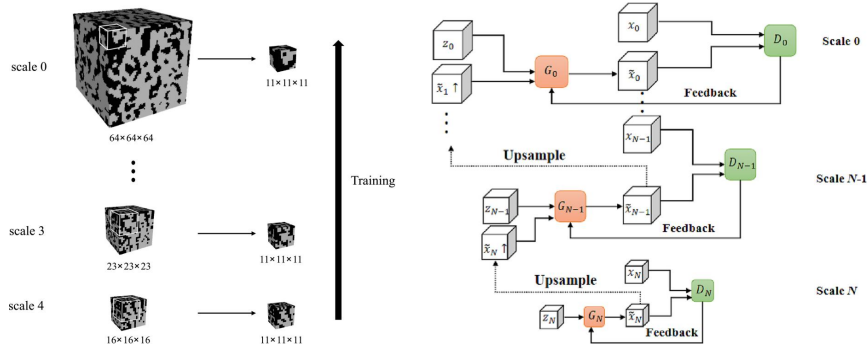

    \centering
    \includecroppedfigure[width=0.9\textwidth]
    \caption{Multi-scale feature extraction process using a fixed receptive field (11×11×11 voxels) across different scales. The input image sizes vary from 64×64×64 at the finest scale (scale 0) to 16×16×16 at the coarsest scale (scale 4), with the same receptive field used to capture features at each scale. (Figure adapted from Zhang et al. \citep{Zhang2023ReconstructionNetworks}, with permission).}
    \label{fig:Zhang2023ReconstructionNetworks1}
\end{figure}

\begin{figure}[H]
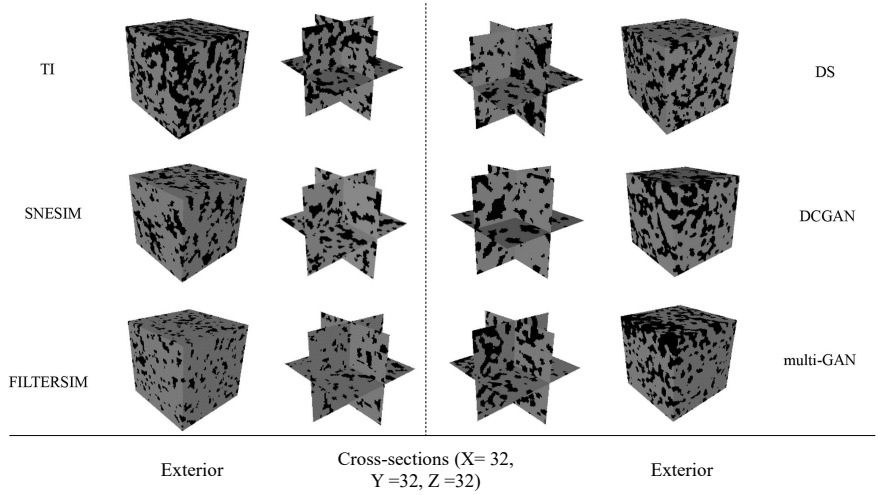

    \centering
    \includecroppedfigure[width=0.9\textwidth]
    \caption{Visual comparison of porous media reconstruction results using different methods: original training image (TI), SNESIM, FILTERSIM, DS, DCGAN, and multi-GAN. Each row shows the exterior view and cross-sectional view (X=32, Y=32, Z=32) of the reconstructed samples. (Figure adapted from Zhang et al. \citep{Zhang2023ReconstructionNetworks}, with permission).}
    \label{fig:Zhang2023ReconstructionNetworks2}
\end{figure}

Zhang et al. \citep{Zhang2023ReconstructionNetworks} introduced a progressive training MS-GAN that employed identical five-layer networks for both generator and discriminator components, operating across five scales from $16^3$ to $64^3$ voxels. Their framework utilized a fixed receptive field ($11 \times 11 \times 11$ voxels) to scan input images across different scales, capturing detailed features at finer scales while evaluating broader structural information at coarser scales (Figure \ref{fig:Zhang2023ReconstructionNetworks1}). The effectiveness of this approach is demonstrated through comparative reconstructions shown in Figure \ref{fig:Zhang2023ReconstructionNetworks2}, where results from different methods including traditional multiple-point statistics (SNESIM, FILTERSIM, DS) and deep learning approaches (DCGAN, multi-GAN) are compared. The visual comparison shows that MS-GAN maintains consistent structural similarity across both exterior and cross-sectional views, particularly in preserving both large-scale features and fine-grained pore distributions.

Recent developments in MS-GAN architectures have been enhanced by the introduction of the Improved Pyramid Wasserstein Generative Adversarial Network (IPWGAN) by Zhu et al. (2024) \citep{Zhu2024GenerationNetworks}. This approach combines three key innovations, a Laplacian pyramid generator for multi-scale feature capture, feature statistics mixing regularization (FSMR) for improved diversity, and Wasserstein GAN with Lipschitz penalty (WGAN-LP) for training stability. IPWGAN demonstrated better performance over conventional DCGAN method, reducing mean relative errors in permeability calculations by up to 79\% for Berea sandstone. The method successfully generated large-scale samples (2,200³ voxels) while maintaining both local and global features, though limitations persist in capturing features beyond the micrometer scale, such as fractures and laminations.

Addressing the cubic memory scaling problem that limited previous 3D super-resolution implementations, Ugolkov et al. \citep{Ugolkov2025OptimizedTomography} introduced a memory-efficient 3D Octree-Based Progressive Growing WGAN-GP for super-resolution enhancement of segmented micro-CT rock images. Their five-stage progressive framework incrementally increases resolution from $32^3$ to $512^3$ voxels, utilizing octree-based sparse convolutional operations to selectively process only ``mixed'' boundary nodes while storing homogeneous ``dense'' regions. The 3D generator operates with a 2D discriminator, enabling training with unpaired 3D low-resolution and 2D high-resolution images. Applied to Berea sandstone, the method achieved 16× super-resolution, increasing detected pore space from 5\% to 17\% with substantially improved pore-throat resolution. The octree implementation reduced memory consumption sufficiently to enable processing of $4096^3$ voxel volumes on standard GPU hardware, representing a substantial advancement over previous 8× limitations through selective refinement with sub-volume processing and seamless stitching.

\subsection{Single-Image Multi-Scale (SinGAN Variants)}

Rott Shaham et al. \citep{Shaham2019SinGAN:Image} introduced SinGAN, demonstrating that a single natural image contains sufficient information to train a generative model through multi-scale patch statistics. This breakthrough enabled training without large datasets, a critical advantage for porous media applications where obtaining extensive training data is challenging. Hinz et al. \citep{Hinz2020ImprovedGANs} improved this approach through concurrent training of multiple stages (ConSinGAN), reducing training time while maintaining generation quality through parameter sharing between stages.

\begin{figure}[H]
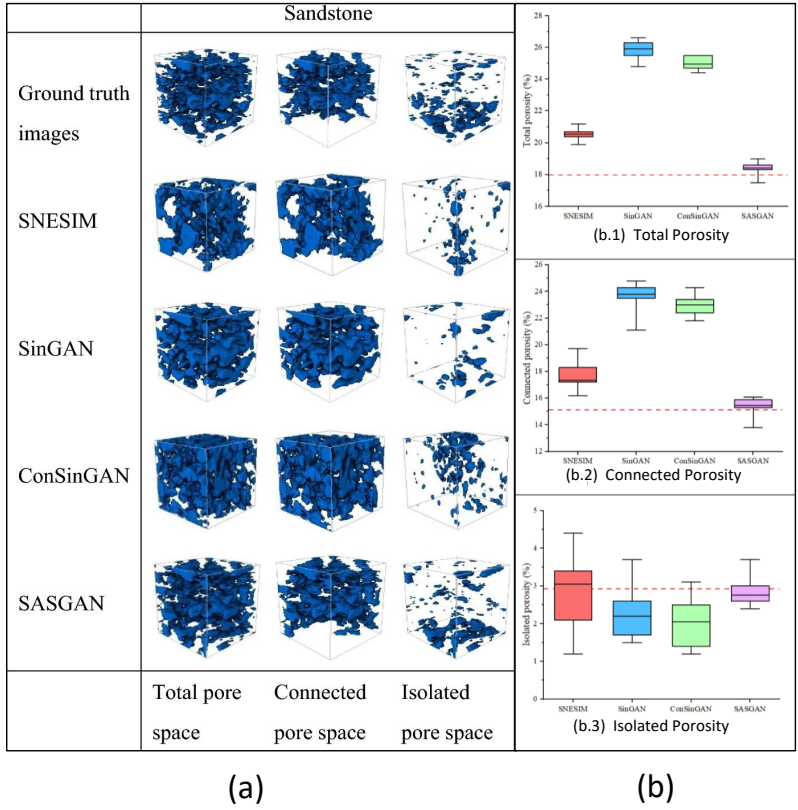

    \centering
    \includecroppedfigure[width=0.9\textwidth]
    \caption{Comparison of pore space characteristics for sandstone samples between ground truth and different reconstructions: (Left) Visual representation of total pore space, connected pore space, and isolated pore space for ground truth and reconstructed samples using SNESIM, SinGAN, ConSinGAN, and SASGAN; (Right) Quantitative comparison through boxplots showing (b.1) total porosity, (b.2) connected porosity, and (b.3) isolated porosity distributions. The horizontal red dotted lines indicate the corresponding porosity values from the ground truth sample, demonstrating SASGAN's ability to preserve pore space characteristics. (Figure adapted from Zhang et al. \citep{Zhang20233DCBAMs}).}
    \label{fig:Zhang20233DCBAMs1}
\end{figure}

Building upon these foundational single-image frameworks, subsequent research has integrated attention mechanisms to enhance feature extraction capabilities. Attention-based architectures incorporating Convolutional Block Attention Modules (CBAMs) have emerged as effective approaches for selectively emphasizing critical microstructural features during reconstruction. Zhang et al. \citep{Zhang20233DCBAMs} developed SASGAN (Super-Resolution Attention based Single-image GAN), which combines three main components: SinGAN (Single-Image GAN), ResNets (Residual Networks), and CBAMs (Convolutional Block Attention Modules). The dual attention mechanisms (channel and spatial attention) and residual connections enable stable training while focusing on critical microstructural features. As demonstrated in Figure \ref{fig:Zhang20233DCBAMs1}, SASGAN achieved superior accuracy in preserving pore space characteristics for sandstone samples, with porosity distributions (total, connected, and isolated) closely matching ground truth values compared to alternative methods including SNESIM, SinGAN, and ConSinGAN.

Adopting a different strategic approach, Zhang et al. \citep{Zhang2023StochasticMechanisms} introduced SMAGAN (Stochastic Multi-scale Attention GAN) for shale reconstruction from single training images. Unlike uniform attention deployment, SMAGAN employed selective CBAM integration across five generator stages, strategically placing attention modules at finer scales (g2-g5) while excluding the coarsest scale where global structure dominates. This selective attention mechanism improved feature diversity while maintaining computational efficiency, demonstrating high performance in preserving anisotropic properties characteristic of shale materials.

Addressing the challenge of maintaining reconstruction diversity while preserving feature quality in single-image frameworks, Zhang et al. (2023) \cite{Zhang2023StochasticNetwork} proposed MD2GAN (Multiple-stage Dual Discriminator GAN) for stochastic reconstruction of shale porous media. While sharing architectural similarities with SMAGAN through multi-stage CBAM integration and parameter sharing strategies, MD2GAN introduces a dual discriminator architecture where discriminator $D$ preferentially evaluates real data while discriminator $D'$ favors generated samples, with their complementary objectives balancing KL and reverse KL divergences to effectively mitigate mode collapse. The architecture progresses through six stages (16$^3$ to 64$^3$ voxels) with parameter reuse between consecutive stages, reducing first reconstruction time from approximately 40,000 seconds (SinGAN) to 18,679 seconds. The method demonstrated superior performance compared to SNESIM, IQ, and original SinGAN across pore space characteristics, multi-point connectivity, and permeability matching, while maintaining sufficient diversity in generated structures.

Concurrent multi-scale training frameworks incorporating residual architectures have demonstrated advantages in computational efficiency and reconstruction quality. Shi et al. \citep{Shi20233D-porous-GAN:Image} developed 3D-porous-GAN, which integrated residual connections between progressive stages ($16^3$ to $80^3$ voxels) with gradient penalty stabilization, achieving reduced 
training time while preserving pore connectivity across scales. Zhang et al. \citep{Zhang2023Super-ResolutionBlocks} further advanced this paradigm through a six-scale concurrent framework specifically designed for super-resolution reconstruction, where residual blocks at each scale facilitated gradient flow 
and enabled consistent porosity distribution and permeability matching. Both implementations demonstrate that residual architectures combined with concurrent training effectively balance computational efficiency with the preservation of critical microstructural properties.

While attention mechanisms and residual connections have enhanced single-image frameworks, alternative architectural approaches have emerged that maintain multi-scale learning through different generator-discriminator configurations. Zhao et al. \cite{Zhao20253DNetworks} introduced CEM3DMG for reconstructing complex multi-phase cement microstructures from a single 2D backscattered electron image. Unlike conventional SinGAN architectures employing multiple generator-discriminator pairs, CEM3DMG implements a single fully convolutional 3D generator paired with multiple 2D discriminators operating at different scales (200$\times$200 to 800$\times$800 pixels$^2$). The multi-scale learning strategy crops patches of varying sizes from the input image and rescales them to uniform resolution, while discriminators evaluate orthogonal 2D slices from the generated 3D volume to establish cross-dimensional correlation. The architecture employs WGAN-GP loss combined with a perceptual loss term and incorporates a block-wise synthesis strategy enabling generation of large-scale volumes. The method successfully captured complex cement microstructure characteristics including pore distribution, particle morphology, and multi-phase interactions, demonstrating that single-image training can be effectively achieved through architectural variations beyond the traditional multi-generator paradigm.


\subsection{Current Limitations and Future Directions}

Scale transition quality remains a central challenge specific to multi-scale architectures. Maintaining feature continuity between scales presents difficulties, particularly for materials with sharp transitions between pore sizes. Zhang T. et al. \citep{Zhang2023StochasticNetwork} addressed this through parameter sharing between consecutive stages, yet the challenge of seamless scale integration persists. The IPWGAN framework by Zhu et al. \citep{Zhu2024GenerationNetworks}, despite achieving large-scale generation up to $2{,}200^3$ voxels with 79\% reduction in permeability prediction errors for Berea sandstone, explicitly noted limitations in capturing features beyond the micrometer scale, stating that their method cannot adequately represent ``features such as fractures and laminations'' that exist at larger scales.

Single-image training approaches, while reducing data requirements, face inherent limitations in capturing material variability. Zhang T. et al. \citep{Zhang20233DCBAMs} and Zhang T. et al. \citep{Zhang2023StochasticMechanisms} demonstrated successful reconstruction from single training images, but this approach inherently limits the statistical diversity of generated structures to variations of the single known sample, and may not capture the full variability present in heterogeneous materials.

Memory constraints significantly impact the scalability of multi-scale approaches. While progressive growing enables generation at higher resolutions, the memory requirements for storing intermediate representations across multiple scales create practical limitations. Zhao et al. \citep{Zhao20253DNetworks} addressed this through block-wise synthesis strategies for their CEM3DMG framework, yet computational resources remain a limiting factor for desktop-scale deployment.

Several architecture-specific future directions emerge from the reviewed implementations. Zhu et al. \citep{Zhu2024GenerationNetworks} suggested that combining IPWGAN with complementary techniques and feature statistics mixing regularisation (FSMR) could address the limitation of capturing larger-scale features, beyond the micrometer range, while improving generation diversity. The success of concurrent training, as demonstrated by Zhang T. et al. \citep{Zhang2023StochasticMechanisms} with one-stage SMAGAN training reducing memory usage to 29\% RAM compared to 45\% for self-attention architectures, indicates that strategic architectural choices can significantly improve  computational efficiency without sacrificing quality. Integration of attention mechanisms with multi-scale frameworks also shows promise; Zhang T. et al. \citep{Zhang20233DCBAMs} combined CBAM modules with the SinGAN framework, achieving 2\% porosity accuracy with 1-second subsequent reconstructions. This suggests that selective feature enhancement through attention could address scale transition challenges while maintaining reasonable computational requirements. Additionally, the development of adaptive scale selection mechanisms, rather than fixed resolution hierarchies, could enable architectures to dynamically determine optimal scale transitions based on material characteristics \citep{Li2023DigitalNetwork}.

\pagebreak

\section{Conditional GAN}

Conditional Generative Adversarial Networks (cGANs) represent a significant advancement in controlled material reconstruction by incorporating additional information to guide the generation process \citep{Mirza2014ConditionalNets}. Unlike traditional GANs, which generate samples from random noise vectors, cGANs produce outputs conditioned on additional information such as class labels, property values, or other auxiliary variables. This conditioning mechanism provides precise control over crucial characteristics during the generation process, making cGANs particularly valuable for applications requiring specific feature control, such as microstructure generation with targeted properties.

The mathematical formulation of cGANs extends the traditional GAN framework by introducing a conditional variable $y$ into both the generator and discriminator:
\begin{equation}
\min_G \max_D V(D,G) = \mathbb{E}_{x \sim p_{data}(x)}[\log D(x|y)] + \mathbb{E}_{z \sim p_z(z)}[\log(1-D(G(z|y)))]
\end{equation}
where $y$ represents the conditioning information that can be properties like porosity, phase fractions, or processing parameters. This formulation enables the network to learn the conditional distribution $p(x|y)$, allowing users to specify desired characteristics in the generated samples.

The earliest implementations focused on controlling basic material properties. Mirza and Osindero \citep{Mirza2014ConditionalNets} introduced cGANs by conditioning both generator and discriminator on auxiliary information, enabling deliberate control over generation characteristics like digit class labels for MNIST generation and image features for automated tag generation. In material science, this approach was adapted to control fundamental properties like phase distribution and porosity in porous media. The development of cGANs for porous media reconstruction can be categorized into three primary approaches based on their conditioning strategies: single-property conditioning systems that target individual material characteristics through either processing parameters or intrinsic properties, multi-property control frameworks that manage multiple interdependent properties through optimization or direct conditioning approaches, and statistical or structural conditioning methods that leverage extracted features or enable reconstruction from limited structural information across dimensions and scales.

\subsection{Single-Property Conditioning}

Single-property conditioning represents the most straightforward application of cGANs in porous media, where implementations focus on controlling individual material properties while demonstrating the feasibility and accuracy of property-specific generation. The development of these approaches has progressed along two parallel tracks, conditioning on processing parameters for manufacturing applications, and conditioning on intrinsic material properties for microstructure generation.

Tang et al. \citep{Tang2021MachineAlumina} pioneered the application of cGANs to manufacturing process control, developing a regression-based conditional Wasserstein GAN with gradient penalty (RCWGAN-GP) for predicting microstructures during laser sintering of alumina. Their approach treats laser power as a continuous conditioning variable, enabling the prediction of microstructures under unexplored processing parameters through interpolation between trained conditions. The framework was validated using experimental SEM micrographs from various laser powers and demonstrated accurate regeneration of microstructural features including grain morphology, pore distribution, and secondary phase fractions. The regression-based conditioning mechanism allows the model to learn the continuous relationship between processing parameters and resulting microstructures, representing a significant advancement for process parameter optimization in advanced manufacturing.

Extending conditioning beyond single scalar properties, Matsuda et al. \citep{Matsuda2022FrameworkNetwork} introduced a cGAN approach for controlled structural hybridization of porous materials. Their system conditions on a 3-component material type intensity vector defining the degree of random, packed-sphere, or sponge structural characteristics, generating realistic porous structures corresponding to any point in this continuous composition space. Unlike direct property control approaches such as Tang et al.'s processing parameter conditioning, performance characteristics (pressure drop and filtration efficiency) are evaluated post-generation using computational fluid dynamics simulations. Coupled with multiobjective Bayesian optimization, the framework discovered hybrid materials exhibiting Pareto-optimal performance tradeoffs, demonstrating the effectiveness of parametric structural control for computational materials discovery.

\begin{figure}[H]
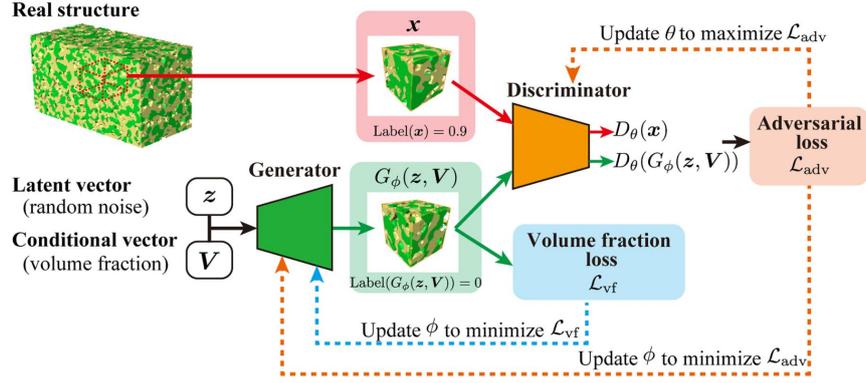

    \centering
    \includecroppedfigure[width=0.9\textwidth]
    \caption{Architecture of Kishimoto et al.'s cGAN for three-phase microstructure generation. The generator $G_{\phi}(z,V)$ accepts a latent vector $z$ (random noise) and conditional vector $V$ (target volume fraction) as inputs. Training employs dual optimization, the discriminator $D_{\theta}$ is trained to distinguish real from generated structures via adversarial loss $\mathcal{L}_{\text{adv}}$, while the generator minimizes both $\mathcal{L}_{\text{adv}}$ and volume fraction loss $\mathcal{L}_{\text{vf}}$ to produce realistic Ni-YSZ anode structures with precise compositional control (Figure adapted from Kishimoto et al. \citep{Kishimoto2023ConditionalFractions}, with permission)}
    \label{fig:kishimoto1}
\end{figure}

\begin{figure}[H]
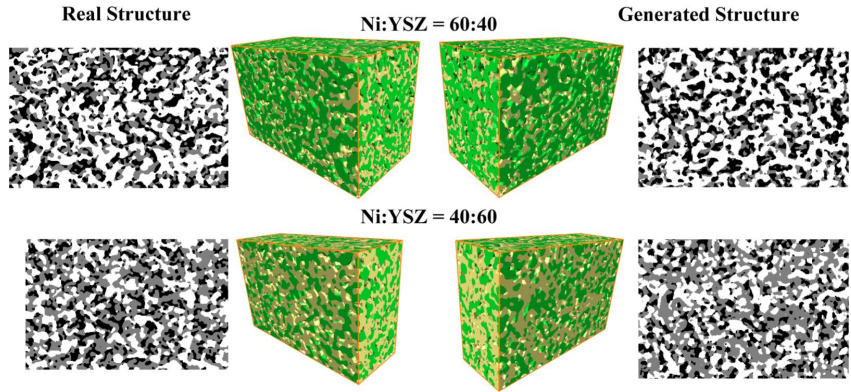

    \centering
    \includecroppedfigure[width=0.9\textwidth]
    \caption{Comparison of real and generated Ni-YSZ anode structures with different volume fractions (60:40 and 40:60), demonstrating the model's ability to generate realistic microstructures for compositions not included in the training data (Figure adopted from Kishimoto et al. \citep{Kishimoto2023ConditionalFractions}, with permission)}
    \label{fig:kishimoto2}
\end{figure}

Shifting toward direct control of intrinsic material properties, Kishimoto et al. \citep{Kishimoto2023ConditionalFractions} developed a cGAN for generating 3D porous structures of solid oxide fuel cell (SOFC) anodes with controllable volume fractions. Their approach combines a conventional GAN with an additional training process to control statistical parameters (Fig. \ref{fig:kishimoto1}). The generator processes both random latent vectors and conditional vectors specifying desired phase fractions, while the loss function integrates adversarial and volume fraction terms. This dual training strategy enables generation of realistic microstructures with precise phase distributions, even for compositions outside the training data, as demonstrated by Ni-YSZ structures with different volume fractions (Fig. \ref{fig:kishimoto2}). The balance between adversarial and volume fraction losses influences both the accuracy of property control and the diversity of generated structures, highlighting the importance of loss function design in conditional generation.

\begin{figure}[H]
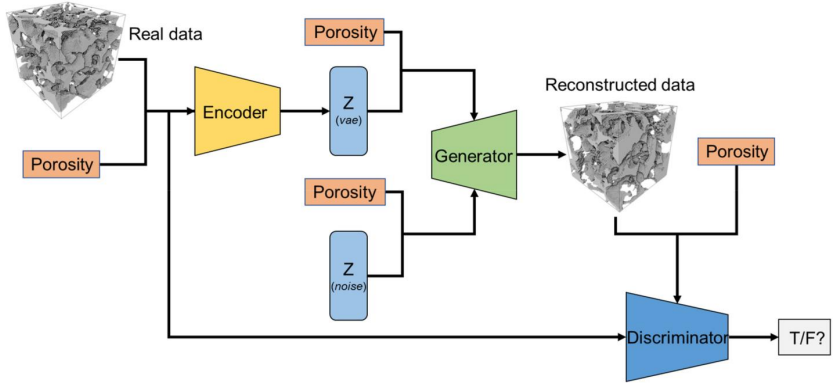

    \centering
    \includecroppedfigure[width=0.9\textwidth]
    \caption{Framework of the CVAE-GAN architecture for digital rock reconstruction with porosity control. The network combines a conditional variational autoencoder with a GAN structure, incorporating porosity conditions at multiple stages to enable controlled generation. (Figure adapted from Chi et al. \citep{Chi2023ReconstructionCVAE-GAN}, with permission)}
    \label{fig:chi1}
\end{figure}

\begin{figure}[H]
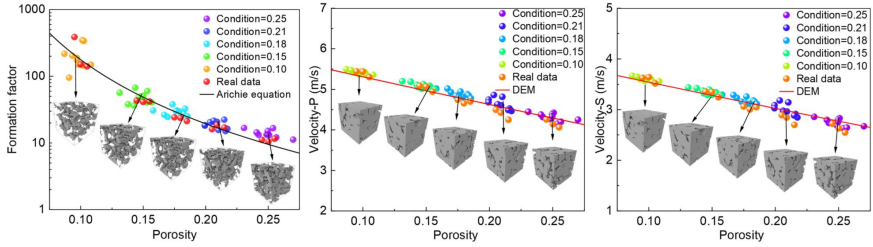

    \centering
    \includecroppedfigure[width=0.9\textwidth]
    \caption{Physical property validation of the CVAE-GAN reconstructed samples showing formation factor, velocity-P, and velocity-S measurements across different porosity conditions, demonstrating consistency with real sample properties. (Figure adapted from Chi et al. \citep{Chi2023ReconstructionCVAE-GAN}, with permission)}
    \label{fig:chi2}
\end{figure}

Building upon these direct property conditioning approaches, Chi et al. \citep{Chi2023ReconstructionCVAE-GAN} proposed a hybrid CVAE-GAN (conditional variational autoencoder generative adversarial network) approach for digital rock reconstruction with controlled porosity. Their framework incorporates porosity information throughout the network architecture, the encoder processes real samples with porosity labels, the generator combines latent vectors with porosity constraints, and the discriminator evaluates both real and synthetic data against porosity conditions (Fig. \ref{fig:chi1}). This comprehensive conditioning enables generation of samples with user-specified porosity values while maintaining realistic pore structures. Validation experiments demonstrated consistent physical properties including formation factor and acoustic velocities across different porosity levels (Fig. \ref{fig:chi2}).

Extending continuous conditioning to asphalt microstructures, Torben et al. \citep{Torben2025ControllableNetworks} developed a Continuous Conditional GAN (CCGAN) framework accommodating continuous regression labels for porosity control. Their architecture builds upon Self-Attention GAN with multi-phase segmentation capabilities. For 3D volume generation from 2D trained models, the framework employs latent space interpolation optimized through correlation analysis. Testing on porous asphalt micro-CT images demonstrated generation of realistic 2D slices assembled into cohesive 3D volumes. The method enables user-defined vertical porosity profile generation, though generated volumes exhibited narrower property distributions and slight regression toward mean porosity values at distribution extremes, attributed to limited representation of extreme porosity values in the training dataset.

\subsection{Multi-Property Control Systems}

As the field matured, researchers developed systems capable of simultaneously controlling multiple material properties, representing a significant advancement in controlled generation capabilities. The evolution of multi-property control frameworks reveals two distinct paradigms, namely optimization-based approaches that iteratively tune parameters toward multiple targets, and direct conditioning methods that incorporate multiple property values as explicit network inputs.

Nguyen et al. \citep{Nguyen2022SynthesizingLearning} pioneered an alternative paradigm for multi-property control, developing a GAN with Actor-Critic reinforcement learning (GAN-AC) framework that achieves control through optimization rather than direct conditioning. Unlike traditional cGANs where property values serve as generator inputs, their approach employs an actor-critic reinforcement learning system to iteratively tune the GAN's morphology parameters toward target properties. The actor network adjusts global morphology parameters controlling structural characteristics, while the critic evaluates generated structures against multiple target quantities of interest. The reward function optimizes across eight properties simultaneously, including porosity, specific surface area, permeability, and topological metrics (average clustering, graph density, degree assortativity, local efficiency, and graph transitivity). This inverse design approach demonstrated property control within 5\% error margins, representing a significant advancement toward materials-by-design methodologies through iterative refinement of complex property combinations.

Transitioning from optimization-based to direct conditioning approaches, the challenge of controlling multiple correlated properties became a central focus. Zheng and Zhang \citep{Zheng2022DigitalNetworks} introduced a conditional framework combining cGAN with progressive growing architecture (ProGAN) for digital rock reconstruction. Their system demonstrates separate control over rock type classification, porosity, or correlation length (both isotropic and anisotropic), while leveraging progressive growing for improved stability and quality. Due to the inherent correlation between properties in binary microstructures, the authors intentionally tested these conditions independently rather than simultaneously, revealing the fundamental challenge of controlling properties that exhibit natural interdependencies. Addressing this limitation, Zhou and Wu \citep{Zhou20233DLearning} introduced a multi-conditional generative adversarial network (MCGAN) for 3D reconstruction of digital rocks guided by petrophysical parameters. Their system incorporates multiple rock properties as conditional inputs to both generator and discriminator, enabling control over porosity, specific surface area, fractal dimension, and tortuosity through a switch structure that allows selective parameter enabling. The loss function includes a feature loss term that accelerates convergence and allows users to choose parameter combinations requiring tighter constraints.

Extending multi-property control to incorporate geological context, Sadeghkhani et al. \citep{Sadeghkhani2025PCP-GAN:Networks} introduced PCP-GAN, a multi-conditional framework conditioning on porosity and sample depth. Trained on blue-epoxy impregnated RGB thin-section samples from multiple carbonate depths, the framework preserves petrographic information through color-coded imagery, distinguishing mineralogical and porosity characteristics. The dual conditioning learns both universal pore network principles and depth-specific geological characteristics. An integrated enhanced U-Net quantifies porosity from images as conditioning inputs. The model achieved high porosity control ($R^2 = 0.95$) with mean absolute errors of 0.001--0.02, and morphological validation confirmed preservation of key pore network characteristics. This uniquely conditions on both material properties and geological context, enabling generation of formation-specific microstructures capturing depth-dependent diagenetic characteristics.

\subsection{Statistical and Structural Conditioning}

A distinct approach emerged focusing on incorporating statistical information and structural features as conditioning inputs, enabling reconstruction from limited data while maintaining statistical fidelity. This paradigm shift from property-based to feature-based conditioning addressed challenges where direct property specification is insufficient or where reconstruction must proceed from incomplete structural information.

Feng et al. developed a series of approaches progressively addressing challenges in statistical conditioning for porous media reconstruction. In their initial work \citep{Feng2018AcceleratingLearning}, they pioneered the application of deep learning to accelerate multi-point statistics (MPS) reconstruction methods. Their conditional GAN framework models the relation between sampling images containing conditioning data and target void-solid images, enabling acceleration of traditional MPS-based reconstruction. Unlike conventional MPS methods that perform point-by-point simulation through searching and matching, their cGAN-based method directly translates sampling images to target structures. This approach achieved significant speedup factors of approximately 760× for 2D and 25× for 3D reconstruction compared to traditional MPS methods. Building upon this foundation, their subsequent work \citep{Feng2019ReconstructionNetworks} extended statistical conditioning to reconstructing complete porous media from extremely limited information (Fig. \ref{fig:Feng}). Their approach uses partial images as small as 26×26 pixels from a 128×128 target as conditional input. The framework integrates four loss components: adversarial loss, L1 loss for pixel-level accuracy, pattern loss for capturing complex pore structure patterns, and porosity loss. The method incorporates Gaussian noise to enable multiple plausible reconstructions. These complementary works established statistical conditioning as a viable pathway when traditional property-based conditioning proves insufficient.

\begin{figure}[H]
    \centering
    \includecroppedfigure[width=0.9\textwidth]
    \caption{Network architecture for conditional reconstruction of porous media from partial images: (a) Generator network architecture showing convolutional layers (blue arrows), transposed convolutional layers (yellow arrows), and skip connections (orange arrows) that processes the partial image input to reconstruct the complete structure. (b) Discriminator architecture that evaluates generated samples against real data by processing both the conditional input and the generated/real output through convolutional layers with LeakyReLU activation and instance normalization (blue arrows), followed by a Sigmoid activation function (purple arrow).  (Figure adapted from Feng et al.\citep{Feng2019ReconstructionNetworks}, with permission).}
    \label{fig:Feng}
\end{figure}

\begin{figure}[H]
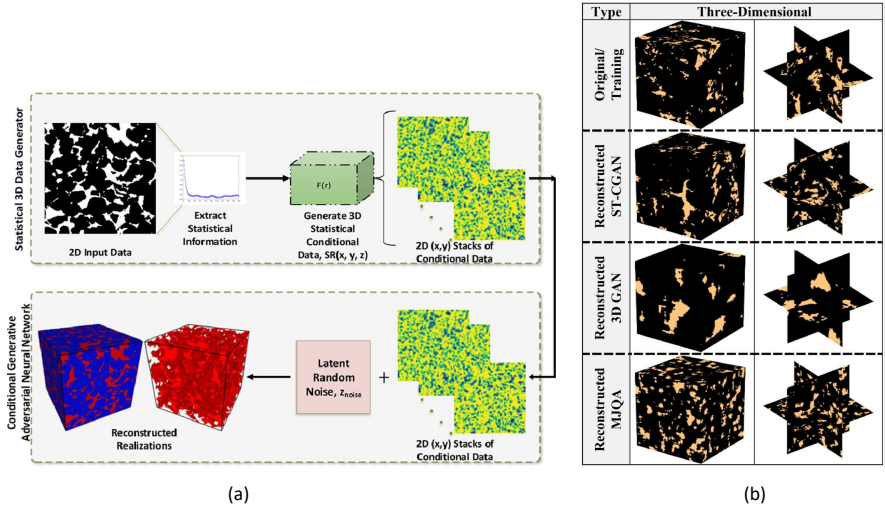

    \centering
    \includecroppedfigure[width=0.9\textwidth]
    \caption{Overview of the ST-CGAN framework for 3D porous media reconstruction: (a) Architecture showing the statistical data generator that extracts information from 2D input data to create conditional data, and the conditional generative adversarial neural network that combines this with random noise to produce 3D reconstructions. (b) Comparison between original/training samples and reconstructed outputs using different methods (ST-CGAN, 3D GAN, and MJQA), demonstrating ST-CGAN's high ability to preserve structural features (Figure adapted from Shams et al. \citep{Shams2021AST-CGAN}, with permission).}
    \label{fig:shams}
\end{figure}

While statistical conditioning addressed reconstruction from limited information, another challenge emerged in transforming structural information across dimensions and scales. Shams et al. \citep{Shams2021AST-CGAN} introduced ST-CGAN, a hybrid approach combining statistical methods with cGANs for 3D porous media reconstruction from 2D images (Figure \ref{fig:shams}). The statistical component extracts conditional information from 2D input data, which is then provided to the cGAN. The U-Net generator receives this statistical conditional data alongside random noise to guide 3D reconstruction, while the discriminator evaluates both the generated 3D image and 2D conditional data for realism and consistency. The method demonstrated approximately 1000-fold acceleration compared to traditional statistical methods while maintaining superior morphological and physical characteristic matching. Extending this dimensional transformation concept to address scale-dependent heterogeneity, Yang et al. \citep{Yang2022Multi-scaleNetworks} developed a multi-scale reconstruction method using low-resolution images as conditional inputs. Their U-Net generator with skip connections preserves macro-pore structures from low-resolution inputs while adding micro-pore details, enabling reconstruction that captures both small-scale and large-scale features simultaneously. Together, these works established that conditional GANs could effectively transform structural information across both dimensions (2D to 3D) and scales (low-resolution to multi-scale), addressing fundamental challenges where complete high-resolution 3D data is prohibitively expensive or technically infeasible to obtain.

\subsection{Current Limitations and Future Directions}

The simultaneous control of multiple strongly correlated properties remains the central challenge specific to conditional architectures. Zheng and Zhang \citep{Zheng2022DigitalNetworks} explicitly noted that properties in binary microstructures often exhibit inherent correlations that make independent control difficult, leading them to test porosity, rock type, and correlation length conditions separately. While Zhou and Wu \citep{Zhou20233DLearning} addressed this through their MCGAN framework with switchable conditioning, the fundamental challenge of navigating complex interdependencies between porosity, permeability, and structural characteristics persists.

Ensuring physical realizability in conditioned generation presents another limitation specific to this paradigm. Current conditional GANs may generate structures that satisfy mathematical conditioning constraints but violate physical principles; for instance, certain combinations of porosity, tortuosity, and specific surface area might be mathematically achievable but physically impossible. Chi et al. \citep{Chi2023ReconstructionCVAE-GAN} addressed this partially through their CVAE-GAN framework that maintains consistent formation factors and acoustic velocities across porosity conditions, yet the broader challenge of enforcing comprehensive physical constraints remains.

Extrapolation beyond training data distributions is a persistent limitation. While Kishimoto et al. \citep{Kishimoto2023ConditionalFractions} demonstrated successful generation for volume fractions not included in training data, and Tang et al. \citep{Tang2021MachineAlumina} showed interpolation capabilities for unexplored processing parameters, the reliability of these extrapolations typically degrades for properties far from training ranges. Relatedly, learning robust conditioning relationships requires diverse training data, yet Feng et al. \citep{Feng2019ReconstructionNetworks} noted that 600--1,000 images per category were needed despite achieving reconstruction from inputs as small as $26 \times 26$ pixels.

Architecture-specific future directions include extending conditioning beyond scalar properties to more complex constraints such as anisotropic property distributions and spatial property gradients. Yang et al. \citep{Yang2022Multi-scaleNetworks} demonstrated multi-scale reconstruction from single low-resolution inputs, suggesting that hierarchical conditioning strategies could enable more sophisticated control mechanisms. The successful demonstration of continuous conditioning by Tang et al. \citep{Tang2021MachineAlumina} through regression-based approaches indicates potential for flexible paradigms incorporating continuous manifolds of properties rather than discrete values. Additionally, statistical conditioning approaches such as that of Shams et al. \citep{Shams2021AST-CGAN}, which achieved 1000-fold acceleration compared to traditional statistical methods, suggest that hybrid statistical-generative frameworks merit further development.

\pagebreak

\section{Attention-Enhanced GAN}
\label{sec:attention-gan}

Attention mechanisms, initially introduced in transformer architectures for natural language processing, have emerged as a powerful enhancement to GANs for image reconstruction \citep{Zhang2018Self-AttentionNetworks, Zhang20233DCBAMs, He2023DigitalNetworks}. These mechanisms address fundamental limitations of traditional convolutional neural networks by enabling capture of long-range dependencies and global contextual information across image features, allowing generators to coordinate fine details across entire structures while discriminators verify their consistency \citep{Wang20233DStructure}. The integration of attention mechanisms into GANs has led to significant improvements in reconstructed material microstructure quality and accuracy, particularly for heterogeneous porous media with complex multiscale features \citep{Chi2024MultiscaleNetwork, Zhang2023StochasticMechanisms}.

The core principle centers on selectively weighting the importance of different spatial locations and feature channels during generation. Rather than treating all features equally as in standard convolutional operations, attention modules emphasize regions and channels contributing most significantly to realistic reconstruction while suppressing less relevant information. This selective focus proves particularly valuable where critical features such as pore connectivity, mineral boundaries, and multi-scale structural patterns require precise representation for accurate physical property prediction. Attention mechanisms in GANs for porous media reconstruction can be categorized into several distinct types, each addressing different aspects of feature learning and generation quality.

\subsection{Self-Attention and Context-Enhanced Mechanisms}

Self-attention mechanisms enable networks to capture long-range dependencies by computing relationships between all positions in the input, proving critical for capturing global structural coherence in porous media \citep{Wang20233DStructure, He2023DigitalNetworks}. The core operation involves transforming input features into Query (Q), Key (K), and Value (V) representations through learned linear transformations. For an input feature map $t$ at position $i$, the self-attention output is computed as:

\begin{equation}
\beta_{j,i} = \frac{\exp(q(t_i)^T k(t_j))}{\sum_j \exp(q(t_i)^T k(t_j))}
\end{equation}

\begin{equation}
t'_i = \gamma \sum_j (\beta_{j,i} \cdot v(t_j)) + t_i
\end{equation}

where $q(\cdot)$, $k(\cdot)$, and $v(\cdot)$ represent learned transformations (typically implemented as $1 \times 1$ convolutions), and $\gamma$ is a learnable scaling parameter often initialized to zero to allow the network to first rely on standard convolutional features before gradually incorporating attention \citep{He2023DigitalNetworks}. This formulation enables each position to aggregate information from all other positions weighted by their relevance, facilitating the capture of global structural patterns essential for maintaining pore network connectivity and geological consistency across large spatial extents.

\begin{figure}[H]
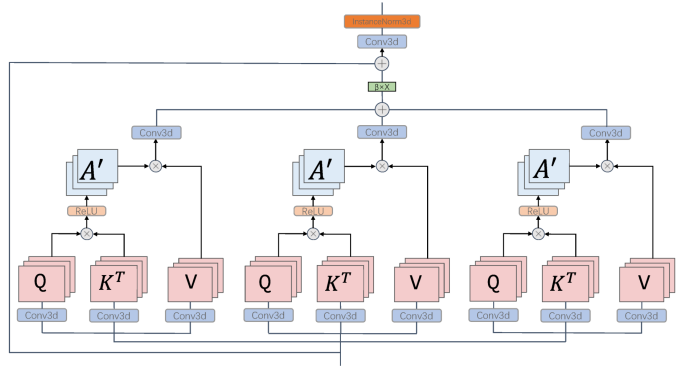

    \centering
    \includecroppedfigure[width=0.7\textwidth]
    \caption{ConvAttentionBlock architecture for hybrid self-attention mechanism. The block processes input through three parallel pathways generating Query (Q), Key ($K^T$), and Value (V) representations via Conv3d operations. Attention weights $A'$ are computed from Q and $K^T$ interactions, then applied to V through element-wise multiplication. The attention-refined features combine with the input via residual connections and element-wise addition before InstanceNorm3d normalization. The architecture shows three repeated attention modules demonstrating the hierarchical processing structure. (Figure adapted from Wang et al. \citep{Wang20233DStructure}).}
    \label{fig:convattention_block}
\end{figure}

Wang et al. \citep{Wang20233DStructure} developed a hybrid self-attention approach combining traditional and fully convolutional self-attention mechanisms specifically for carbonate rock reconstruction. The hybrid architecture addresses carbonate's complex heterogeneity and dissolution features through ConvAttentionBlocks that integrate query-key-value transformations with 3D convolutional operations. As shown in Figure \ref{fig:convattention_block}, the ConvAttentionBlock processes 3D input through parallel pathways for query (Q), key ($K^T$), and value (V) transformations, each consisting of Conv3d operations, ReLU activations, and element-wise operations. The attention mechanism, denoted as $A'$, computes relationships between all spatial positions before combining with the input through element-wise multiplication and residual connections, followed by InstanceNorm3d normalization. This combination enables the network to capture both long-range dependencies between distant features and spatial relationships through convolutional inductive biases, proving particularly effective for replicating carbonate dissolution phenomena and variable pore size distributions that challenge purely convolutional approaches.

He et al. \citep{He2023DigitalNetworks} introduced Residual Self-Attention Blocks (RSAB) that combine residual learning with self-attention for digital core image reconstruction. The architecture integrates self-attention modules with residual connections to capture both high-resolution details and long-distance dependencies while facilitating gradient flow during training. A key innovation involves using a learnable parameter $\gamma$ initialized to zero in the output combination $t' = \gamma \cdot o + t$, allowing gradual incorporation of attention-based refinements. The approach was rigorously evaluated using Fréchet Inception Distance (FID) and Kernel Inception Distance (KID), representing one of the first applications of these distribution-similarity metrics to porous media reconstruction, demonstrating strong statistical consistency between generated and real shale samples.

For materials with nano-micron scale features, context-enhanced attention mechanisms have been developed to explicitly incorporate local surrounding information alongside global context \citep{Pingquan2023ShaleEstimation}. Unlike standard self-attention that computes relationships independently at each spatial location, context-enhanced approaches explicitly model neighboring region relationships where local context significantly influences feature interpretation.

Wang et al. \citep{Pingquan2023ShaleEstimation} developed context-aware attention modules integrated with high-resolution optical flow estimation (COFRnet-3DWGAN) for nano-micron scale pore reconstruction in shale. The mechanism enhances feature extraction by considering both local details and global context, proving particularly valuable for nano-scale pores where traditional attention mechanisms struggle. This enables the network to distinguish between structurally similar but contextually different features, such as isolated pores versus connected pore networks, that require different treatments during reconstruction. The synergistic combination with optical flow estimation captures structural continuity between 2D slices, enabling 3D shale reconstruction with improved accuracy for nano-micron features that critically influence transport properties despite their small volumetric fractions.

\subsection{Channel and Spatial Attention (CBAM) Approaches}

\begin{figure}[H]
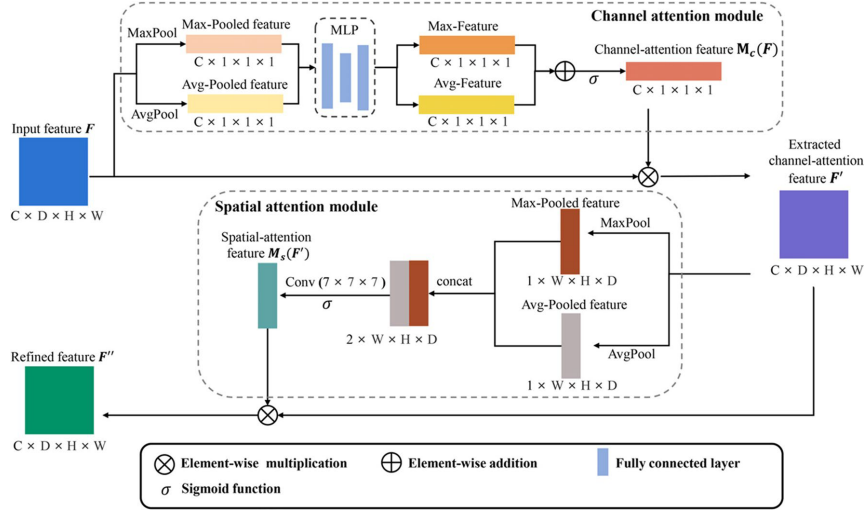

    \centering
    \includecroppedfigure[width=0.9\textwidth]
    \caption{Structure of the Convolutional Block Attention Module (CBAM) showing channel and spatial attention pathways. The channel attention module utilizes max and average pooling operations followed by MLP layers, while the spatial attention module employs convolution operations to generate attention maps. Element-wise multiplication and addition operations combine attention-refined features. (Figure adapted from Zhang et al. \citep{Zhang2023StochasticMechanisms}).}
    \label{fig:cbam_module}
\end{figure}

Channel and spatial attention mechanisms take a complementary approach to self-attention by focusing on ``what'' and ``where'' aspects of feature maps respectively. Channel attention identifies which feature channels contain the most relevant information, while spatial attention determines which spatial locations deserve emphasis \citep{Zhang20233DCBAMs, Zhang2023StochasticMechanisms}. The Convolutional Block Attention Module (CBAM), combining both attention types sequentially, has proven particularly effective for porous media applications. The channel attention module computes:

\begin{equation}
M_c(F) = \sigma(\text{MLP}(\text{AvgPool}(F)) + \text{MLP}(\text{MaxPool}(F)))
\end{equation}

where $\sigma$ denotes the sigmoid activation function, and both average and max pooling operations capture complementary channel-wise statistics. The refined features $F' = M_c(F) \otimes F$ then pass through spatial attention:

\begin{equation}
M_s(F') = \sigma(f^{7 \times 7}([\text{AvgPool}(F'); \text{MaxPool}(F')]))
\end{equation}

where $f^{7 \times 7}$ represents a $7 \times 7$ convolutional layer, and the final output becomes $F'' = M_s(F') \otimes F'$. This sequential refinement enables the network to first determine which feature channels to emphasize, then identify where within those channels to focus attention, creating a hierarchical feature selection process well-suited to the multi-scale nature of porous structures \citep{Zhang20233DCBAMs}. Figure \ref{fig:cbam_module} illustrates the CBAM architecture, showing how channel attention (utilizing both max and average pooling operations followed by MLP layers) and spatial attention (employing $7 \times 7 \times 7$ convolution operations) work sequentially to refine feature maps.

The application of channel and spatial attention mechanisms to porous media reconstruction began with Shan et al. \citep{Shan2022RockNetwork}, who developed the Residual Dual-Channel Attention Generative Adversarial Network (RDCA-SRGAN) for rock CT image super-resolution. The approach employs both global average pooling and global max pooling in parallel pathways to capture complementary channel-wise statistics, with average pooling providing smoothed summaries while max pooling emphasizes peak responses. The generator cascades 16 Residual Dual-Channel Attention Blocks (RDCAB) while the discriminator strategically places 2 RDCAB units to enhance feature discrimination. Cross-material validation across carbonate, sandstone, and coal demonstrated the approach's effectiveness at recovering high-frequency details and producing clear, sharp edges, particularly excelling in capturing subtle features such as coal cleats and dissolved carbonate regions.

Building upon these attention-based super-resolution capabilities, Zhang et al. \citep{Zhang20233DCBAMs} extended the approach to address limited training data scenarios by integrating CBAM with the Single-Image GAN (SinGAN) framework for 3D super-resolution reconstruction. The architecture combines residual networks and CBAM to learn structural characteristics from a single LR 3D training image, addressing both the degradation problem in deep networks and the high-resolution/large-field-of-view trade-off in physical imaging. Each residual block integrates CBAM that sequentially applies channel attention (determining which feature channels merit emphasis) followed by spatial attention (identifying where to focus within those channels).

\begin{figure}[H]
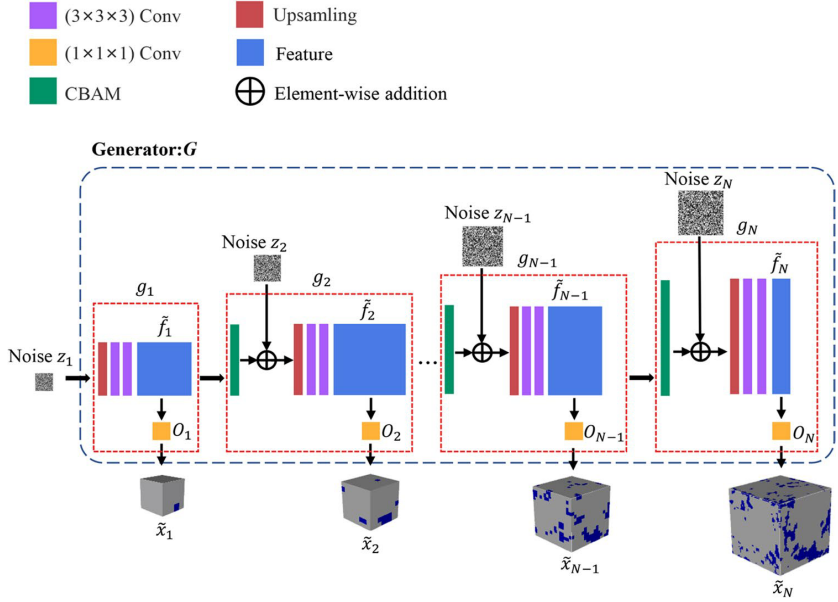

    \centering
    \includecroppedfigure[width=0.9\textwidth]
    \caption{Architecture of the SMAGAN generator showing progressive feature generation across multiple scales. The network processes input noise $z_1$ through $z_N$ at each stage, incorporating $(3 \times 3 \times 3)$ convolutions (purple), $(1 \times 1 \times 1)$ convolutions (orange), upsampling operations (red), feature blocks (blue), and CBAM modules (green) at each scale level. The generator progressively creates porous media structures ($\tilde{x}_1$ through $\tilde{x}_N$) at increasing resolutions, with CBAM enhancing feature extraction at finer scales to capture detailed structural characteristics while maintaining computational efficiency. Element-wise addition operations combine features across scales. (Figure adapted from Zhang et al. \citep{Zhang2023StochasticMechanisms}).}
    \label{fig:smagan_architecture}
\end{figure}

Further advancing single-image training approaches, Zhang et al. \citep{Zhang2023StochasticMechanisms} extended CBAM integration to concurrent multi-scale training through SMAGAN (Stochastic Multi-scale Attention GAN) for shale reconstruction. The architecture employs five generator stages processing progressively finer scales, with CBAM modules strategically integrated at finer scales (g2-g5) while excluding the coarsest scale where global structure dominates. As illustrated in Figure \ref{fig:smagan_architecture}, the SMAGAN generator processes input noise through multiple progressive stages, with each stage ($g_1$ through $g_N$) incorporating convolutional layers (shown in purple for $3 \times 3 \times 3$ convolutions and orange for $1 \times 1 \times 1$ convolutions), upsampling operations (red blocks), feature extraction (blue blocks), and CBAM modules (green blocks) integrated at stages $g_2$ through $g_N$. The architecture progressively generates porous media structures at increasing resolutions ($\tilde{x}_1$ through $\tilde{x}_N$), with noise injection at each scale enabling stochastic variation while CBAM modules enhance feature extraction by focusing attention on critical structural elements at each resolution level.

Unlike progressive growing approaches training scales sequentially, SMAGAN implements one-stage concurrent training where all scales train simultaneously, enabling parameter sharing and reducing training time while maintaining quality. The approach incorporates hinge-GAN loss to address gradient vanishing and diversity loss to prevent mode collapse, proving particularly valuable for single-image training scenarios.

\subsection{Multi-Mask Attention for Multiscale Image Fusion}

While the attention mechanisms discussed thus far focus on reconstruction enhancement from noise or resolution improvement, attention has also proven effective for multi-scale image fusion tasks where complementary imaging modalities are integrated. Multi-mask attention mechanisms employ multiple attention masks that can simultaneously focus on different structural components, unlike single-attention approaches (self-attention, CBAM) that produce one attention map \citep{Chi2024MultiscaleNetwork}. This approach generates multiple attention masks alongside content masks, enabling selective weighting of different features during fusion.

\begin{figure}[H]
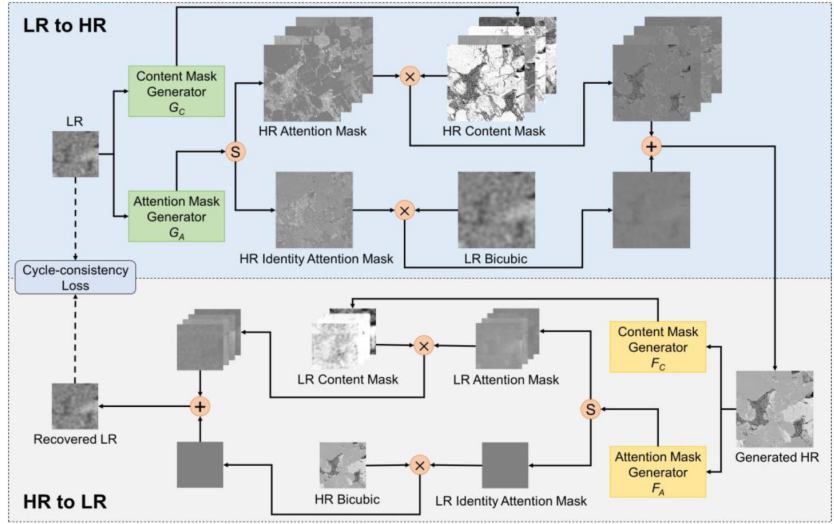

    \centering
    \includecroppedfigure[width=0.9\textwidth]
    \caption{Overall framework of AttentionGAN for multiscale fusion of tight sandstone digital rocks. The network consists of bidirectional transformations: LR to HR (top) with generators $G_C$ and $G_A$ producing content and attention masks, and HR to LR (bottom) with generators $F_C$ and $F_A$. Both pathways use cycle-consistency loss to maintain feature consistency. The architecture generates HR attention masks and HR content masks that guide the fusion process, enabling integration of micro-CT (LR) and SEM (HR) images while preserving multiscale structural features. (Figure adapted from Chi et al. \citep{Chi2024MultiscaleNetwork}, with permission).}
    \label{fig:attentiongan_architecture}
\end{figure}

Chi et al. \citep{Chi2024MultiscaleNetwork} introduced AttentionGAN for combining micro-CT and SEM images of tight sandstone, addressing the challenge that single-modality imaging cannot simultaneously achieve wide field of view and high resolution necessary for characterizing heterogeneous materials. As illustrated in Figure \ref{fig:attentiongan_architecture}, the framework employs bidirectional transformations with content mask generators ($G_C$, $F_C$) and attention mask generators ($G_A$, $F_A$) that produce masks guiding the fusion process. Both pathways utilize cycle-consistency loss to maintain feature consistency across scales. Attention masks focus on structurally important regions across scales, while content masks preserve fine-scale SEM features during fusion with coarser micro-CT data, balancing global structural coherence with detailed pore-scale features.

\begin{figure}[H]
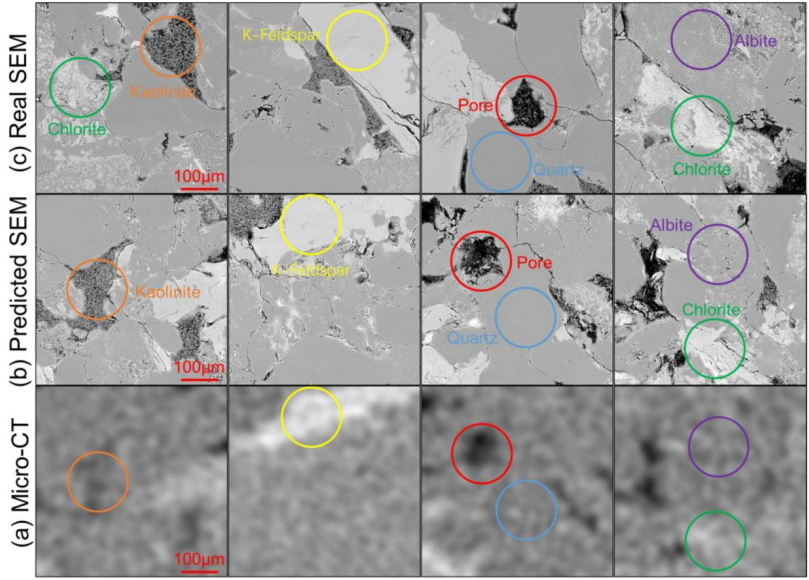

    \centering
    \includecroppedfigure[width=0.9\textwidth]
    \caption{Comparison between (a) micro-CT images (resolution: 8~$\mu$m), (b) predicted SEM images, and (c) real SEM images (resolution: 0.5~$\mu$m). Various mineral components are highlighted: kaolinite (orange), K-feldspar (yellow), pores (red), quartz (blue), albite (purple), and chlorite (green), demonstrating the network's ability to preserve mineral features across scales. The predicted images show strong consistency with real SEM images while maintaining correspondence with micro-CT structural features, validating AttentionGAN's effectiveness for multiscale fusion. (Figure adapted from Chi et al. \citep{Chi2024MultiscaleNetwork}, with permission).}
    \label{fig:attentiongan_results}
\end{figure}

Figure \ref{fig:attentiongan_results} demonstrates the effectiveness of AttentionGAN through comparison of micro-CT images at 8~$\mu$m resolution, predicted SEM images, and real SEM images at 0.5~$\mu$m resolution. The predicted images show high consistency with real SEM images in capturing mineral distributions and pore structures while maintaining correspondence with micro-CT structural features, validating the network's capability to preserve features across nearly two orders of magnitude in resolution. Comprehensive validation through two-point correlation functions, representative volume element analysis, pore radius distributions, and mineral content analysis confirmed that fused images maintain statistical similarity to source images at their respective scales while capturing the full spectrum from macro-scale intergranular pores to nano-scale micropores.

\subsection{Current Limitations and Future Directions}

The quadratic scaling of self-attention with input size represents the defining constraint of attention-enhanced architectures. Zhang T. et al. \citep{Zhang2023StochasticMechanisms} reported that self-attention mechanisms require 45\% RAM usage compared to 29\% for CBAM-based approaches in comparable architectures. For 3D volumetric data, attention maps grow cubically with spatial dimensions, and current implementations are effectively restricted to $80^3$ voxels for single-GPU configurations \citep{Zhang20233DCBAMs, Zhang2023StochasticMechanisms}. This limitation proves particularly acute for applications requiring representative volume elements of highly heterogeneous materials, making large-scale reconstruction computationally prohibitive with current attention formulations.

Training time requirements, while showing significant improvement in subsequent reconstructions, remain substantial for initial model training. Zhang T. et al. \citep{Zhang20233DCBAMs} reported approximately 6 hours for initial training of their SASGAN architecture while achieving 1-second subsequent reconstructions, and Zhang T. et al. \citep{Zhang2023StochasticMechanisms} documented initial training times of approximately 1.6 hours with 6-second subsequent generations. However, Chi et al. \citep{Chi2024MultiscaleNetwork} required approximately 12 hours for their AttentionGAN framework, demonstrating that more complex attention mechanisms demand considerably greater computational investment. This sharp contrast between initial training cost and subsequent generation speed is characteristic of attention-enhanced architectures.

The effectiveness of attention mechanisms depends critically on training data diversity. Single-image approaches such as SASGAN \citep{Zhang20233DCBAMs} and SMAGAN \citep{Zhang2023StochasticMechanisms} successfully operate with minimal data but inherently sacrifice generalisation capability, generating variations of a single known structure rather than diverse samples representing broader geological variability. He et al. \citep{He2023DigitalNetworks} addressed this partially through data augmentation, expanding 20,000 original samples to 100,000, yet the fundamental tension between data scarcity and generation diversity remains.

A further limitation specific to attention-enhanced architectures is their limited integration with conditional generation frameworks. Unlike conditional GANs that explicitly incorporate property constraints, attention mechanisms focus primarily on feature importance weighting rather than property targeting. This represents a significant gap for inverse design applications where specific material properties must be achieved, as the attention mechanisms excel at preserving structural coherence but lack direct mechanisms for property-controlled generation.

Architecture-specific future directions centre on addressing these constraints. Memory-efficient attention through sparse patterns that compute relationships only for relevant feature pairs, rather than full attention maps, could substantially reduce resource requirements. Zhang T. et al. \citep{Zhang2023StochasticMechanisms} demonstrated that strategic CBAM placement at finer scales while excluding coarse scales reduced memory usage from 45\% to 29\%, suggesting that selective attention deployment merits further investigation. The development of 3D-specific attention mechanisms also deserves particular focus, as current implementations largely adapt 2D attention concepts to volumetric data. Chi et al. \citep{Chi2024MultiscaleNetwork} demonstrated multi-mask attention for multiscale fusion of micro-CT and SEM images across nearly two orders of magnitude in resolution, indicating the potential for hierarchical 3D attention operating efficiently at multiple scales. 

Adaptive attention strategies show further promise; building on Wang B. et al.'s \citep{Wang20233DStructure} hybrid self-attention for carbonate rocks and Wang P. et al.'s \citep{Pingquan2023ShaleEstimation} context-aware modules for nano-micron scale features, future architectures could dynamically select attention types based on local material characteristics, applying self-attention for regions requiring long-range dependency modelling while using efficient channel attention for locally correlated features.

\pagebreak

\section{Style-based GAN}
\label{Style-based GAN}

Style-based GAN (StyleGAN), initially introduced by Karras et al. \citep{Karras2018ANetworks}, represents a significant advancement in generative adversarial network architecture that fundamentally changes how latent codes influence the generation process. Unlike traditional GANs that directly feed latent codes to the generator, StyleGAN employs a mapping network to transform input latent codes into an intermediate latent space, enabling hierarchical control over generated features through style injection at multiple network depths. This hierarchical control mechanism has proven particularly effective for microstructure reconstruction in porous materials, allowing independent manipulation of coarse structural attributes and fine-grained details \citep{Fokina2020MicrostructureNetworks, Cao2022ReconstructionGAN, Li2024Three-dimensionalNetwork}.

\begin{figure}[H]
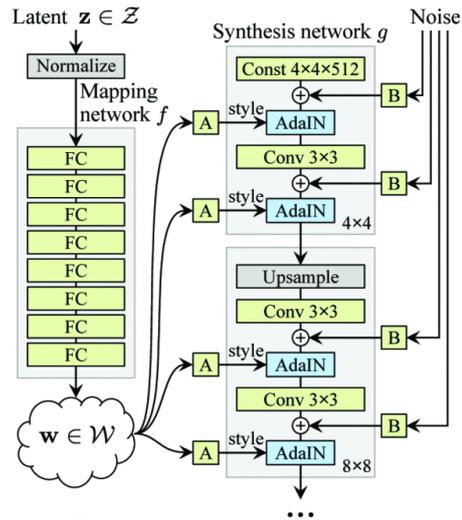

    \centering
    \includecroppedfigure[width=0.5\textwidth]
    \caption{Style-based GAN architecture. The mapping network transforms input latent code z into intermediate latent space w through 8 fully-connected layers. The synthesis network progressively generates images from 4×4 to higher resolutions, with style information injected at each scale via adaptive instance normalization (AdaIN) operations. Learned affine transformations (A) modulate styles, while per-channel scaling factors (B) apply noise for stochastic variation. This architecture enables hierarchical control over generated features at different scales. (Figure adapted from Karras et al. \citep{Karras2018ANetworks})}
    \label{fig:stylegan}
\end{figure}

\begin{figure}[H]
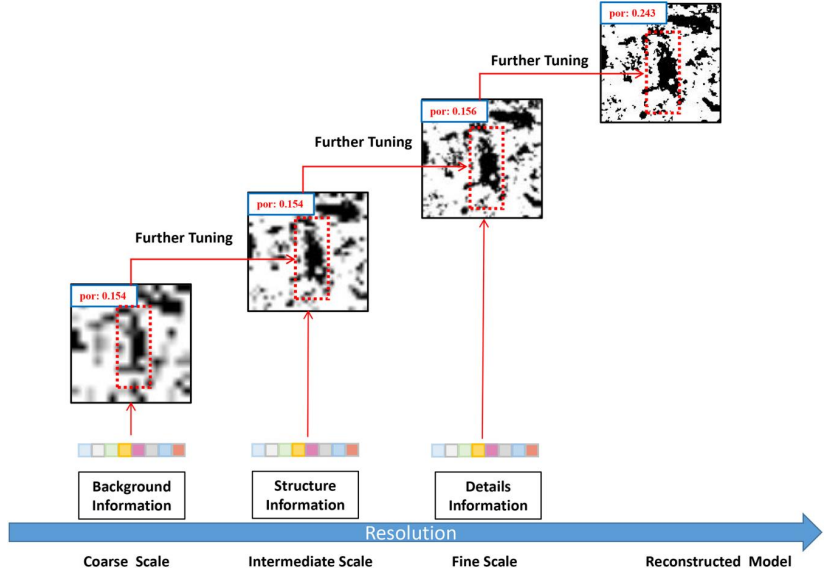

    \centering
    \includecroppedfigure[width=0.9\textwidth]
    \caption{Progressive multi-scale reconstruction in StyleGAN showing hierarchical feature control across different resolutions. Coarse scales define background pore distribution, intermediate scales enrich pore structure and boundaries, and fine scales add microscopic details. Porosity values (por.) indicate progressive refinement at each scale. (Figure adapted from Cao et al. \citep{Cao2022ReconstructionGAN}, with permission).}
    \label{fig:multiscale}
\end{figure}

The intermediate latent codes (w) are strategically injected at multiple points throughout the StyleGAN generator, creating a hierarchical control system over the generated features (Figure \ref{fig:stylegan}). When these style codes are introduced at earlier layers of the network, they govern coarse attributes of the generated output such as background pore distribution, while injection at deeper layers enables control over fine-grained details including pore boundaries and microscopic features (Figure \ref{fig:multiscale}). This multi-scale style injection mechanism aligns with the natural hierarchical organization of features in generative networks, where earlier layers typically handle broad structural elements while deeper layers refine local details \citep{Karras2018ANetworks, Cao2022ReconstructionGAN}.

The transformed latent code influences the generation process through adaptive instance normalization (AdaIN), which is incorporated after each convolutional layer. AdaIN applies style-based modulation using the transformed latent code according to the following operation \citep{Karras2018ANetworks}:
\begin{equation}
\text{AdaIN}(x_i, y) = y_{s,i}\left(\frac{x_i - \mu(x_i)}{\sigma(x_i)}\right) + y_{b,i}
\end{equation}
where $x_i$ represents feature maps, and $y$ contains style information derived from the intermediate latent space.

Furthermore, after each convolution layer, random noise is added to introduce stochastic variations, which helps capture the natural randomness present in microstructures \citep{Karras2018ANetworks}. Additionally, StyleGAN employs a progressive growing framework that gradually increases image resolution during training, ensuring stable high-resolution image generation. Overall, these innovations yield several practical advantages including enhanced control over specific image features, improved image quality and resolution, and a more structured latent space allowing independent attribute modification. Further improvements were introduced in StyleGAN2 \citep{Karras2019AnalyzingStyleGAN}, which addressed image artifacts and improved the quality of generated samples, particularly relevant for microstructure generation where unrealistic features could affect physical property predictions.

The application of StyleGAN architectures to porous media reconstruction has evolved significantly since 2020, with implementations progressing from initial demonstrations to sophisticated hybrid systems capable of multi-scale reconstruction and precise property control. As shown in Table \ref{tab:stylegans-variants}, this evolution demonstrates distinct architectural approaches, each addressing specific technical requirements in material generation and reconstruction.

\subsection{Core StyleGAN Architectures}

The application of StyleGAN to porous media reconstruction began with the original StyleGAN architecture and subsequently evolved through StyleGAN2-ADA, which incorporated adaptive augmentation mechanisms for improved training stability and performance with limited datasets.

\begin{figure}[H]
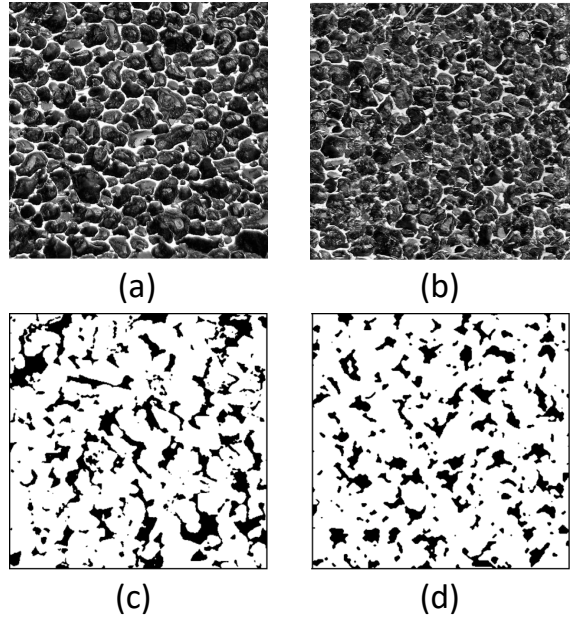

    \centering
    \includecroppedfigure[width=0.7\textwidth]
    \caption{Visual comparison of obtained results for Alporas aluminum foam: (a) original image of alporas, (b) result of StyleGAN; Berea sandstone: (c) original image, (d) result of StyleGAN+quilting. (Figure adapted from Fokina et al. \citep{Fokina2020MicrostructureNetworks}, with permission).}
    \label{fig:stylegan_Fokina}
\end{figure}

The original StyleGAN architecture, implemented by Fokina et al. (2020) \citep{Fokina2020MicrostructureNetworks}, establishes the foundational framework with an 8-layer mapping network and six connected blocks. Its structure features progressive resolution growth from 8×8 to 128×128 pixels and implements a dual-channel control system. The architecture employs Adaptive Instance Normalization (AdaIN) operations after each layer, combined with noise injection for local variation control. This implementation demonstrated effectiveness in microstructure synthesis across various porous media, including Alporas aluminum foam and Berea sandstone (Fig. \ref{fig:stylegan_Fokina}).

\begin{figure}[H]
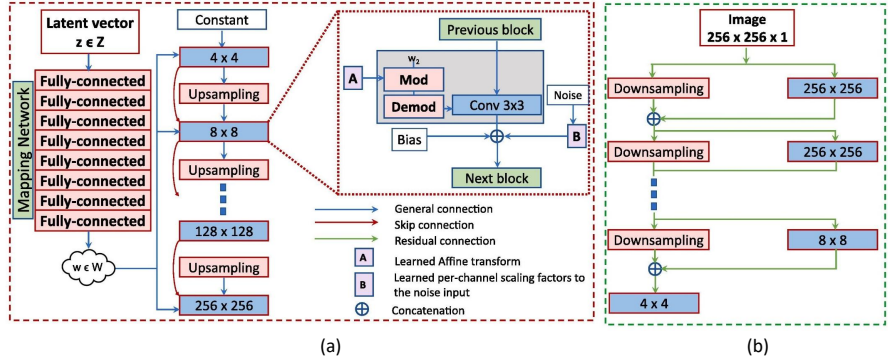

    \centering
    \includecroppedfigure[width=0.9\textwidth]
    \caption{Architectural overview of StyleGAN2-ADA generator. (a) The generator network starts with a latent vector z mapped through fully-connected layers to intermediate latent space w. The synthesis network begins with a constant input, progressively upsampling through multiple resolution levels (4×4 to 256×256) with modulation, demodulation, and noise injection at each scale. Style injection occurs through learned affine transformations (A), while noise is added with learned per-channel scaling factors (B). Skip, general, and residual connections are indicated by different colored arrows. (b) Discriminator network, employing a symmetric downsampling path from 256×256 to 4×4 resolution with residual connections and concatenation operations. (Figure adapted from Haribabu et al. \citep{Haribabu2023AData}, with permission)}
    \label{fig:stylegan2-ada}
\end{figure}

\begin{figure}[H]
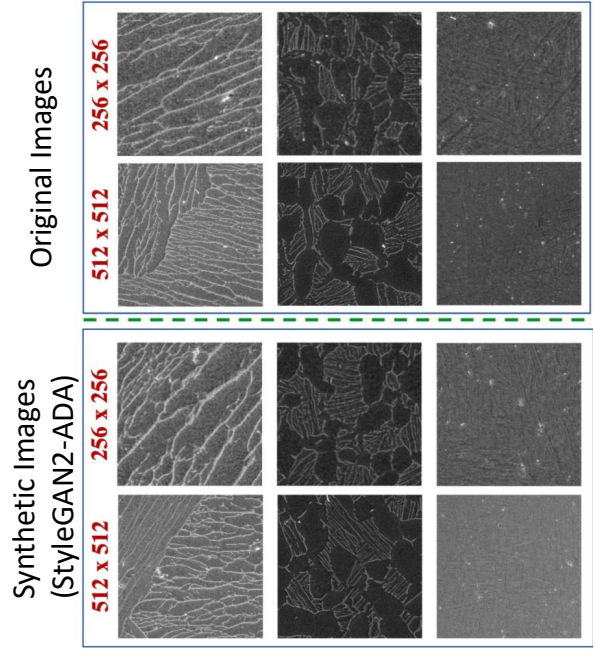

    \centering
    \includecroppedfigure[width=0.7\textwidth]
    \caption{Qualitative comparison between original titanium alloy microstructures (top) and synthetic images generated by StyleGAN2-ADA (bottom) at two different resolutions: 256×256 pixels and 512×512 pixels. Images showcase three distinct microstructural morphologies: bimodal (left), lamellar (middle), and acicular (right). The synthetic images demonstrate high fidelity to the original microstructural features across different scales. (Figure adapted from Haribabu et al. \citep{Haribabu2023AData}, with permission).}
    \label{fig:stylegan2-ada_haribu}
\end{figure}

Building upon this foundation, StyleGAN2-ADA introduced adaptive discriminator augmentation with dynamic training adjustment, as implemented by Thakre et al. (2023) \citep{Thakre2023QuantificationModels}, Haribabu et al. (2023) \citep{Haribabu2023AData}, and Huang et al. (2022) \citep{Huang2022Deep-learning-basedMethod}. As shown in Figure \ref{fig:stylegan2-ada}, the architecture begins with a mapping network of fully-connected layers that transforms the input latent vector z into an intermediate latent space w. The generator network employs a series of upsampling blocks with modulation, demodulation, and noise injection components. The constant input block is progressively upsampled through convolution operations from 4×4 to 256×256 resolution, with style injection occurring at each scale through learned affine transformations. Haribabu et al. \citep{Haribabu2023AData} demonstrated the effectiveness of this architecture by successfully generating synthetic titanium alloy microstructures at multiple resolutions (Fig. \ref{fig:stylegan2-ada_haribu}), achieving high-fidelity results particularly for lamellar and bimodal morphologies.

The architecture incorporates style demodulation and path length regularization, proving particularly effective for materials like dual-phase steels and titanium alloys, with phase fraction variations typically ranging from 1--5\% compared to original samples. The discriminator follows a symmetrical downsampling path, utilizing adaptive augmentation to maintain training stability despite limited training data \citep{Thakre2023QuantificationModels}.

\subsection{Hybrid StyleGAN Architectures}

Hybrid StyleGAN architectures extend the core framework by integrating complementary networks to address multi-scale reconstruction challenges, with different implementations targeting distinct reconstruction objectives.

\begin{figure}[H]
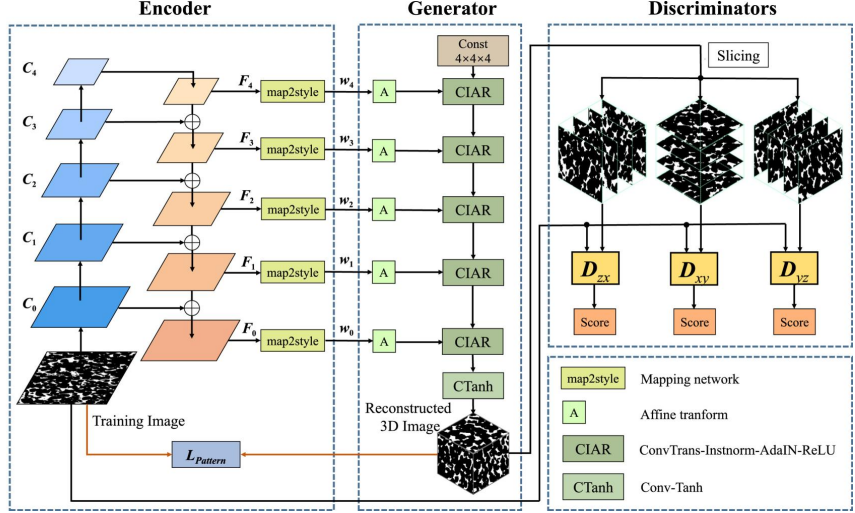

    \centering
    \includecroppedfigure[width=0.9\textwidth]
    \caption{Architecture of the EFPN-StyleGAN framework for 3D porous media reconstruction from 2D images. The encoder extracts multi-scale features (C0 to C4) using an extended feature pyramid network, which are transformed into style codes (w0 to w4) via map2style modules. The generator incorporates these features through CIAR (ConvTrans-InstNorm-AdaIN-ReLU) blocks to progressively construct the 3D structure. Three discriminators (Dzx, Dxy, Dyz) evaluate orthogonal cross-sections (xy, xz, yz planes) to ensure 3D consistency. (Figure adapted from Li et al. \citep{Li2024Three-dimensionalNetwork}, with permission).}
    \label{fig:Li2024Three-dimensionalNetwork}
\end{figure}

Li et al. (2024) \citep{Li2024Three-dimensionalNetwork} integrated an extended feature pyramid network (EFPN) with StyleGAN for effective feature extraction. As shown in Figure \ref{fig:Li2024Three-dimensionalNetwork}, their architecture consists of an encoder that extracts multi-grid image features through EFPN, a generator incorporating these features via style transfer, and three discriminators to evaluate orthogonal cross-sections. This design enables the effective capture of features at different resolutions and scales.

\begin{figure}[H]
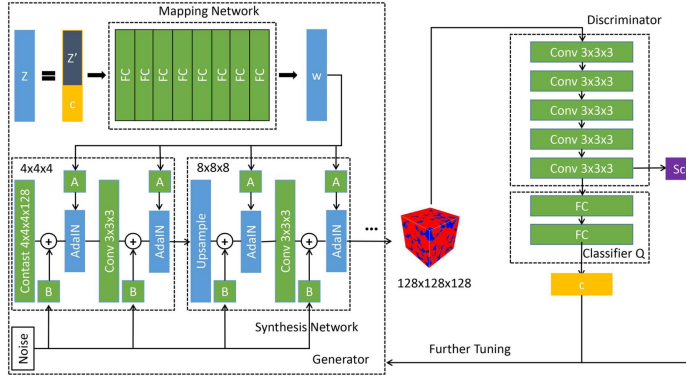

    \centering
    \includecroppedfigure[width=0.75\textwidth]
    \caption{The architecture of CISGAN (Combined InfoGAN and StyleGAN) for porous media reconstruction. The network consists of a mapping network that transforms latent code and prior information into an intermediate latent space w, a synthesis network that progressively generates the 3D structure through AdaIN operations and convolutions, and a discriminator with a classifier Q to ensure porosity constraints. (Figure adapted from Cao et al. \citep{Cao2022ReconstructionGAN}, with permission).}
    \label{fig:Cao2022ReconstructionGAN}
\end{figure}

Addressing property-controlled generation, Cao et al. (2022) \citep{Cao2022ReconstructionGAN} proposed combining InfoGAN with StyleGAN to incorporate prior porosity information while handling multi-scale features. Their architecture, illustrated in Figure \ref{fig:Cao2022ReconstructionGAN}, processes information hierarchically from coarse background features to fine microscopic details through adaptive instance normalization. A mapping network transforms the latent space containing both random noise and porosity information, while a classifier ensures the porosity remains within reasonable bounds.

For unpaired data scenarios, Liu et al. (2022) \citep{Liu2022MultiscaleNetworks} enhanced StyleGAN by incorporating CycleGAN for unpaired data handling, enabling style transfer between domains without requiring paired training data. These hybrid implementations facilitate multi-scale information fusion and domain-specific style transfer, making them particularly effective for complex reconstruction tasks that demand handling of varied resolutions and features.

Each architectural variation has demonstrated distinct performance characteristics across different materials and scales. StyleGAN2-ADA implementations have achieved impressive similarity metrics, with Thakre et al.'s work \citep{Thakre2023QuantificationModels} showing SSIM values ranging from 0.94 to 0.99 for dual-phase steels, and Haribabu et al.'s implementation \citep{Haribabu2023AData} demonstrating excellent phase fraction accuracy with differences of less than 1.2\% between original and synthetic images. In the realm of hybrid approaches, Cao et al.'s implementation \citep{Cao2022ReconstructionGAN} maintained relative errors around 10\% in total porosity reconstruction, while Li et al.'s EFPN integration \citep{Li2024Three-dimensionalNetwork} showed particular strength in preserving both structural and physical properties of porous media.

These advancements in StyleGAN architectures have significantly enhanced the field's capability to generate high-quality, physically accurate microstructure reconstructions. The various architectural innovations have enabled more precise control over generated features while successfully maintaining the statistical and physical properties essential for material science applications, thereby establishing StyleGAN as an effective framework for microstructure reconstruction tasks.

\subsection{Current Limitations and Future Directions}

The extension of StyleGAN architectures to large three-dimensional volumes remains the primary constraint specific to this paradigm. While StyleGAN excels at high-resolution 2D generation, current 3D implementations typically operate at resolutions of $128^3$ voxels or smaller \citep{Huang2022Deep-learning-basedMethod}. The computational and memory requirements scale cubically with resolution, creating practical barriers for direct 3D generation at scales necessary for representative volume element analysis or accurate flow simulation.

The interpretability of the latent space, while improved compared to traditional GANs, remains incomplete for physical property control. Cao et al. \citep{Cao2022ReconstructionGAN} addressed this through their CISGAN architecture combining InfoGAN with StyleGAN, achieving relative errors around 10\% in total porosity reconstruction. However, the direct mapping between latent dimensions and specific physical properties such as porosity, tortuosity, or connectivity remains unclear. Current implementations achieve property control through additional conditioning mechanisms rather than through direct latent space manipulation, suggesting that disentanglement is not complete with respect to physically meaningful properties.

Multi-phase and complex geometry reconstruction presents further challenges specific to style-based architectures. Most StyleGAN implementations focus on binary or simple multi-phase systems. Li et al. \citep{Li2024Three-dimensionalNetwork} successfully reconstructed sandstone, limestone, and silica particles, but required 16,000 2D images for each material type. The progressive growing and style injection mechanisms may not adequately capture the distinct statistical properties of each phase simultaneously in more complex multi-phase systems.

Training data requirements, while reduced compared to traditional GANs through adaptive augmentation, vary considerably across implementations. Haribabu et al. \citep{Haribabu2023AData} achieved high-quality generation with 1,225 training images through StyleGAN2-ADA, while Thakre et al. \citep{Thakre2023QuantificationModels} utilised only 300 images across six classes of dual-phase steels. The tension between data availability and generation quality persists, particularly for novel materials where extensive characterisation data may not exist.

Architecture-specific future directions address these constraints along several fronts. The development of hierarchical multi-resolution frameworks could overcome 3D scalability limitations; rather than generating complete structures at high resolution, future architectures could generate coarse-scale features at low resolution and progressively refine local regions requiring higher detail. Huang et al. \citep{Huang2022Slice-to-voxelNetworks} demonstrated that StyleGAN's progressive growing naturally supports this approach, achieving comparable quality to voxel-to-voxel methods through their slice-to-voxel implementation. Transfer learning and few-shot generation capabilities merit attention, as the success of StyleGAN2-ADA with limited datasets \citep{Haribabu2023AData} suggests potential for pre-training on diverse microstructures followed by fine-tuning on limited target data, which could dramatically reduce data requirements for new material systems. The exploration of latent space structure for property optimisation represents another frontier; Thakre et al. \citep{Thakre2023QuantificationModels} demonstrated comprehensive similarity quantification achieving SSIM values between 0.94 and 0.99, and future work could develop systematic mapping functions between latent codes and target properties, enabling inverse design through latent space optimisation. Multi-modal integration also shows particular promise, as Liu et al. \citep{Liu2022MultiscaleNetworks} successfully combined micro-CT and SEM images at different scales through their StyleGAN2-ADA and CycleGAN hybrid approach, maintaining porosity distribution and morphological characteristics across scales.

\begin{table}[htbp]
    \centering
    \caption{Technical Variations of StyleGAN Architectures in Material Reconstruction}
    \label{tab:stylegans-variants}
    \footnotesize
    \begin{tabular}{|p{1.8cm}|p{3cm}|p{3cm}|p{1.8cm}|p{2.2cm}|}
        \hline
        \textbf{Architecture Type} & \textbf{Key Features} & \textbf{Style Injection Method} & \textbf{Notable Implementations} & \textbf{Primary Applications} \\
        \hline
        Original StyleGAN & 
        Mapping network (8 layers) \newline
        Six connected blocks \newline
        Progressive resolution growth \newline
        Dual-channel control system & 
        AdaIN operations after each layer \newline
        Noise injection for local variations \newline
        Progressive resolution control (8$\times$8 to 128$\times$128) & 
        Fokina et al. (2020) & 
        Microstructure synthesis for porous media \\
        \hline
        StyleGAN2-ADA & 
        Adaptive discriminator augmentation \newline
        Dynamic training adjustment \newline
        Improved training stability \newline
        Path length regularization & 
        AdaIN with augmentation-aware mixing \newline
        Multi-resolution style control \newline
        Style demodulation & 
        Thakre et al. (2023) \newline
        Haribabu et al. (2023) \newline
        Huang et al. (2022) & 
        Limited dataset applications (DP steels, titanium alloys) \\
        \hline
        Hybrid StyleGAN Architectures & 
        InfoGAN + StyleGAN \newline
        CycleGAN + StyleGAN2 \newline
        Multi-scale feature fusion & 
        Combined style injection methods \newline
        Domain-specific style transfer \newline
        Multi-scale information fusion & 
        Li et al. (2024) \newline
        Cao et al. (2022) \newline
        Liu et al. (2022) & 
        Multi-scale reconstruction, unpaired data handling \\
        \hline
    \end{tabular}
    \par\vspace{4pt}
    \noindent\textit{Note:} ADA = Adaptive Discriminator Augmentation; EFPN = Extended Feature Pyramid Network
\end{table}

\pagebreak

\section{Hybrid Architecture GANs}
\label{AE-GAN}

Hybrid architecture GANs represent a significant advancement in generative modeling for porous media reconstruction by combining the generative capabilities of GANs with complementary architectures that address specific limitations. The integration of Auto-Encoders or Variational Auto-Encoders with GANs has emerged as an approach to address some of the fundamental challenges of GAN models while enhancing reconstruction quality and training stability. Traditional GANs face several key limitations including the difficulties in generating realistic and well-connected porous structures \citep{Shams2020CoupledMedia}, instability during the training process, and challenges in 2D to 3D reconstruction \citep{Zhang2021Slice-to-voxelModel}, and issues with model collapse and generation diversity \citep{Zhang20213DAuto-encoders}. Additionally, conventional GANs struggle with capturing multi-scale features and maintaining accurate statistical properties of the porous structures, particularly when dealing with complex 3D reconstructions \citep{Shams2020CoupledMedia, Li2023DeepImage}.

This chapter examines hybrid architectures organized into four primary categories including VAE-GAN based architectures that leverage probabilistic latent spaces for controlled generation, GAN with Autoencoder (GAN-AE) approaches using traditional autoencoders for feature extraction, GAN with Specialized Architectures incorporating domain-specific neural networks, and Multi-Model Integration combining multiple distinct generative paradigms.

\subsection{VAE-GAN Based Architectures}

VAE-GAN architectures combine the probabilistic encoding of Variational Autoencoders with the adversarial training of GANs. This hybridization addresses the blurriness often associated with VAE reconstructions while providing GANs with a structured latent space that enables controlled generation and improved training stability. The VAE component learns a probabilistic mapping from input images to a latent distribution, while the GAN component ensures generated samples maintain high visual quality. The evolution of VAE-GAN architectures in porous media reconstruction has progressed along several distinct but interconnected research directions, each addressing specific challenges in microstructure generation and reconstruction fidelity.

The application of VAE-GAN to porous media reconstruction began with frameworks designed to generate diverse plausible structures from limited input data. Feng et al. (2020) \citep{Feng2020AnLearning} introduced a BicycleGAN framework that combines conditional VAE-GAN with conditional Latent Regressor GAN for 3D reconstruction from single 2D images. Their architecture employs an encoder with convolutional layers and residual blocks, enabling the generation of diverse 3D structures from the same 2D input through Gaussian noise injection. This dual-pathway approach addresses a fundamental challenge in reconstruction where a single 2D image corresponds to multiple plausible 3D structures, and the BicycleGAN framework captures this one-to-many mapping. The implementation demonstrated instantaneous reconstruction achieving a $3.6 \times 10^4$ speedup compared to classical methods for $128^3$ voxel reconstructions, establishing the practical viability of VAE-GAN approaches for porous media applications.

Building upon the multimodal reconstruction capabilities demonstrated by BicycleGAN, Li et al. (2023) \citep{Li20233DReconstruction} developed GANDB-SD, which extends the BicycleGAN framework to achieve unlimited-size reconstruction. Their approach establishes eight multimodal mappings using the BicycleGAN model as a dictionary for reconstruction at arbitrary scales. The method introduces a three-dimensional co-occurrence matrix function as the loss function and implements a diversity preservation strategy to prevent pattern repetition during reconstruction. This advancement addresses scalability limitations inherent in the original BicycleGAN formulation, enabling reconstruction at arbitrary scales through the multimodal dictionary approach.

While BicycleGAN-based approaches successfully generated diverse reconstructions, they inherited fundamental training challenges associated with VAE-GAN architectures. These challenges, including mode collapse, training instability, variational inference errors, and data scarcity issues, motivated a major line of research focusing on resolving the inherent limitations of the VAE-GAN framework. Zhang et al. (2021) \citep{Zhang20213DAuto-encoders} proposed a VAE-GAN architecture that directly addressed the issues of training instability and model collapse through the incorporation of determinantal point processes (DPPs) into the framework. The GDPP implementation transforms complex probability calculations into simple determinant calculations, providing a mathematically principled approach to improve generation diversity. This method particularly focused on balancing the complementary strengths of both architectures, combining GANs' ability to generate clear images with VAEs' probabilistic encoding capabilities. The determinantal point process framework ensures that generated samples maintain diversity by explicitly penalizing similarity between outputs, effectively addressing the mode collapse problem where generators produce limited variations.

Extending the effort to resolve VAE-GAN training challenges, Zhang et al. (2024) \citep{Zhang2024StochasticVAE-GAN} developed a two-discriminator VAE-GAN that tackles the more fundamental issue of error problems arising from variational inference in the VAE component. Their architecture features a multi-layer encoder network with progressive channel expansion and introduces a code discriminator as a second discriminator component. By replacing variational inference with adversarial learning through the code discriminator, the method addresses limitations of traditional VAE-GAN that stem from the assumption of Gaussian or mixed Gaussian posterior distributions. When the true posterior distribution differs significantly from these assumed distributions, traditional VAE-GAN produces blurred reconstructions. The code discriminator learns to distinguish between encoded latent vectors and random samples, training the encoder through adversarial learning rather than variational approximation. This approach maintains stable training for grayscale digital core reconstruction while avoiding the distributional mismatch problems inherent to variational inference.

Addressing a different but equally critical challenge in VAE-GAN training, Zhang et al. (2024) \citep{Zhang2024DA-VEGAN:Sets} proposed DA-VEGAN incorporating a $\beta$-variational autoencoder into a hybrid GAN architecture with differentiable data augmentation. Their model uses five Conv2D layers in the encoder with progressive filter increases, allowing penalization of strong nonlinearities in the latent space through the $\beta$ parameter. The $\beta$-VAE formulation provides explicit control over the trade-off between reconstruction fidelity and latent space regularization, enabling more interpretable latent representations. The differentiable augmentation scheme represents a key innovation for handling extremely small datasets, as it allows the model to learn from augmented data without the quality degradation typically associated with naive augmentation strategies. This method enables learning from datasets as small as 16 samples without mode collapse, achieving significant improvements for both stationary and non-stationary materials. The differentiable nature of the augmentation ensures that transformations are integrated into the learning process rather than simply expanding the training set with potentially unrealistic modified images.

In parallel with efforts to stabilize VAE-GAN training, researchers explored architectural innovations in the encoder component to improve feature extraction and representation learning. These approaches focused on enhancing the VAE encoder's ability to capture complex structural information through specialized network architectures. Li et al. (2022) \citep{Li2022CascadedImage} introduced the Cascaded Progressive Generative Adversarial Network (CPGAN) using a U-Net based generator architecture with encoder-decoder structure. Their model employs deep gray-padding structures and three-dimensional convolution to capture spatial characteristics, maintaining gray level distribution through specialized loss functions. The U-Net architecture, with its skip connections between encoder and decoder layers, enables the network to preserve fine-scale structural details during the encoding-decoding process. The cascaded architecture ensures continuity and variability between reconstruction layers for grayscale core images, demonstrating that hierarchical feature extraction through U-Net-style connections enhances reconstruction quality for complex microstructures.

Taking a fundamentally different approach to latent representation, Phan et al. (2022) \citep{Phan2022Size-invariantImage} proposed a multi-step framework combining Vector-Quantized Variational AutoEncoder (VQ-VAE) with size-invariant GANs and Image Transformers. Their approach addresses scalability challenges by first using VQ-VAE for discrete latent representation learning, then applying GANs for size-invariant generation. Unlike traditional VAE architectures that use continuous latent spaces, VQ-VAE quantizes the latent representation into discrete codes from a learned codebook. This discrete representation provides more stable training and enables the generation of structures at arbitrary scales without retraining. The method enables reconstruction of structures larger than $1000^3$ voxels while maintaining statistical representativeness, demonstrating that discrete latent representations can overcome size limitations of continuous VAE formulations.

Further advancing encoder architectures, Yin et al. (2023) \citep{Yin2023Three-dimensionalModels} presented a hybrid generative model integrating VAE and GAN with self-attention modules for granular porous media reconstruction. Their architecture replaces random noise in traditional GANs with hidden vectors from the VAE encoder's latent space, while self-attention modules capture both long-range and short-range dependencies in the structural patterns. The self-attention mechanism enables the network to identify relationships between distant spatial locations, which is particularly important for capturing the connectivity and spatial arrangement of granular materials. The method generates grain sets as a whole rather than treating them as simplified individuals, accurately reproducing complex morphology without artificiality. This holistic generation approach, enabled by the attention mechanism, ensures that reconstructed structures maintain physically realistic grain arrangements and contact networks.

\begin{figure}[H]
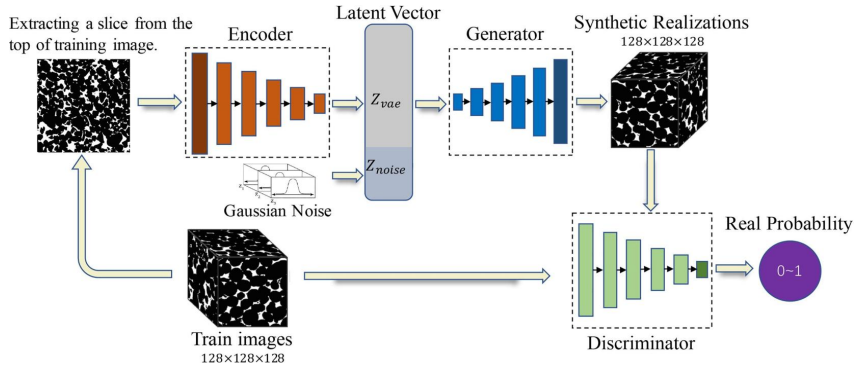

    \centering
    \includecroppedfigure[width=0.9\textwidth]
    \caption{Hybrid GAN-VAE architecture for 2D-to-3D reconstruction. A 2D slice is encoded into latent vector $z_{vae}$, concatenated with Gaussian noise $z_{noise}$, and fed to the generator to produce 3D synthetic realizations (128³ voxels). The discriminator distinguishes between real training volumes and generated outputs. This approach combines learned structural features with stochastic sampling for diverse, high-quality 3D reconstructions. (Figure adapted from Zhang et al. \citep{Zhang2021Slice-to-voxelModel}, with permission).}
    \label{fig:zhang}
\end{figure}

A distinct research direction emerged focusing on two-dimensional to three-dimensional reconstruction tasks using Wasserstein distance-based training strategies. These approaches specifically address the challenges of dimensional expansion while maintaining training stability through improved distance metrics. Zhang et al. (2021) \citep{Zhang2021Slice-to-voxelModel} introduced a hybrid GAN-VAE model designed for slice-to-voxel reconstruction. In this architecture, the encoder is connected before the generator, where the encoded latent vector is concatenated with Gaussian noise to serve as input to the generator (Fig. \ref{fig:zhang}). This design offers advantages from two perspectives, the encoded latent vector helps the generator better understand the structures and features of training images compared to pure Gaussian sampling, while the adversarial training strategy enhances the quality of synthetic realizations. The model establishes mapping between two-dimensional slices and three-dimensional structures through this hybrid approach, demonstrating that conditioning the generator on encoded features from 2D inputs enables more accurate 3D reconstruction than purely random generation.

Applying Wasserstein distance specifically to specialized applications, Feri et al. (2021) \citep{Feri2021APenalty} developed 3D-IWGAN that uses a Variational Autoencoder to generate latent vectors for the improved Wasserstein GAN for permeable pavement microstructure reconstruction. Their VAE processes 2D images through multiple convolutional layers to extract latent features, which are then used by the GAN component with enhanced gradient penalty for 3D generation. The improved Wasserstein GAN formulation with gradient penalty provides more stable gradients during training compared to traditional GAN losses, which is particularly important for the dimensional expansion task. This approach specifically addresses clogging analysis in permeable pavement applications, demonstrating the practical applicability of Wasserstein-based VAE-GAN to engineering problems requiring detailed pore-scale simulation.

\begin{figure}[H]
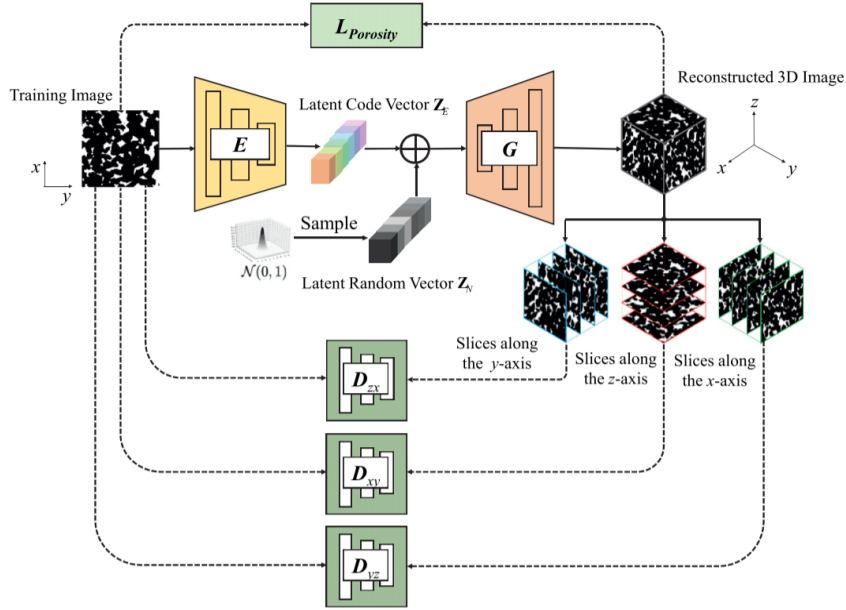

    \centering
    \includecroppedfigure[width=0.9\textwidth]
    \caption{EWGAN-GP architecture for 2D-to-3D reconstruction with multi-directional discrimination. The encoder (E) processes a 2D training image into latent code vector $\mathbf{Z}_E$, which is combined with random latent vector $\mathbf{Z}_N$ and fed to generator (G) to produce a 3D reconstruction. Three discriminators ($D_{xy}$, $D_{yz}$, $D_{zx}$) evaluate orthogonal cross-sections along x, y, and z axes to ensure statistical consistency across all spatial orientations. A porosity loss function ($L_{Porosity}$) constrains global structural properties. This multi-discriminator approach with Wasserstein distance and gradient penalty enables stable training while preserving pore network connectivity. (Figure adapted from Li et al. \citep{Li2023DeepImage}, with permission).}
    \label{fig:Li}
\end{figure}

Extending the Wasserstein approach to more complex 3D reconstruction scenarios, Li et al. (2023) \citep{Li2023DeepImage} introduced the EWGAN-GP method, combining VAE principles with Wasserstein GAN with gradient penalty. This implementation features an encoder working in concert with multiple discriminators operating along three orthogonal directions, specifically designed to tackle the challenges of 2D to 3D reconstruction while preserving structural characteristics and pore network connectivity (Fig. \ref{fig:Li}). The three discriminators evaluate cross-sections of the generated three-dimensional structure along x, y, and z directions, ensuring that the reconstruction maintains statistical consistency across all spatial orientations. The Wasserstein distance with gradient penalty serves as the evaluation metric, providing stable training even with the increased complexity of multiple discriminator networks. An additional porosity loss function constrains the overall three-dimensional structure, ensuring that global properties match the target while the discriminators ensure local slice-wise consistency.

The evolution of VAE-GAN architectures for porous media reconstruction demonstrates a progression from basic framework applications to sophisticated solutions addressing training stability, encoder design, and dimensional expansion challenges. These advances have established VAE-GAN as a robust and versatile approach for microstructure reconstruction, with ongoing research continuing to expand its capabilities and applicability.

\subsection{GAN with Autoencoder (GAN-AE)}

GAN-AE architectures integrate traditional autoencoders with GANs primarily for dimensionality reduction, feature extraction, or multi-scale reconstruction. Unlike VAE-GAN approaches, these methods use deterministic autoencoders without probabilistic constraints, focusing on hierarchical feature learning or post-processing enhancement. The autoencoder component typically handles either pre-processing of input data or post-processing of GAN outputs to add fine-scale details.

\begin{figure}[H]
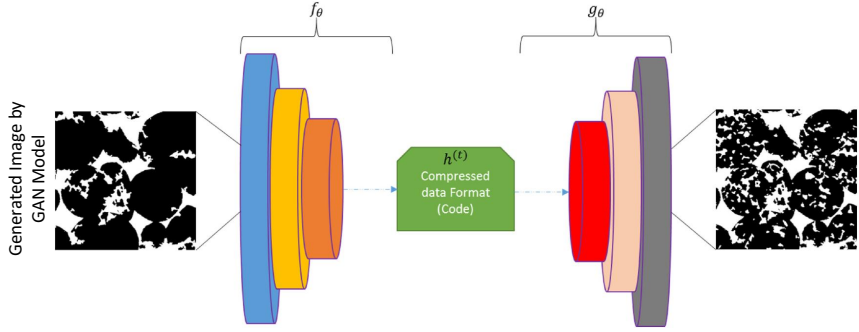

    \centering
    \includecroppedfigure[width=0.9\textwidth]
    \caption{Autoencoder architecture for post-processing GAN-generated images. The encoder $f_\theta$ compresses GAN-generated images (containing inter-grain pores) through progressively smaller feature layers into a compressed latent code representation $h^{(t)}$. The decoder $g_\theta$ reconstructs the image through progressively larger layers, incorporating fine-scale intra-grain porosity learned from high-resolution SEM training data. This two-stage approach enables multi-scale reconstruction by combining GAN-generated macro-porosity with AE-enhanced micro-porosity features. (Figure adapted from Shams et al. \citep{Shams2020CoupledMedia}, with permission).}
    \label{fig:shams2}
\end{figure}

Early developments in this field, as demonstrated by Shams et al. \citep{Shams2020CoupledMedia}, introduced a coupled GAN-AE approach where the Auto-Encoder acts as a post-processing enhancement tool. In this implementation, the GAN first generates images containing inter-grain pores, which are then fed into an Auto-Encoder specifically trained on high-resolution SEM images to incorporate intra-grain porosity (Fig. \ref{fig:shams2}). This approach proved particularly effective for multi-scale reconstruction of both sandstone and carbonate samples, successfully capturing both macro and micro-porosity features that traditional GANs struggled to represent.

Volkhonskiy et al. (2022) \citep{Volkhonskiy2022GenerativeSlices} proposed a deep learning architecture combining a convolutional autoencoder with GAN for three-dimensional reconstruction from two-dimensional slices. Their approach uses an encoder that transforms input 2D slices to latent vector representations, with the generator serving dual function as both the autoencoder decoder and GAN generator. This architecture fits a distribution on all possible three-dimensional structures, recovering the most probable 3D structure from given central slices.

\begin{figure}[H]
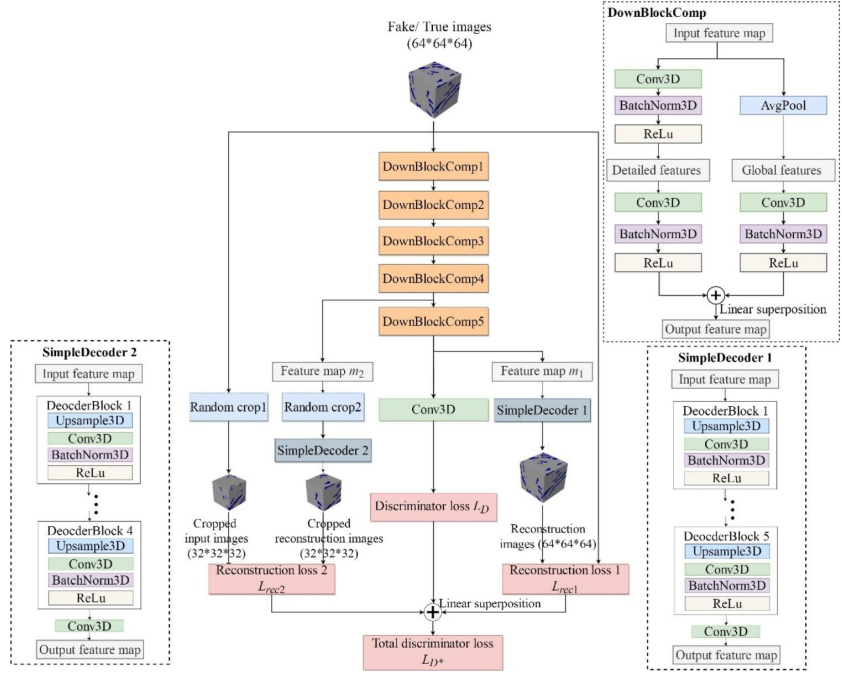

    \centering
    \includecroppedfigure[width=0.9\textwidth]
    \caption{The discriminator architecture of 3D-FGAN showing AE-based regularization with random crop operations and multiple decoder blocks for reconstruction. The discriminator processes fake/true images through DownBlockComp modules, applies random cropping, and uses SimpleDecoders to compute reconstruction losses alongside the discriminator loss for enhanced training stability. (Figure adapted from Zhang et al. \citep{Zhang20243D-FGAN:Cores}, with permission).}
    \label{fig:zhang24_2}
\end{figure}

Most recently, Zhang et al. (2024) \citep{Zhang20243D-FGAN:Cores} introduced the 3D-FGAN architecture where the Auto-Encoder is integrated into the discriminator rather than the generator. This novel approach employs the Auto-Encoder as a regularization technique within the discriminator, combined with random crop operations to prevent overfitting (Fig. \ref{fig:zhang24_2}). The architecture is further enhanced by incorporating skip-layer excitation (SLE) blocks in the generator, resulting in faster convergence and high-quality reconstructions while maintaining computational efficiency.

\subsection{GAN with Specialized Architectures}

This category encompasses hybrid architectures that integrate GANs with specialized neural networks designed for specific reconstruction tasks. Two distinct architectural paradigms have emerged, U-Net-based approaches leveraging hierarchical spatial feature extraction through encoder-decoder structures with skip connections, and RNN-based approaches capturing sequential dependencies for layer-by-layer generation.

\begin{figure}[H]
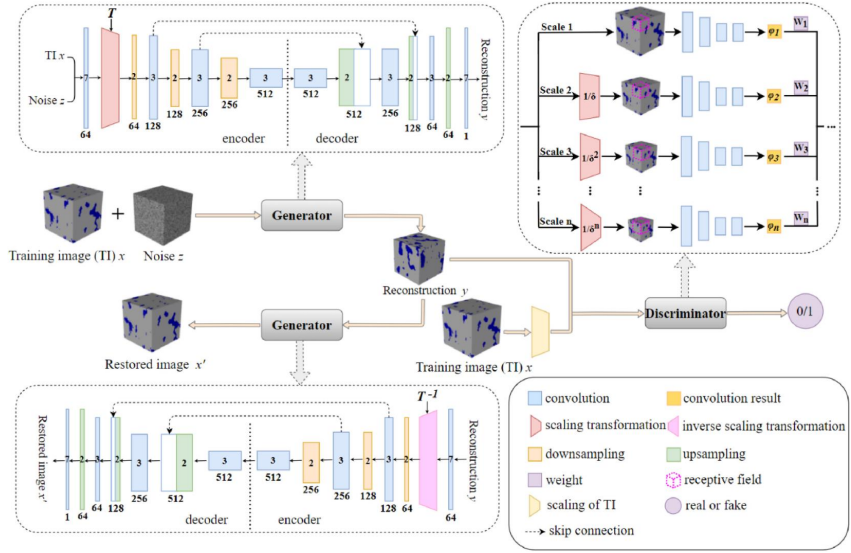

    \centering
    \includecroppedfigure[width=0.9\textwidth]
    \caption{Architecture of 3D-MSPGAN showing the dual-generator framework. The network combines multi-scale discrimination with scaling transformation to enable single-image training. The first generator (top) creates reconstructions while the second generator (bottom) performs inverse scaling for validation. Skip connections (dotted lines) facilitate information flow between corresponding encoder-decoder layers. The discriminator (right) evaluates features across multiple scales through scaling transformations. (Figure adapted from Zhang et al. \citep{Zhang20223DNetworks}, with permission).}
    \label{fig:Zhang20223DNetworks}
\end{figure}

Zhang et al. (2022) \citep{Zhang20223DNetworks} proposed 3D-MSPGAN (Three-dimensional Multi-scale Pattern Generative Adversarial Network), employing a U-Net-based generator with scaling transformation for super-resolution output and multi-scale discrimination. A key innovation is the dual-generator mechanism leveraging the generator's automorphism property, where the system attempts to restore the original image after reconstruction, ensuring preservation of both local and global characteristics. The architecture enables single-image training while achieving up to eight-fold volume increase with maintained structural fidelity (Fig. \ref{fig:Zhang20223DNetworks}).

Building upon the U-Net framework for large-scale applications, Zhang et al. (2022) \citep{Zhang2022FastLearning} proposed LGCNN based on Wasserstein GAN with a U-Net-inspired generator containing encoder-decoder structure with skip connections. Their framework combines an ML-based reconstruction method with an adjustable splicing algorithm for representative element volume reconstruction, achieving $600^3$ voxel reconstruction in 10 minutes while maintaining higher accuracy on structural parameters compared to traditional methods.

Zheng et al. (2024) \citep{Zheng2024AEvaluation} further extended U-Net-based architectures by combining U-Net multiscale features with attention mechanisms. Their architecture incorporates three specialized modules including a multiscale channel aggregation module, a hierarchical feature aggregation module, and a convolutional block attention module. The loss function combines image regularization loss with Wasserstein distance loss, achieving high-fidelity 3D structure generation with anisotropy index measurement.

In contrast to spatial hierarchical processing in U-Net architectures, Zhang et al. (2023) \citep{Zhang2023PM-ARNN:Network} introduced PM-ARNN incorporating recurrent neural networks with GANs for 2D-to-3D reconstruction. Their hybrid architecture features an encoder with 2D CNNs and fully-connected layers, an LSTM-based frame predictor, and a reference module with prior and inference LSTM networks. The adversarial training strategy with RNN enables better perception of 3D pore structures and enhanced reconstruction of adjacent layers while preserving long-distance connectivity. Unlike U-Net approaches that process entire volumes through hierarchical spatial decomposition, PM-ARNN's recurrent structure explicitly models sequential generation, maintaining hidden states that capture information from previously generated layers to inform subsequent layer generation.

These specialized architectures demonstrate distinct approaches to incorporating domain-specific inductive biases into GAN frameworks. U-Net-based architectures leverage spatial hierarchical decomposition through skip connections, enabling simultaneous multi-scale feature processing and proving effective for holistic volume generation. RNN-based architectures impose sequential processing constraints that explicitly model inter-layer dependencies through recurrent connections, providing advantages when maintaining consistency across adjacent layers is critical. The choice between paradigms involves trade-offs in computational efficiency, structural consistency mechanisms, and applicability to different reconstruction scenarios, with U-Net architectures enabling parallel spatial processing while RNN architectures provide explicit sequential dependency modeling.

\subsection{Multi-Model Integration}

Multi-model integration represents the most complex category of hybrid architectures, combining GANs with multiple distinct generative paradigms to address diverse reconstruction challenges. These approaches leverage complementary strengths of different frameworks including InfoGAN's disentangled representations, StyleGAN's style control, reinforcement learning's optimization capabilities, and CycleGAN's unpaired translation.

Addressing the challenge of property-controlled generation with multi-scale feature integration, Cao et al. (2022) \citep{Cao2022ReconstructionGAN} introduced CISGAN combining InfoGAN and style-based GAN architectures guided by prior information. Their architecture incorporates porosity distribution as prior information into the latent space to control pore distribution, with a classifier Q in the discriminator ensuring porosity remains within reasonable ranges. Multi-scale information is applied through style transfer at each layer, progressively optimizing background information, pore structure, and micro-scale details for natural digital rock generation. This hierarchical style transfer approach enables systematic control over features at different spatial scales.

While architectural integration approaches achieve property control through direct conditioning, inverse design problems requiring simultaneous optimization of multiple interdependent properties motivated alternative strategies. Nguyen et al. (2022) \citep{Nguyen2022SynthesizingLearning} developed the GAN-AC framework combining generative adversarial networks with actor-critic reinforcement learning. Their architecture reduces the GAN component from five to four convolutional blocks for computational efficiency while adding stride-1 convolutional layers after half-stride convolutions. The actor-critic component iteratively optimizes generation across eight properties simultaneously (porosity, specific surface area, permeability, average clustering, graph density, degree assortativity, local efficiency, and graph transitivity), enabling generation of microstructures matching user-defined physical quantities within 5\% error margin of target values.

Addressing data scarcity challenges in super-resolution tasks, Chen et al. (2020) \citep{Chen2020Super-resolutionNetworks} introduced SRCycleGAN for real-world rock microcomputed tomography image enhancement from unpaired training data. The approach leverages CycleGAN's bidirectional mapping with two ResNet-based generators and PatchGAN discriminators, trained using combined least-squares adversarial loss, cycle consistency loss, and identity loss to preserve structural features during resolution enhancement. Experimental validation across multiple upsampling factors ($1.8\times$, $2.0\times$, $4.0\times$) on sandstone samples demonstrated superior peak signal-to-noise ratio compared to bicubic interpolation while maintaining statistical accuracy in porosity, two-point correlation functions, and pore size distributions within 1\% of ground truth, enabling large field-of-view high-resolution reconstruction without requiring spatially registered LR-HR image pairs.

Extending CycleGAN-based unpaired learning from two-dimensional super-resolution to three-dimensional reconstruction, Wu et al. (2023) \citep{Wu2023AnFeatures} proposed an end-to-end approach combining FastGAN and CycleGAN for heterogeneous porous media reconstruction from limited unpaired data. Their FastGAN architecture features a generator with GLU/BatchNorm components and SLE modules for multi-scale processing. The framework synthesizes thousands of plausible images from approximately 100 unpaired samples, then uses CycleGAN to assimilate fine-scale structures from 2D high-resolution images into 3D low-resolution images for digital rock reconstruction, eliminating the requirement for spatially registered paired training data.

The choice among these multi-model approaches depends on the specific reconstruction scenario: CISGAN when prior information is available for property control, GAN-AC when explicit conditioning relationships are complex or unknown and iterative optimization is acceptable, CycleGAN-based super-resolution when paired LR-HR training data cannot be obtained while resolution enhancement is needed, and FastGAN-CycleGAN when dimensional expansion from 2D to 3D with limited unpaired data is required.

\subsection{Multi-Stage and Transformer-Enhanced Architectures}

Recent developments have integrated transformer architectures and multi-stage processing pipelines to enhance feature extraction and address multiple reconstruction challenges simultaneously. Zhang and Zhang \citep{Zhang20253DTransformer} developed Style-Transformer GAN (STGAN), integrating StyleGAN with Transformer attention mechanisms for 3D digital rock reconstruction. Their dual-architecture employs Transformer blocks to extract features across multiple resolution scales while StyleGAN provides style transfer through adaptive instance normalization. The discriminator incorporates a classifier leveraging porosity distribution as priori information. For $64^3$ shale reconstructions, STGAN achieved porosity statistics closely matching training data with minimal Morphological Pattern Comparison difference degrees across all three directions, and pore network analysis closely approximated target values. Though computational requirements exceed StyleGAN baselines, the architecture demonstrates balanced efficiency.

Addressing data scarcity through latent space manipulation, Zhang et al. \citep{Zhang2025ForADA-PGGAN} introduced a hybrid combining ResNet-VGG Inversion Optimization Network (RVION) with Adaptive Data Augmentation Progressive Growing GAN (ADA-PGGAN). RVION employs a two-stage encoder where ResNet50 predicts latent codes followed by VGG16 optimization, with these codes interpolated to generate intermediate slices between sparse scanned images. ADA-PGGAN implements nine progressive training stages from $4 \times 4$ to $1024 \times 1024$ resolution with adaptive augmentation mechanisms. Applied to sandstone, the method reconstructed large volumes using only 100 sparse slices, reducing imaging costs 10-fold compared to full 3D scanning. Sliced Wasserstein Distance evaluation and pore structure analysis demonstrated close approximation to ground truth. Comparative analysis against MS-GAN, IWGAN-GP, and improved SliceGAN showed superior microstructure parameter preservation despite requiring only one training sample versus 50-124 samples for competing methods.

Implementing comprehensive multi-stage processing, Liu et al. \citep{LIU2025PoreAuto-segmentation} developed a framework integrating three distinct GAN architectures to address shale imaging limitations: StyleGAN2-ADA for HR image augmentation from limited real images, SRCycleGAN for unpaired super-resolution achieving 8× enhancement without paired training samples, and U-Net-based SegGAN for multi-mineral auto-segmentation. The sequential pipeline addresses unavailable paired LR-HR datasets through cycle-consistency loss with bidirectional mapping. Testing on shale samples demonstrated that StyleGAN2-ADA accurately reproduced porosity and organic matter proportions, SRCycleGAN achieved higher PSNR and entropy values compared to bicubic interpolation, and SegGAN improved multi-mineral segmentation accuracy. The framework's capability to characterize complex multiscale and multi-mineral microstructures was validated through statistical and physical property analyses, with potential applicability to other heterogeneous porous media including carbonates, coal, and tight sandstones.

Ma et al.\ \citep{Ma2026SDWGAN} introduced the Stationary and Discrete Wavelet-Enhanced Generative Adversarial Network (SDWGAN) to address persistent artifact generation in GAN-based super-resolution of digital rock CT images. The architecture pairs a lightweight SwinIR-based transformer generator with a discriminator operating exclusively in the Stationary Wavelet Transform (SWT) domain, enabling direct discrimination between genuine high-frequency rock textures and GAN-induced artifacts. A hybrid fidelity loss combining SWT and Discrete Wavelet Transform (DWT) constraints provides multi-scale frequency supervision across both full- and half-resolution subbands. Evaluated on the DeepRock-SR dataset across carbonate, sandstone, and coal lithologies at $4\times$ super-resolution, SDWGAN achieved 0.63--2.12~dB PSNR and 0.01--0.11 SSIM gains over RGB-domain loss-based models, with porosity and permeability estimates reaching 98\% similarity to high-resolution references.

\subsection{Current Limitations and Future Directions}

Training complexity represents a primary limitation specific to hybrid architectures. The integration of multiple network components significantly increases architectural complexity and training difficulty. Zhang T. et al. \citep{Zhang2024StochasticVAE-GAN} reported approximately 68,000 seconds for their two-discriminator VAE-GAN architecture, and Zhang T. et al. \citep{Zhang20243D-FGAN:Cores} required 7.3 hours for their 3D-FGAN incorporating auto-encoders within the discriminator. The increased complexity consistently results in longer convergence times compared to standard GAN implementations.

The balance between reconstruction accuracy and generation diversity remains a fundamental challenge unique to VAE-GAN architectures. Zhang F. et al. \citep{Zhang2021Slice-to-voxelModel} noted that while their hybrid GAN-VAE model improved 3D reconstruction quality from 2D slices, the trade-off between visual quality and latent space interpretability persists. Yin et al. \citep{Yin2023Three-dimensionalModels} addressed this partially through self-attention modules, yet the inherent tension between the VAE's tendency toward blurred reconstructions and the GAN's focus on sharp features remains unresolved.

Memory constraints affect hybrid architectures particularly severely due to the simultaneous storage requirements of multiple network components. Li et al. \citep{Li20233DReconstruction} demonstrated unlimited-size reconstruction with their GANDB-SD framework (up to $1000^3$ voxels), but this required 2.6 GB of GPU memory for just $128^3$ elements with batch size of 2. Phan et al. \citep{Phan2022Size-invariantImage} achieved structures exceeding $1000^3$ voxels through their VQ-VAE and GAN combination, but required approximately 9 hours for a single $1024^3$ volume.

The additional hyperparameters introduced by hybrid architectures complicate optimisation, particularly under data-scarce conditions. Zhang Y. et al. \citep{Zhang2024DA-VEGAN:Sets} successfully demonstrated training on datasets as small as 16 samples through $\beta$-VAE incorporation with differentiable augmentation, but achieving this required careful tuning of the $\beta$ parameter to balance reconstruction fidelity and latent space regularisation.

Architecture-specific future directions address several of these constraints. The development of more efficient training strategies is a critical need; Zhang Y. et al. \citep{Zhang2024DA-VEGAN:Sets} demonstrated that differentiable augmentation could reduce training time to approximately 30 minutes for 20,000 epochs, and future work should explore meta-learning and transfer learning strategies to further reduce both data requirements and training time. The exploration of adaptive hybrid architectures shows particular promise, as Zhang et al. \citep{Zhang20243D-FGAN:Cores} demonstrated that saving parameters for reuse could reduce subsequent generation time from 26,304 seconds to just 41 seconds, suggesting that adaptive parameter sharing could significantly improve efficiency. 

Multi-model integration beyond binary combinations represents another frontier; Cao et al. \citep{Cao2022ReconstructionGAN} combined InfoGAN and StyleGAN, while Chen et al. \citep{Chen2020Super-resolutionNetworks} demonstrated CycleGAN for unpaired super-resolution, and future architectures could integrate three or more complementary models, potentially combining VAE for latent space structure, GAN for generation quality, and reinforcement learning for property optimisation within a unified framework. The development of hierarchical hybrid architectures also deserves attention, as Li et al. \citep{Li2022CascadedImage} achieved up to $1024^3$ voxel generation through cascaded progressive reconstruction, suggesting that different architectural components could handle different scales to combine the efficiency of progressive growing with the stability of VAE-based approaches.

The evolution of GAN architectures for porous media reconstruction has demonstrated significant advancement from fundamental implementations to specialized approaches addressing specific challenges. Each architectural category - fundamental GANs, multi-scale GANs, conditional GANs, attention-enhanced GANs, style-based GANs, and hybrid architectures - has contributed unique capabilities to the field. While fundamental GANs established the baseline feasibility, subsequent developments have enhanced control over generated features, improved handling of multi-scale structures, and increased reconstruction quality through various architectural innovations. These developments reflect the field's progression toward more sophisticated and capable reconstruction systems, though challenges remain in areas such as computational efficiency, training stability, and handling of large-scale reconstructions. For implementation details and code availability of the discussed methods, readers are referred to Table \ref{tab:code_availability} in the Appendix.

\pagebreak

\section{Results and Discussions}

The systematic analysis of 96 peer-reviewed articles implementing GAN-based approaches for porous media reconstruction reveals significant technological progress alongside persistent challenges. The field has evolved through six distinct architectural paradigms, each addressing specific reconstruction requirements while introducing unique trade-offs between accuracy, computational efficiency, and controllability.

Reconstruction capabilities have expanded dramatically since initial implementations. Early work by Mosser et al. \citep{Mosser2017ReconstructionNetworks} achieved porosity accuracy within 1\% for $64^3$ voxel volumes, establishing baseline feasibility. Current architectures demonstrate substantially enhanced performance: Zhu et al.'s IPWGAN \citep{Zhu2024GenerationNetworks} generates structures up to $2{,}200^3$ voxels with 79\% reduction in permeability prediction errors, while Zhang et al.'s implementations \citep{Zhang2022AWGAN-GP} maintain porosity accuracy within 0.1--0.3\% of reference values. This volumetric expansion, representing a 50,000-fold increase, enables representative volume element analysis previously impossible with GAN-based methods.

Architectural evolution reflects systematic responses to identified limitations. Approximately 40\% of recent implementations employ WGAN-GP loss functions \citep{Li2022DigitalPenalty, Xu2024Multi-criteriaProperties, Chi2024ReconstructionNetwork}, addressing mode collapse and gradient instability that plagued early architectures. The introduction of Hinge Loss for handling training-generation size disparities \citep{Zhang2025PredictionModelling} and batch normalization techniques \citep{Zhang2022AWGAN-GP} has further stabilized training processes. Multi-scale architectures particularly excel at capturing hierarchical features, with concurrent training reducing computational time from 40,000 seconds (SinGAN) to 5,863 seconds (SMAGAN) \citep{Zhang2023StochasticMechanisms} while maintaining reconstruction quality.

A clear trajectory toward reduced data requirements emerges across architectural paradigms. The 2D-to-3D reconstruction approach, pioneered by SliceGAN \citep{Kench2021GeneratingExpansion} and derivative frameworks, eliminated the need for expensive volumetric imaging by generating 3D structures from 2D microscopy images, achieving porosity accuracy within 0.6\% of $\mu$CT measurements \citep{Amiri2024NewGANs}. Style-based architectures further reduced data demands through adaptive augmentation; StyleGAN2-ADA achieves phase fraction accuracy within 1.2\% using only 300 training images \citep{Thakre2023QuantificationModels}. Hybrid approaches push this boundary further: DA-VEGAN demonstrates successful training on datasets as small as 16 samples \citep{Zhang2024DA-VEGAN:Sets}, while FastGAN-CycleGAN combinations generate thousands of synthetic images from approximately 100 unpaired samples \citep{Wu2023AnFeatures}. This progression, from requiring thousands of volumetric training samples to functioning with minimal 2D inputs, represents a fundamental shift in accessibility, enabling reconstruction for laboratories and applications where extensive imaging data is unavailable.

The introduction of controllability mechanisms has transformed GANs from purely generative tools to targeted design instruments, though each mechanism introduces specific trade-offs. Conditional architectures provide the most direct property control. Single-property conditioning achieves volume fraction accuracy within 1\% error \citep{Kishimoto2023ConditionalFractions}, multi-property systems simultaneously manage up to eight parameters with 5\% accuracy \citep{Nguyen2022SynthesizingLearning}, and dual conditioning on porosity and geological depth achieves $R^2=0.95$ \citep{Sadeghkhani2025PCP-GAN:Networks}. However, controlling strongly correlated properties remains challenging, as interdependencies between porosity, permeability, and structural characteristics create complex optimisation landscapes \citep{Zheng2022DigitalNetworks}.

Attention mechanisms improve structural preservation through selective feature weighting, with implementations such as SASGAN \citep{Zhang20233DCBAMs} achieving excellent fidelity and 1-second subsequent reconstruction times. However, the quadratic memory scaling of self-attention with input size imposes a fundamental volume ceiling of approximately $80^3$ voxels on single-GPU configurations \citep{Zhang2023StochasticMechanisms, Wang20233DStructure}. Style-based architectures enable hierarchical control of coarse and fine features through latent space manipulation \citep{Cao2022ReconstructionGAN}, yet remain constrained to $128^3$ voxels for 3D implementations \citep{Huang2022Deep-learning-basedMethod}. Hybrid VAE-GAN combinations improve training stability while enabling probabilistic latent space interpretation \citep{Zhang20213DAuto-encoders}, but require extended training times often exceeding 48 hours. This fundamental tension between control sophistication and computational feasibility remains a central challenge as the field advances.

Recent developments have demonstrated continued progress in addressing fundamental challenges. Memory efficiency advances through octree-based sparse convolution \citep{Ugolkov2025OptimizedTomography} enable 16× super-resolution to $512^3$ voxels while processing only mixed boundary nodes, representing substantial improvement over previous 8× limitations. Data scarcity mitigation through latent space manipulation techniques \citep{Zhang2025ForADA-PGGAN} achieves reconstruction from single training samples versus 50-124 samples required by competing methods. Specialized applications to complex materials including tight carbonates \citep{Zhou2025DigitalSliceGAN}, organic matter in shales \citep{Liu2025Three-dimensionalShales}, and multi-mineral systems \citep{LIU2025PoreAuto-segmentation} validate the versatility of GAN-based approaches across diverse geological formations.

Computational requirements vary by orders of magnitude across architectures, creating significant accessibility barriers. Basic DCGAN implementations converge within 800--1,000 iterations on standard hardware \citep{Zhang2022DigitalNetworks}, whereas sophisticated hybrid systems require specialised GPUs with 24--48GB memory and training times extending to several days \citep{Matsuda2022FrameworkNetwork}. Between these extremes, attention-enhanced architectures demand 6--12 hours of initial training but achieve 1-second subsequent reconstructions \citep{Zhang20233DCBAMs, Chi2024MultiscaleNetwork}, and style-based implementations require up to 54 hours for multi-scale hybrid configurations \citep{Liu2022MultiscaleNetworks}. This resource disparity risks limiting advanced methods to well-funded research groups, suggesting that the development of computationally efficient architectures, without sacrificing reconstruction quality, should be a priority for broadening practical adoption.

Evaluation metrics remain inconsistent across implementations, hindering systematic comparison. While porosity and two-point correlation functions appear universally, their adoption reflects convention rather than demonstrated sufficiency for capturing reconstruction quality. Advanced metrics such as Fr\'echet Inception Distance and Kernel Inception Distance \citep{He2023DigitalNetworks}, multi-point connectivity curves \citep{Zhang2023ReconstructionNetworks}, and comprehensive systems combining SSIM, PSNR, and statistical functions \citep{Zheng2024AEvaluation} have been introduced by individual groups but lack community-wide standardisation. The absence of benchmark datasets and unified evaluation protocols makes direct comparison between architectures difficult, and future work should establish standardised frameworks that assess both generation quality and computational efficiency across material types.

The integration of physics-based constraints into GAN architectures remains one of the most significant unresolved challenges across all paradigms. Current implementations predominantly optimise statistical similarity metrics, porosity, two-point correlation, Minkowski functionals, without explicitly enforcing physical conservation laws. Nguyen et al. \citep{Nguyen2022SynthesizingLearning} demonstrated one pathway by combining GANs with actor-critic reinforcement learning, achieving property matching within 5\% error margins, but this required iterative post-hoc optimisation over 1,000 training episodes rather than direct architectural integration. Zhang Y. et al. \citep{Zhang2025PredictionModelling} integrated DCGAN with pore network modelling for CO$_2$ storage efficiency prediction, and Cao et al.'s \citep{Cao2022ReconstructionGAN} multi-scale style injection suggests that physics constraints could be incorporated at appropriate scales within generation hierarchies. Wu et al. \citep{Wu2023AnFeatures} further suggested that combining data-driven approaches with physics-based models could enhance both accuracy and interpretability. However, direct incorporation of physical laws into the training loss or network architecture itself remains largely unexplored, and achieving this would represent a fundamental advance ensuring that generated structures are not only statistically realistic but also physically realisable.

Uncertainty quantification frameworks for GAN-based reconstruction remain largely underdeveloped despite their importance for engineering applications. Zhang Y. et al. \citep{Zhang2025PredictionModelling} made notable progress by developing methods to determine the minimum number of digital realisations needed to reproduce representative statistics of geometric and flow properties, generating an ensemble of 1,000 realisations for CO$_2$ storage efficiency prediction. Feng et al. \citep{Feng2019ReconstructionNetworks} incorporated Gaussian noise to generate multiple plausible reconstructions from limited input data, and Fokina et al. \citep{Fokina2020MicrostructureNetworks} validated generated structures through finite element analysis of elastic properties. However, comprehensive frameworks that characterise how uncertainty propagates from input data through the generation process to predicted structural and transport properties, providing confidence bounds essential for reliability-critical applications in reservoir engineering, battery design, and structural materials, remain absent.

Taken together, these cross-cutting challenges, physics integration, standardised evaluation, and uncertainty quantification, represent the primary barriers to transitioning GAN-based reconstruction from research demonstrations to reliable engineering tools. Addressing them will require not only architectural innovations within individual paradigms but also community-wide coordination on benchmarks, datasets, and validation protocols.

\begin{landscape}
    \begin{figure}[!p]
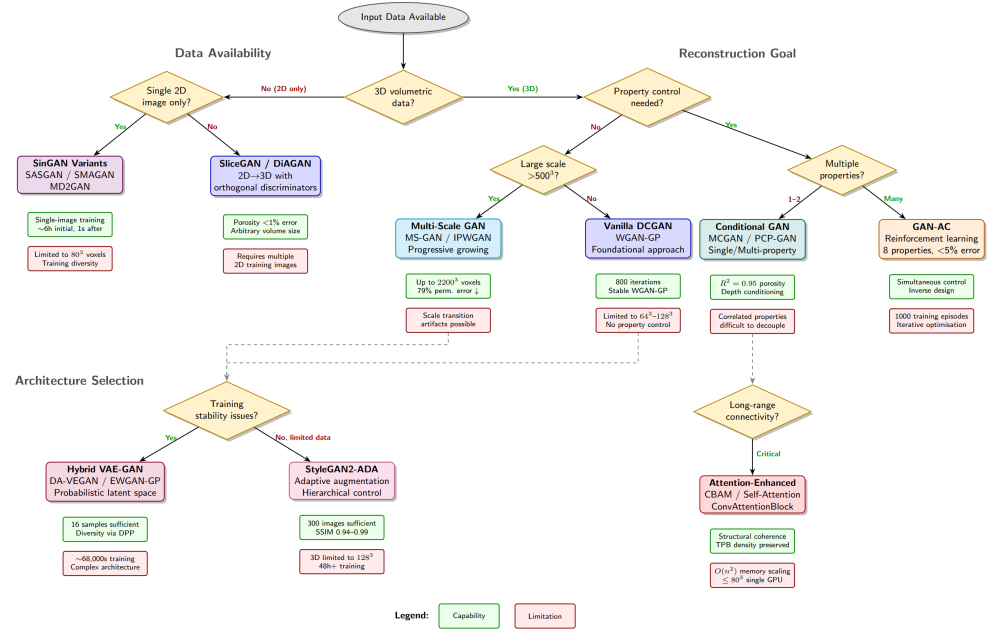

        \centering
        \includecroppedfigure[width=1\textwidth]
        \caption{Decision framework for selecting GAN architectures for porous media reconstruction. The flowchart guides practitioners through critical selection criteria: input data characteristics (2D vs 3D, quantity), reconstruction objectives (property control, volume scale), and architectural constraints (training stability, connectivity preservation). Each architecture pathway displays demonstrated capabilities (green boxes) with quantitative performance benchmarks and known limitations (red boxes) from reviewed literature. Dashed arrows indicate fallback pathways when primary methods encounter limitations. Three decision phases progress from data availability assessment through reconstruction goal specification to final architecture selection.}
        \label{fig:gan_selection_flowchart}
    \end{figure}
\end{landscape}

To guide practitioners through the complex landscape of GAN architectures, we present a decision framework (Figure \ref{fig:gan_selection_flowchart}) that synthesizes the quantitative benchmarks and architectural trade-offs from the reviewed implementations. The framework operates through three sequential phases including (1) data availability assessment (3D volumetric, multiple 2D, or single 2D image), (2) reconstruction objective specification (property control, volume scale, connectivity requirements), and (3) architectural constraint evaluation (training stability, data scarcity). Each pathway presents demonstrated capabilities alongside known limitations quantified through literature benchmarks.

The framework explicitly represents trade-offs at each decision point. Conditional GANs achieve $R^2=0.95$ porosity control \citep{Sadeghkhani2025PCP-GAN:Networks} but struggle with correlated properties. Multi-scale approaches generate up to $2{,}200^3$ voxels with 79\% permeability error reduction \citep{Zhu2024GenerationNetworks} yet face scale transition artifacts. Attention-enhanced architectures preserve structural coherence but impose $O(n^2)$ memory scaling limiting practical volumes to $\leq 80^3$ voxels \citep{Zhang20233DCBAMs, Zhang2023StochasticMechanisms}. Dashed arrows indicate alternative pathways when initial architectures encounter limitations, such as transitioning from vanilla DCGAN to hybrid VAE-GAN approaches when training stability issues arise. This framework serves as both a starting point for newcomers and a navigation tool for experienced practitioners exploring alternative approaches.

The implementation accessibility of the reviewed GAN-based approaches varies significantly across research groups. Table~\ref{tab:code_availability} provides a curated overview of publicly available implementations, organised by GAN architecture category, offering a direct starting point for researchers aiming to reproduce or build upon existing methods.

\begin{table}[!htbp]
    \centering
    \small
    \caption{Code Availability for GAN-based Porous Media Reconstruction Methods}
    \label{tab:code_availability}
    \begin{tabular}{|p{2cm}|p{3cm}|p{6cm}|}
        \hline
        \textbf{GAN Category} & \textbf{Reference} & \textbf{Code Availability} \\
        \hline
        \multirow{8}{2cm}{Vanilla GAN} 
        & Zhang Y. et al. \citep{Zhang2025PredictionModelling} & \url{https://github.com/ImperialCollegeLondon/porescale} \newline \url{https://github.com/iPMLab/Multi-physics-Network-Model} \\
        & Amiri H. et al. \citep{Amiri2024NewGANs} & \url{https://github.com/hamediut/True2Dto3Drecon} \\
        & Zhang T. et al. \citep{Zhang2023ANetwork} & \url{https://github.com/Hugsazp/3DRGAN} \\
        & Zhang T. et al. \citep{Zhang2022AWGAN-GP} & \url{https://github.com/vevvve/IWGAN-GP} \\
        & Kench S. et al. \citep{Kench2021GeneratingExpansion} & \url{https://github.com/stke9/SliceGAN} \\
        & Gayon-Lombardo A. et al. \citep{Gayon-Lombardo2020PoresBoundaries} & \url{https://github.com/agayonlombardo/pores4thought} \\
        & Coiffier G. et al. \citep{Coiffier20203DNetworks} & \url{https://github.com/randlab/diaGAN} \\
        & Mosser L. et al. \citep{Mosser2017ReconstructionNetworks} & \url{https://github.com/LukasMosser/PorousMediaGan} \\
        \hline
        \multirow{6}{2cm}{Multi-Scale GAN} 
        & Zhao X. et al. \citep{Zhao20253DNetworks} & \url{https://github.com/NBICLAB/CEM3DMG} \\
        & Zhu L. et al. \citep{Zhu2024GenerationNetworks} & \url{https://github.com/ImperialCollegeLondon/IPWGAN} \\
        & Zhang T. et al. \citep{Zhang20233DCBAMs} & \url{https://github.com/vevvve/SASGAN} \\
        & Xia P. et al. \citep{Xia2022Multi-scaleNetworks} & \url{https://github.com/FPXMU/MS-GAN} \\
        & You N. et al. \citep{You20213DGAN} & \url{https://doi.org/10.5281/zenodo.4437911} \\
        & Ugolkov et al. \citep{Ugolkov2025OptimizedTomography} & \url{https://github.com/EvgenyUgolkov/x16-Octree-Bassed-Super-Resolution} \\
        \hline
        Conditional GAN
        & Sadeghkhani A. et al. \citep{Sadeghkhani2025PCP-GAN:Networks} & \url{https://github.com/AliSadeghkhani1990/PCP-GAN} \\
        \hline
        Attention-Enhanced GAN & Zhang T. et al. \citep{Zhang20233DCBAMs} & \url{https://github.com/vevvve/SASGAN} \\
        \hline
        \multirow{4}{2cm}{Style-based GAN} 
        & Thakre S. et al. \citep{Thakre2023QuantificationModels} & \url{https://github.com/vir-k01/StyleGAN2-ADA-MicrostructureGeneration} \\
        & Cao D. et al. \citep{Cao2022ReconstructionGAN} & \url{https://github.com/Berthou817/CISGAN} \\
        & Liu M. et al. \citep{Liu2022MultiscaleNetworks} & \url{https://github.com/theanswer003/MultiscaleDRPNet} \\
        & Fokina D. et al. \citep{Fokina2020MicrostructureNetworks} & Used StyleGAN TensorFlow implementation from \url{https://github.com/NVlabs/stylegan} \\
        \hline
        \multirow{3}{2cm}{Hybrid Architecture GAN} 
        & Chen H. et al. \citep{Chen2020Super-resolutionNetworks} & \url{https://github.com/III-SCU/SRCycleGAN} \\
        & Cao D. et al. \citep{Cao2022ReconstructionGAN} & \url{https://github.com/Berthou817/CISGAN} \\
        & Zhang Y. et al. \citep{Zhang2025ForADA-PGGAN} & \url{https://github.com/QUST-SmartData/ADA-PGGAN} \\
        \hline
    \end{tabular}
\end{table}
\FloatBarrier

\pagebreak

\section{Conclusions}
\label{sec:con}

This review has traced the evolution of GAN architectures for porous media reconstruction from foundational implementations to specialised systems addressing multi-scale generation, property control, and data scarcity. The analysis of 96 peer-reviewed articles published between 2017 and 2026 reveals that GAN-based reconstruction has matured into a field with quantitative capabilities, in volumetric scale, porosity accuracy, and permeability prediction, that validate these methods as practical alternatives to traditional imaging and statistical reconstruction approaches.

A central finding of this review is that each architectural paradigm resolves a specific limitation while introducing a distinct trade-off. Multi-scale architectures capture hierarchical features but face scale transition challenges. Conditional architectures enable property control but struggle with correlated properties. Attention mechanisms preserve structural coherence but impose quadratic memory costs. Style-based approaches offer data efficiency but remain limited to small 3D volumes. Hybrid architectures provide the most comprehensive solutions but at the cost of training complexity and extended convergence times. This recurring pattern of capability-versus-cost suggests that no single architecture is universally optimal; rather, the choice of GAN paradigm should be guided by the specific reconstruction requirements, available data, and computational resources of a given application.

Three cross-cutting challenges, discussed in detail in the preceding section, represent the primary barriers to broader adoption. These are (1) the integration of physics-based constraints directly into network architectures to ensure physical realisability beyond statistical similarity; (2) the establishment of standardised evaluation benchmarks enabling systematic comparison across methods and material types; and (3) the development of comprehensive uncertainty quantification frameworks providing confidence bounds essential for engineering applications. Addressing these challenges will require both architectural innovation within individual paradigms and community-wide coordination on datasets, metrics, and validation protocols.

The field trajectory suggests convergence toward hybrid architectures combining multiple innovations for comprehensive solutions. Recent successes with minimal training data \citep{Zhang2024DA-VEGAN:Sets}, multi-property control \citep{Nguyen2022SynthesizingLearning, Zhou20233DLearning}, and unpaired image scenarios \citep{Chen2020Super-resolutionNetworks, Wu2023AnFeatures} indicate a transition from research demonstrations to practical engineering tools. As computational resources become more accessible and architectures continue to improve in efficiency, GAN-based reconstruction is positioned to become a standard component of computational materials workflows, contingent upon resolving the current challenges in physics integration, standardisation, and uncertainty quantification.

\pagebreak

\section*{Author Declarations}
\subsection*{Funding}
A.S. was supported by a full PhD scholarship from the School of Computer Science, University of Leeds. No additional external funding was received for this research.

\subsection*{Conflicts of interest/Competing interests}
The authors declare that they have no known competing financial interests or personal relationships that could have appeared to influence the work reported in this paper.

\subsection*{Ethics approval}
This study is a systematic review of published literature and involved no experimental work, human participants, human tissue, or animal subjects. Therefore, no ethical approval was required.

\subsection*{Consent to participate}
Not applicable.

\subsection*{Consent for publication}
Not applicable.

\subsection*{Availability of data and materials}
This article is a systematic review. All data supporting the findings of this review are derived from previously published studies, which are cited and listed in the reference section. The compiled summary tables of all 96 reviewed papers (organised by GAN architecture category) are publicly available as Excel files in the supplementary data repository at: 
\url{https://github.com/AliSadeghkhani1990/GAN-Materials-Archive}. Detailed architectural specifications for all reviewed GAN implementations are provided 
in the Supplementary Material.

\subsection*{Code availability}
No new computer codes or software were developed in this study. A curated list of all open-source implementations identified in this review, organised by GAN architecture category, is publicly available at: \url{https://github.com/AliSadeghkhani1990/GAN-Materials-Archive}

\subsection*{Authors' contributions}
A.S. conducted the systematic literature search, categorised and analysed the reviewed studies, and wrote the original draft of the manuscript. B.B. provided supervision and contributed to the conceptualisation and critical revision of the manuscript. M.B. contributed to the interpretation of results and manuscript revision. A.R. conceived and supervised the project, contributed to the review methodology and scope definition, and contributed to writing and revising the manuscript. All authors read and approved the final manuscript.
\bibliography{references}



\end{document}


\title{Supplementary Material: A Decade of Generative Adversarial Networks 
for Porous Material Reconstruction}
\author{Ali Sadeghkhani, Brandon Bennett, Masoud Babaei, Arash Rabbani}

\maketitle

\section*{Appendix: Detailed Architecture Tables}

This appendix provides comprehensive architectural details and technical specifications for all GAN-based methods reviewed in this paper. Tables \ref{tab:gan_overview} through \ref{tab:hybrid_gan_overview} systematically document the evolution of GAN architectures for porous media reconstruction, organized by architectural category: Vanilla GANs (Table \ref{tab:gan_overview}), Multi-Scale GANs (Table \ref{tab:multiscale_gan_overview}), Conditional GANs (Table \ref{tab:conditional_gan_overview}), Attention-Enhanced GANs (Table \ref{tab:attention_gan_overview}), Style-based GANs (Table \ref{tab:style_gan_overview}), and Hybrid Architectures (Table \ref{tab:hybrid_gan_overview}). Each table presents detailed information including publication year, loss functions employed, training and output specifications (resolution and dimensionality), image types (binary, grayscale, multi-phase), material types studied, key novelties and innovations, evaluation metrics used for validation, and computational requirements. This systematic documentation enables direct comparison of methods within and across categories, facilitating informed selection of appropriate architectures for specific reconstruction tasks and providing a comprehensive reference for researchers building upon these approaches.

\begin{landscape}
    \tiny
    \raggedright  
    \setlength{\LTleft}{0pt}
    \setlength{\LTright}{0pt}
    \begin{longtable}{|>{\raggedright\arraybackslash}p{1.4cm}|>{\raggedright\arraybackslash}p{0.4cm}|>{\raggedright\arraybackslash}p{1.2cm}|>{\raggedright\arraybackslash}p{1.2cm}|>{\raggedright\arraybackslash}p{0.5cm}|>{\raggedright\arraybackslash}p{1.2cm}|>{\raggedright\arraybackslash}p{0.5cm}|>{\raggedright\arraybackslash}p{1cm}|>{\raggedright\arraybackslash}p{1.8cm}|>{\raggedright\arraybackslash}p{3cm}|>{\raggedright\arraybackslash}p{3cm}|>{\raggedright\arraybackslash}p{3cm}|}

        \caption{Summary of Vanilla GAN Architectures for Digital Porous Material Reconstruction}\label{tab:gan_overview}\\

        \hline
        \textbf{Reference} & \textbf{Year} & \textbf{Loss Function} & \multicolumn{2}{c|}{\textbf{Training Images}} & \multicolumn{2}{c|}{\textbf{Generated Images}} & \textbf{Image Type} & \textbf{Material} & \textbf{Novelty} & \textbf{Evaluation Metrics} & \textbf{Computational Requirements} \tabularnewline
        \hline
        & & & \textbf{Size} & \textbf{Dim} & \textbf{Size} & \textbf{Dim} & & & & & \tabularnewline
        \hline
        \endfirsthead

        \multicolumn{12}{c}{{\tablename\ \thetable{} -- continued from previous page}} \\
        \hline
        \textbf{Reference} & \textbf{Year} & \textbf{Loss Function} & \multicolumn{2}{c|}{\textbf{Training Images}} & \multicolumn{2}{c|}{\textbf{Generated Images}} & \textbf{Image Type} & \textbf{Material} & \textbf{Novelty} & \textbf{Evaluation Metrics} & \textbf{Computational Requirements} \tabularnewline
        \hline
        & & & \textbf{Size} & \textbf{Dim} & \textbf{Size} & \textbf{Dim} & & & & & \tabularnewline
        \hline
        \endhead

        \hline
        \multicolumn{12}{r}{{Continued on next page}} \\
        \endfoot

        \hline
        \endlastfoot

        Zhou Y. et al.\textsuperscript{\citep{Zhou2025DigitalSliceGAN}} & 2025 & WGAN & $383^2$ and $400^2$ pixels & 2D & $383^3$ voxels & 3D & Binary and Multi-phase (3-phase) & Tight carbonate rocks (Ordovician Majiagou Formation) and Berea sandstone & Application of SliceGAN to tight carbonate reconstruction with MAPS and QEMSCAN integration for multi-component mineral identification (dolomite, calcite), addressing complex pore structures and heterogeneity in tight carbonate reservoirs & Auto-correlation function
Two-point probability function
Minkowski functions
Porosity
Mineral content
Formation factor
Permeability (LBM) & Dataset: Berea 9 samples ($383^3$), Carbonate 7 MAPS images ($400^2$); GPU RTX 4090; CPU Xeon W-2195 @ 2.30 GHz; 136GB RAM; $\sim$2 hours; 20--30 epochs \\
        \hline

        Liu D. et al.\textsuperscript{\citep{Liu2025Three-dimensionalShales}} & 2025 & WGAN-GP & $614^2$ pixels & 2D & $384^3$ voxels & 3D & Multi-phase (4-phase) & Shale (Longmaxi Formation, Sichuan Basin) & Novel REV-size digital rock reconstruction workflow integrating high-resolution MAPS with OM watershed classification into three connectivity types based on surface porosity (Type A >20\%, B 10--20\%, C <10\%), demonstrating Type C OM dominates gas transport at REV scale & Porosity comparison
Pore diameter distributions
Coordination number distributions
Two-point correlation
Connectivity analysis
Area ratio stability & Dataset: 3 samples, 144 sub-images per sample ($614^2$); FIB-SEM validation $1600\times1600\times1000$ voxels; GPU RTX 4000 (8GB); CPU Intel i9-10900K; 64GB RAM; $\sim$24 hours; 50 epochs \\
        \hline

        Zhang Y. et al.\textsuperscript{\citep{Zhang2025PredictionModelling}} & 2025 & Hinge Loss & $64^3$ voxels & 3D & $512^3$ voxels & 3D & Binary & Bentheimer sandstone & DCGAN with $8^3$ upscaling combined with PNM for CO2 storage efficiency prediction and quantifying minimum simulations needed & Minkowski functionals
Two-point correlation
Morphological properties & Dataset: 6000 sub-volumes ($64^3$); GPU/training time not reported \\
        \hline

        Amiri H. et al.\textsuperscript{\citep{Amiri2024NewGANs}} & 2024 & WGAN-GP & $256^2$ pixels & 2D & $256^3$ voxels & 3D & Binary (2-phase) & Berea sandstone & Systematic 3D reconstruction from purely 2D images without 3D training data, enabling diverse reconstructions for uncertainty assessment & Two-point correlation functions
Fr\'{e}chet Inception Distance
Pore characteristics distribution
Permeability estimates & Dataset: Large 2D images (7.78--15.56mm BSE, 11.95--23mm optical); GPU RTX A6000 (48GB); 24 hours \\
        \hline

        Chi P. et al.\textsuperscript{\citep{Chi2024DigitalFramework}} & 2024 & WGAN-GP + Reconstruction Loss (MSE) & 3D LR: $128^3$ (S1), $512^3$ (S2) voxels; 2D HR: up to $4096^2$ pixels & 3D + 2D & Up to $512^3$ voxels (4--8$\times$ upsampling) & 3D & Grayscale and Binary & Tight sandstone & Novel 2D-3D fusion framework combining CT and SEM images
Lightweight 3D generator with 2D discriminator architecture
Direct high-resolution 3D generation with better continuity
4--8$\times$ super-resolution capability & Porosity
Fractal dimension
Pore structure parameters
Permeability (Lattice Boltzmann)
Autocorrelation function
Pore connectivity & Dataset: S1 (CT: 4$\mu$m HR/16$\mu$m LR); S2 (CT 8$\mu$m + SEM 2$\mu$m); GPU RTX 3060; $\sim$20 hours; Batch 16 \\
        \hline

        Chi P. et al.\textsuperscript{\citep{Chi2024ReconstructionNetwork}} & 2024 & WGAN-GP & $64^2$ pixels & 2D & Up to $1024^3$ voxels & 3D & Binary & Berea sandstone, clay, and shale samples & Large-scale anisotropic 3D from 2D using overlapping splitting and multiple discriminators for preserving anisotropy & Porosity distribution
Fractal dimension
Euler number
Two-point correlation functions
Flow velocity distribution
Pore connectivity
Permeability (LBM) & Dataset: 32,000 sub-samples per epoch; GPU RTX 3060; $\sim$5 hours; 10 epochs, batch 16 \\
        \hline

        Li X. et al.\textsuperscript{\citep{Li20243DLearning}} & 2024 & WGAN-GP & $64^2$ pixels & 2D & $64^3$ voxels & 3D & Binary & Porous ceramics (SOFC electrode materials and ceramic atomizer core materials) & 3D reconstruction from single 2D image using multi-criteria statistical controller with four two-point descriptors, validated with FIB-SEM & Phase volume fraction
Specific surface area
Tortuosity factor
Pore size distribution
Two-point correlation function & Dataset: Single 2D image; GPU 2$\times$ RTX 3090; $\sim$30 minutes \\
        \hline

        Xu W. et al.\textsuperscript{\citep{Xu2024Multi-criteriaProperties}} & 2024 & WGAN-GP & $64^2$ pixels & 2D & $64^3$ voxels & 3D & Binary & Cementitious materials (hardened cement paste, cellular concrete) & Multi-criteria DCGAN with four two-point descriptors for 3D reconstruction from single 2D image with transport properties prediction & Four two-point statistical functions
Absolute permeability
Effective thermal conductivity & Dataset: Single 2D image; CPU AMD EPYC 9754 (2.25 GHz); 60,000 iterations; 8GB RAM \\
        \hline

        Liu Y. et al.\textsuperscript{\citep{Liu2023ReconstructionDCGAN}} & 2023 & Wasserstein distance & $256^2$ pixels & 2D & $256^2$ pixels & 2D & Binary & Concrete & DCGAN for meso-scale concrete with advanced CT preprocessing, data augmentation, and novel fractal dimension evaluation metrics & Aggregate area fraction
Aggregate fractal dimension
Siamese network similarity
Stress-strain curve consistency & Dataset: 3,240 images (with augmentation); CPU Intel i7-11800H @ 2.3 GHz; GPU RTX 3060 (8GB DDR4); 15,400 epochs \\
        \hline

        Zhang T. et al.\textsuperscript{\citep{Zhang2023ANetwork}} & 2023 & WGAN-GP + Content Loss (MSE) + Feature Loss (VGG) & HR: $80^3$ voxels; LR: $40^3$ voxels & 3D & $80^3$ voxels (2$\times$ upsampling) & 3D & Binary & Shale & 3D super-resolution reconstruction method specifically for shale (3DRGAN)
Residual networks combined with GAN for enhanced feature learning
VGG-based perceptual feature loss for improved quality
Addresses field-of-view vs resolution trade-off
Validated through multiple statistical and flow simulation measures & Multi-point connectivity (MPC)
Porosity
Pore size distribution
Permeability (flow simulation)
Visual quality comparison & Dataset: 64 training images ($80^3$ voxels), 10 test images ($40^3$ voxels); GPU RTX 3080 (10GB); REV: $62\times65\times65$ voxels minimum \\
        \hline

        Zhang T. et al.\textsuperscript{\citep{Zhang2022AWGAN-GP}} & 2022 & WGAN-GP & $64^3$ voxels & 3D & $64^3$ voxels & 3D & Binary & Shale and Berea sandstone & Improved WGAN-GP with CNN replacing MLP and BatchNorm3d for enhanced 3D reconstruction with stabilized training & Multiple-point connectivity curves
Porosity accuracy
Autocorrelation functions
Permeability alignment & Dataset: 50 subsamples ($64^3$); CPU AMD R9 3900X 3.8 GHz; GPU RTX 2070 Super (8GB); 16GB RAM; 2000 epochs (25,000 iterations) \\
        \hline

        Li Y. et al.\textsuperscript{\citep{Li2022DigitalPenalty}} & 2022 & WGAN-GP & $64^2$ pixels & 2D & Up to $128^2$ pixels & 2D & Binary & Berea sandstone and Ketton limestone & Wasserstein distance with gradient penalty for stable training, demonstrating correlation between quality and discriminator loss & Two-point correlation functions
Minkowski functionals
Image similarity validation & Dataset: Berea $64\times64$ (8px stride), Ketton $64\times64$ (4px stride) with augmentation; GPU not reported; 10,000 gen. updates; Batch 64 \\
        \hline

        Sugiura K. et al.\textsuperscript{\citep{Sugiura2022HourlySliceGAN}} & 2022 & WGAN & $64^2$ pixels & 2D & $64^3$ voxels & 3D & Binary (2-phase) & Dual phase steels & Hourly 3D reconstruction (vs. daily serial sectioning) with comparable accuracy and consistent volume fractions & Volume fraction accuracy & GPU RTX 2070 Super; 30 epochs; 4h generation time \\
        \hline

        Zhang Q. et al.\textsuperscript{\citep{Zhang2022DigitalNetworks}} & 2022 & BCE Loss & $64^3$ voxels & 3D & $64^3$ voxels & 3D & Binary (2-phase) & Berea sandstone samples & Pretrained discriminator reducing training to 800 iterations (vs. thousands) with automated 3D reconstruction workflow & Porosity accuracy
Specific surface area
Two-point correlation functions & Dataset: Pre: 300 samples (300 iter.), Main: 1331 samples (800 iter.); Batch: Pre 16, Main 32; GPU not reported \\
        \hline

        Liu X. et al.\textsuperscript{\citep{Liu20223DLearning}} & 2022 & Not explicitly specified & $512^2$ pixels & 2D & $512\times512\times600$ voxels & 3D & Binary (2-phase) & Fuel cell catalyst layer & Latent space interpolation combining 2D DCGAN with two interpolation methods for spatial-continuous 3D structures & Phase ratio accuracy
Porosity
Pore size distribution
Particle size distribution
Tortuosity
Relative diffusion coefficient
FID score & Dataset: 600 FIB-SEM images; GPU/training time not reported \\
        \hline

        Zhang H. et al.\textsuperscript{\citep{Zhang2021ScaffoldGAN:Networks}} & 2021 & Combined Loss (Adversarial + Gram + Structural) & $64^3$ voxels & 3D & Up to $420^3$ voxels & 3D & Binary & Bone scaffolds (vertebral bone samples) and metal foams (Ni/Cu/Al foams) & Novel structural loss term for spatial coherence enabling end-to-end 3D scaffold synthesis with combined GAN, gram, and structural loss & Porosity error
Two-point correlation function
Connectivity metrics
Physical properties matching & Dataset: $\sim$6,000 samples ($64^3$, stride 2); CPU Intel i7-6700; GPU GTX 1080; 32GB RAM; $\sim$4 hours; Batch 100 \\
        \hline

        Hsu T. et al.\textsuperscript{\citep{Hsu2021MicrostructureMaterials}} & 2021 & WGAN & $96^3$ voxels & 3D & $96^3$ voxels & 3D & Grayscale & Solid oxide fuel cell (SOFC) anode containing yttria-stabilized zirconia (YSZ), nickel (Ni), and pore phases & First practical-scale 3D multiphase microstructure for ICME with high fidelity in visual, statistical, and electrochemical properties & Volume fractions
Particle sizes
Tortuosity factors
Interfacial surface area
Formation factors
TPB density
Exchange current density
Ohmic resistance
Active TPB density & Dataset: 323 subvolumes; GPU 64$\times$ Tesla P100; 39 hours (32,565 iterations) \\
        \hline

        Kench S. et al.\textsuperscript{\citep{Kench2021GeneratingExpansion}} & 2021 & WGAN-GP & $64^2$ pixels & 2D & $64^3$ voxels & 3D & Multi-phase (2-3-phase), RGB, Grayscale & Polycrystalline grains, ceramics, carbon fiber rods, battery separators, steel, and battery cathodes & SliceGAN generating 3D from 2D slices using uniform information density, enabling arbitrary large volumes for isotropic/anisotropic materials & Statistical metrics
Diffusivity
Two-point correlation functions
Triple-phase boundary densities & GPU Titan Xp; $\sim$4 hours \\
        \hline

        Zhao J. et al.\textsuperscript{\citep{Zhao20213DLearning}} & 2021 & BCE & $32^3$ voxels & 3D & $32^3$ voxels & 3D & Binary & Tight sandstone & DCGAN with 3D convolution for tight sandstone digital rock reconstruction, first application to tight sandstone & Pore size distributions
Spatial correlations
Two-point probability functions
Pore connectivity and geometry & Dataset: 910 CT slices; GPU GTX 1050; 20--30 min; 6000 epochs \\
        \hline

        Samaei A.\textsuperscript{\citep{Samaei2021MechanicalFEM}} & 2021 & BCE Loss & $64^3$ voxels & 3D & $64^3$ voxels & 3D & Binary & Zirconia-silica bilayer coating on aluminum alloys & GAN-based porous microstructure generation combined with mechanics of structure genome for multiscale modeling framework & Two-point correlation functions
Directional covariance
Visual inspection
Load-displacement curves
Stress-strain behavior
Elastic moduli & Dataset: 5203 images ($64^3$); D:G training ratio 3:1; GPU/time not reported \\
        \hline

        Sciazko A.\textsuperscript{\citep{Sciazko2021UnsupervisedImage}} & 2021 & Not explicitly specified & $64^2$ pixels & 2D & $64^3$ voxels & 3D & Multi-phase (3-phase) & Ni-GDC SOFC anodes & Novel 3D generator with 2D discriminator architecture, using only 2D training images to generate validated 3D structures & Volume fractions
Visual quality
TPB densities
Microstructural parameters & Dataset: 8 images ($\sim$2,500 $\mu$m$^2$), 192 patches/iter.; GPU RTX 3090 \\
        \hline

        Zha W. et al.\textsuperscript{\citep{Zha2020ReconstructionPenalty}} & 2020 & WGAN-GP & $256^2$ pixels & 2D & $256^2$ pixels & 2D & Grayscale & Shale & WGAN-GP with gradient penalty for improved training stability and quality of generated shale samples & FID Score
KID Score
Two-point correlation functions & Dataset: 10,000 images; GPU GTX 1080 Ti; 4 hours \\
        \hline

        Valsecchi A. et al.\textsuperscript{\citep{Valsecchi2020StochasticGANs}} & 2020 & Not explicitly specified & $64^3$ voxels & 3D & $64^3$ voxels & 3D & Binary (2-phase) & Beadpack, Berea sandstone, and Ketton limestone samples & Improved 3D-to-3D GAN reconstruction architecture with enhanced quality results for porous media & Two-point correlation functions
Minkowski functionals & Dataset: Beadpack: 55,296, Berea: 24,300, Ketton: 3,888; 8000 epochs ($\sim$8 hours); GPU not reported \\
        \hline

        Coiffier G.\textsuperscript{\citep{Coiffier20203DNetworks}} & 2020 & WGAN-GP & $64^2$ pixels & 2D & $64^3$ voxels & 3D & Binary & Packed spheres, folded layers, sand grains, channels, and sedimentary deposits & DiAGAN generating 3D from 2D using random cut sampling with periodic boundaries and arbitrary volume sizes without retraining & Indicator variograms
Connectivity functions
Fr\'{e}chet Inception Distance & GPU Tesla K40C (12GB RAM); Hours to days \\
        \hline

        Gayon-Lombardo A.\textsuperscript{\citep{Gayon-Lombardo2020PoresBoundaries}} & 2020 & BCE Loss with Label Smoothing & $64^3$ voxels & 3D & $64^3$ voxels (minimum) up to arbitrary large sizes & 3D & Multi-phase (3-phase) & Li-ion battery cathode and SOFC anode & First multi-phase 3D microstructure generation with periodic boundaries and arbitrary large volume capability & Volume fraction
Specific surface area
Triple-phase boundary density
Relative diffusivity
Two-point correlation function & Dataset: >10,000 sub-volumes; GPU Titan Xp; 72 epochs ($\sim$48 hours) \\
        \hline

        Valsecchi A. et al.\textsuperscript{\citep{Valsecchi2020StochasticNetworks}} & 2020 & BCE Loss & $64^2$ pixels & 2D & $64^3$ voxels & 3D & Binary (2-phase) & Beadpack, Berea sandstone, and Ketton limestone samples & First 2D-to-3D GAN reconstruction for porous media with novel discriminator working on 2D cross-sections of 3D volumes & Two-point probability function
Minkowski functionals & Dataset: Beadpack: 55,296, Berea: 24,300, Ketton: 3,888; GPU Titan X (12GB); $\sim$8 hours; 20--30ms gen. \\
        \hline

        Liu S. et al.\textsuperscript{\citep{Liu2019ANetworks}} & 2019 & BCE Loss & $64^3$ voxels for Berea, $128^3$ voxels for Estaillades & 3D & Up to $528^3$ voxels & 3D & Binary (2-phase) & Berea sandstone and Estaillades carbonate & 3D rotational data augmentation for improved training with sub-sampling strategy for homogeneous and heterogeneous samples & Visual and structural agreement
Training stability metrics & GPU GTX 1080 Ti; $\sim$3,000 iterations \\
        \hline

        Mosser L. et al.\textsuperscript{\citep{Mosser2018StochasticNetworks}} & 2018 & BCE Loss & $64^3$ voxels & 3D & Up to $450^3$ voxels & 3D & Grayscale & Oolitic Ketton limestone & First GAN application to unsegmented grayscale micro-CT data with Minkowski functionals for morphological validation & Minkowski functionals
Two-point probability functions
Velocity distributions
Permeability
Kolmogorov-Smirnov test & Dataset: 5,832 sub-volumes ($64^3$); GPU 8$\times$ K40; 8 hours (170 epochs); Batch 64 \\
        \hline

        Mosser L. et al.\textsuperscript{\citep{Mosser2017ReconstructionNetworks}} & 2017 & BCE Loss & $64^3$ voxels & 3D & Up to $350^3$ voxels & 3D & Binary (2-phase) & Beadpack, Berea sandstone, and Ketton limestone samples & First GAN application for 3D microstructure reconstruction with efficient generation of larger samples than training images & Two-point correlation functions
Minkowski functionals
Directional permeability
Mean chord lengths
Interface morphology & Dataset: Beadpack: 1,727, Berea: 10,647, Ketton: 15,624; 24 hours; GPU not specified \\
        \hline
    \end{longtable}
\end{landscape}

\begin{landscape}

    \tiny
    \raggedright  
    \setlength{\LTleft}{0pt}
    \setlength{\LTright}{0pt}
    \begin{longtable}{|>{\raggedright\arraybackslash}p{1.4cm}|>{\raggedright\arraybackslash}p{0.4cm}|>{\raggedright\arraybackslash}p{1.2cm}|>{\raggedright\arraybackslash}p{1.2cm}|>{\raggedright\arraybackslash}p{0.5cm}|>{\raggedright\arraybackslash}p{1.2cm}|>{\raggedright\arraybackslash}p{0.5cm}|>{\raggedright\arraybackslash}p{1cm}|>{\raggedright\arraybackslash}p{1.8cm}|>{\raggedright\arraybackslash}p{3cm}|>{\raggedright\arraybackslash}p{3cm}|>{\raggedright\arraybackslash}p{3cm}|}
    
        \caption{Summary of Multi-Scale GAN Architectures for Digital Porous Material Reconstruction}\label{tab:multiscale_gan_overview}\\
        
        \hline
        \textbf{Reference} & \textbf{Year} & \textbf{Loss Function} & \multicolumn{2}{c|}{\textbf{Training Images}} & \multicolumn{2}{c|}{\textbf{Generated Images}} & \textbf{Image Type} & \textbf{Material} & \textbf{Novelty} & \textbf{Evaluation Metrics} & \textbf{Computational Requirements} \tabularnewline
        \hline
        & & & \textbf{Size} & \textbf{Dim} & \textbf{Size} & \textbf{Dim} & & & & & \tabularnewline
        \hline
        \endfirsthead
        
        \multicolumn{12}{c}{{\tablename\ \thetable{} -- continued from previous page}} \\
        \hline
        \textbf{Reference} & \textbf{Year} & \textbf{Loss Function} & \multicolumn{2}{c|}{\textbf{Training Images}} & \multicolumn{2}{c|}{\textbf{Generated Images}} & \textbf{Image Type} & \textbf{Material} & \textbf{Novelty} & \textbf{Evaluation Metrics} & \textbf{Computational Requirements} \tabularnewline
        \hline
        & & & \textbf{Size} & \textbf{Dim} & \textbf{Size} & \textbf{Dim} & & & & & \tabularnewline
        \hline
        \endhead
        
        \hline
        \multicolumn{12}{r}{{Continued on next page}} \\
        \endfoot
        
        \hline
        \endlastfoot
        
        Ugolkov E. et al.\textsuperscript{\citep{Ugolkov2025OptimizedTomography}} & 2025 & WGAN-GP with MSE Reconstruction & 3D LR: $32^3$ voxels; 2D HR LSM: 5 stages from $64^2$ to $1024^2$ pixels & 3D + 2D & $512^3$ voxels per sub-volume (up to $4096^3$ via stitching) & 3D & Segmented (4-phase) & Berea sandstone & Memory-efficient octree-based progressive growing DC WGAN-GP achieving 16× super-resolution (7 µm to 0.44 µm/voxel) using Minkowski Engine for sparse convolutions
Uses unpaired 3D micro-CT and 2D LSM images for training
Corrects segmentation inaccuracies in minerals with similar X-ray attenuation
Generates arbitrary-sized images via sub-volume stitching with 50\% overlap
Addresses cubic memory scaling limitation of standard Conv3D & Volume fraction
Relative surface area
Two-point correlation function
Pore size distribution (PSD)
Pore network extraction (PNE) & Dataset: 3D micro-CT volume $1192 \times 1266 \times 1540$ voxels (7 µm/voxel), 2D HR LSM images (0.44 µm/pixel)
GPU: Two NVIDIA A100 80GB
RAM: 450 GB
Training: 71 hours total (100 epochs per stage, 5 stages)
Batch size: 2 \\
\hline
        Zhao X. et al.\textsuperscript{\citep{Zhao20253DNetworks}} & 2025 & WGAN-GP & Variable: $200^2$ to $800^2$ pixels & 2D & Variable: $200^3$ to $800^3$ voxels & 3D & Grayscale & Hardened cement paste & Multi-scale learning strategy capturing cement microstructure at various scales
Block-wise synthesis for generating large-size 3D images
Slicing strategy for cross-dimensional 2D-3D correlation
Successfully handles complex multi-phase cement microstructures
Single 3D generator with multiple 2D discriminators at different scales & Gray level histogram
Phase proportions (pores, hydration products, unreacted clinkers)
Pore size distribution (PSD) & Dataset: Single 2D BSE images at various resolutions
GPU: NVIDIA A100 Tensor Core
CPU: Intel Xeon Silver 4214R
Training: 15.8 hours \\

\hline

        Zhu L. et al.\textsuperscript{\citep{Zhu2024GenerationNetworks}} & 2024 & WGAN-LP with Lipschitz Penalty & $64^3$ and $128^3$ voxels & 3D & Up to $2,200^3$ voxels & 3D & Binary & Sandstone (Berea, Bentheimer, Mt. Simon) and carbonate (Estaillades, Savonnières) & Laplacian pyramid generator with WGAN-LP for multi-scale feature capture and large-scale generation using FSMR-enhanced discriminator & Porosity
Absolute permeability
Curvature
Specific surface area
Euler characteristic
Two-point correlation functions
Pore shape factors & GPU: NVIDIA Quadro RTX 5000
CPU: Dual Intel Xeon Gold 5222
RAM: 64GB
Training: 800 iterations max
Generation time: 2.47-5.62s ($300^3$), ~5h ($2,200^3$) \\
        \hline
        
        Zhang T. et al.\textsuperscript{\citep{Zhang20233DCBAMs}} & 2023 & WGAN-GP with L2 Reconstruction & HR: $80^3$ voxels, LR: $40^3$ voxels (6 scales: $16^3$ to $80^3$) & 3D & $80^3$ voxels & 3D & Binary & Sandstone and Shale & Combines CBAM attention mechanisms and residual blocks with SinGAN for 3D super-resolution from single low-resolution training image & Pore space analysis
Absolute permeability in X, Y, Z directions
Multi-point connectivity curves
Autocorrelation functions
Pore/throat size distributions and coordination number
& GPU: GeForce RTX3080
CPU: AMD R9 3900X 3.8 GHz
RAM: 16GB
Training time: 3000 epochs per scale, ~6 hours for first reconstruction \\
        \hline
        
        Li Y. et al.\textsuperscript{\citep{Li2023DigitalNetwork}} & 2023 & WGAN-GP with L2 Reconstruction & Shale: 2 slices from $120^3$ voxel, Berea: 2 slices from $100^3$ voxel & 2D & Shale: $120^3$, Berea: $100^3$ & 3D & Binary (2-phase) & Shale (anisotropic), Berea Sandstone (isotropic) & Combines MCMC with GAN for 3D reconstruction from only 2 training images, addressing connectivity issues in 2D-to-3D methods & Two-point correlation function in x, y, z directions
Porosity
Minkowski functionals & Hardware: Not specified
Training time: Berea sandstone ($100^3$): 0.32h; Shale ($100^3$): 0.82h \\
        \hline
        
        Zhang T. et al.\textsuperscript{\citep{Zhang2023ReconstructionNetworks}} & 2023 & Combined Loss (WGAN-GP + L2 Reconstruction + Porosity) & $64^3$ voxels (5 scales: $16^3$ to $64^3$) & 3D & $64^3$ voxels & 3D & Binary (2-phase) & Shale (anisotropic), Beadpack (isotropic) & Multi-scale GAN with convolutional pyramid structure learning from single 3D image, with reusable model for fast subsequent reconstructions & Variogram curves in X, Y, Z
Multiple-point connectivity
Lattice Boltzmann permeability
Pore size distributions
& GPU: RTX 2080 Ti
CPU: Intel Core i7-7800X
RAM: 24 GB
Training time: 36,905 seconds (first reconstruction), 6 seconds (subsequent) \\
        \hline
        
        Zhang T. et al.\textsuperscript{\citep{Zhang2023StochasticMechanisms}} & 2023 & Combined Loss (Hinge + L2 Reconstruction + Diversity) & $64^3$ voxels (5 scales: $4^3$ to $64^3$) & 3D & $64^3$ voxels & 3D & Binary (2-phase) & Shale & Multi-scale GAN with CBAM attention mechanisms for single-image training with one-stage approach versus traditional multi-stage methods & Multi-point connectivity curves
Pore space distribution
Permeability matching
Feature diversity & GPU: NVIDIA RTX 3080
CPU: AMD Ryzen 9 3900X
RAM: 16GB
Training time: Initial ~5,863s, Subsequent ~6s
Training rounds: 30,000
Memory usage: 29\% RAM, 95\% GPU, 70\% GPU memory \\
        \hline
        Zhang T. et al.\textsuperscript{\citep{Zhang2023StochasticNetwork}} & 2023 & Dual Discriminator Adversarial Loss + Reconstruction Loss (L2) & $64^2$ pixels & 2D & $256^2$ pixels & 2D & Binary & Sandstone & Multi-stage GAN with CBAM attention mechanisms for single-image training
Dual discriminator architecture balancing KL and reverse KL divergences for diversity preservation
Progressive multi-scale learning from coarse to fine features
Parameter sharing between consecutive stages reducing training time
Significantly improved efficiency over SinGAN while maintaining quality & Porosity
Pore/throat number and radii distributions
Two-point correlation function
Lineal-path function
Multi-point connectivity (MPC)
Euler number
Permeability & Dataset: Single training image
GPU: NVIDIA RTX 3080 (10GB)
CPU: Intel Core i5-8250U (1.60 GHz)
RAM: 16GB
Training: 3000 iterations per stage, 8-10 stages total \\

\hline
        
        Zhang T. et al.\textsuperscript{\citep{Zhang2023Super-ResolutionBlocks}} & 2023 & WGAN-GP with L2 Reconstruction & 6 scales: $16^3$ to $80^3$ & 3D & $80^3$ voxels & 3D & Binary & Sandstone and Shale & Concurrent multi-scale training with residual blocks for super-resolution from single low-resolution image with reusable model & Porosity distribution
Connected/isolated pore space ratios
Permeability matching
Multi-point connectivity curves & GPU: RTX3080
CPU: AMD R9 3900X
RAM: 16GB
Training time: 3000 epochs per stage \\
        \hline
        
        Shi X. et al.\textsuperscript{\citep{Shi20233D-porous-GAN:Image}} & 2023 & WGAN-GP with L2 Reconstruction & Progressive scales: $16^3$ to $80^3$ & 3D & $80^3$ voxels, $64^3$ voxels & 3D & Binary & Sandstone and Shale & Multi-scale concurrent training with progressive growing architecture, residual connections between stages, and gradient penalty for stability & Porosity matching
Pore connectivity
Two-point correlation
Minkowski functionals & Hardware: Not specified
Training time: Not specified \\
        \hline
        
        Xia P. et al.\textsuperscript{\citep{Xia2022Multi-scaleNetworks}} & 2022 & WGAN-GP & 6 stages: $4^3$ to $128^3$ & 3D & $128^3$ voxels, $64^3$ voxels & 3D & Binary (2-phase) & Sandstone and Shale & Progressive growing architecture with sequential multi-scale training using gradual layer integration and scale transition via trilinear interpolation & Porosity Analysis
Multi-point Connectivity
Pore Network Analysis
Permeability Analysis & GPU: NVIDIA GeForce RTX 3080 (10 GB)
CPU: Intel Core i5-6200U (2.40 GHz)
RAM: 8 GB
Training: $4 \times 10^4$ epochs \\
        \hline
        
        You N. et al.\textsuperscript{\citep{You20213DGAN}} & 2021 & WGAN-GP with L2 Reconstruction & $1024^2$ pixels & 2D & $1720 \times 1024 \times 1024$ voxels & 3D & Grayscale CT scans & Estaillades carbonate rock & Progressive growing GAN achieving 9x speedup and 4,587x data compression by needing only every 9th slice for large-scale reconstruction & Effective porosity comparison
Pore network analysis
Visual assessment metrics
Data compression metrics
Pixel-wise reconstruction error
Pore radius distribution & GPU: Tesla V100 (32 GB)
Training time: 243 hours
Model parameters: Generator 18.3M, Discriminator 23.1M \\
        \hline
    \end{longtable}
\end{landscape}

\begin{landscape}

    \tiny
    \raggedright  
    \setlength{\LTleft}{0pt}
    \setlength{\LTright}{0pt}
    \begin{longtable}{|>{\raggedright\arraybackslash}p{1.4cm}|>{\raggedright\arraybackslash}p{0.4cm}|>{\raggedright\arraybackslash}p{1.2cm}|>{\raggedright\arraybackslash}p{1.2cm}|>{\raggedright\arraybackslash}p{0.5cm}|>{\raggedright\arraybackslash}p{1.2cm}|>{\raggedright\arraybackslash}p{0.5cm}|>{\raggedright\arraybackslash}p{1cm}|>{\raggedright\arraybackslash}p{1.8cm}|>{\raggedright\arraybackslash}p{3cm}|>{\raggedright\arraybackslash}p{3cm}|>{\raggedright\arraybackslash}p{3cm}|}
    
        \caption{Summary of Conditional GAN Architectures for Digital Porous Material Reconstruction}\label{tab:conditional_gan_overview}\\
        
        \hline
        \textbf{Reference} & \textbf{Year} & \textbf{Loss Function} & \multicolumn{2}{c|}{\textbf{Training Images}} & \multicolumn{2}{c|}{\textbf{Generated Images}} & \textbf{Image Type} & \textbf{Material} & \textbf{Novelty} & \textbf{Evaluation Metrics} & \textbf{Computational Requirements} \tabularnewline
        \hline
        & & & \textbf{Size} & \textbf{Dim} & \textbf{Size} & \textbf{Dim} & & & & & \tabularnewline
        \hline
        \endfirsthead
        
        \multicolumn{12}{c}{{\tablename\ \thetable{} -- continued from previous page}} \\
        \hline
        \textbf{Reference} & \textbf{Year} & \textbf{Loss Function} & \multicolumn{2}{c|}{\textbf{Training Images}} & \multicolumn{2}{c|}{\textbf{Generated Images}} & \textbf{Image Type} & \textbf{Material} & \textbf{Novelty} & \textbf{Evaluation Metrics} & \textbf{Computational Requirements} \tabularnewline
        \hline
        & & & \textbf{Size} & \textbf{Dim} & \textbf{Size} & \textbf{Dim} & & & & & \tabularnewline
        \hline
        \endhead
        
        \hline
        \multicolumn{12}{r}{{Continued on next page}} \\
        \endfoot
        
        \hline
        \endlastfoot
        
        Torben J. et al.\textsuperscript{\citep{Torben2025ControllableNetworks}} & 2025 & CCGAN Loss (reformulated empirical cGAN) & $128^2$ pixels & 2D & $128^2$ pixels, $128 \times 128 \times 901$ voxels & 2D/3D & Binary (2-phase) & Porous Asphalt & CCGAN-based framework with continuous porosity regression label, user-defined vertical porosity profile generation, latent space interpolation for cohesive 3D volume synthesis from 2D slices, integrated segmentation via SoftMax and one-hot encoding, optimal interpolation step determination via RMSE minimization & Porosity profile control
Mean correlation between slices
RMSE between correlation curves
Material properties: porosity, surface area, effective diffusivity
TauFactor metrics
Statistical analysis (50 volumes) & GPU: NVIDIA 3090 RTX Ti
Dataset: 32,000 patches from 2000³ voxel CT scan (cropped to $1000 \times 1000 \times 901$)
Generation time: 0.7s per volume
Training time: Not specified \\
\hline

Sadeghkhani A. et al.\textsuperscript{\citep{Sadeghkhani2025PCP-GAN:Networks}} & 2025 & Binary Cross Entropy Loss & $480^2$ pixels & 2D & $480^2$ pixels & 2D & Colour (RGB) & Carbonate formations (Dolomite, Dolomite-Anhydrite) & Multi-conditional GAN with dual constraints (porosity + depth), RGB thin section processing preserving petrographic information, superior representativeness: 1.9-11.3\% error vs. 36.4-578\% for real sub-images, automated framework with REV analysis and U-Net & Porosity Control,
Morphological parameters: pore radius, specific surface area, tortuosity
Dual-property performance: 1.9-11.3\% vs. 36.4-578\% 
Statistical tests: KS-test, t-test, Cohen's d
Visual quality and architecture comparison & GPU: NVIDIA A2 (15 GB)
CPU: AMD EPYC 9354P 32-Core
Dataset: $\sim$6,400 images (1,600 per sample $\times$ 4 samples)
Training time: 200 epochs \\
\hline

        Zhou R. et al.\textsuperscript{\citep{Zhou20233DLearning}} & 2023 & WGAN-GP with Porosity Feature Loss & $64^3$ voxels & 3D & $64^3$ voxels & 3D & Binary (2-phase) & Berea sandstone & Multi-conditional GAN with petrophysical parameters, dynamic penalty feature loss, and switchable conditioning & Two-point correlation function
Statistical distribution of petrophysical parameters
Porosity
Fractal dimension
Specific surface area
Tortuosity & GPU: NVIDIA GeForce RTX 3080
CPU: Intel Core i9-12900K
Dataset: 3,375 samples
Training time: 61 hours \\
        \hline
        
        Kishimoto M. et al.\textsuperscript{\citep{Kishimoto2023ConditionalFractions}} & 2023 & cGAN Adversarial Loss + Volume Fraction Loss (MSE) & $64^3$ voxels & 3D & $64^3$ to $256^3$ voxels & 3D & Multi-phase (3-phase) & Ni-YSZ anode for SOFC & Volume fraction control via conditional input, fully convolutional for expandable outputs, extrapolation beyond training & Volume fraction control
Specific surface area
TPB density & GPU: NVIDIA GeForce RTX 3090
Dataset: 3,000 structures per composition
Training time: 100 epochs
Batch size: 32 \\
        \hline
        
        Kishimoto M. et al.\textsuperscript{\citep{Kishimoto2023ConditionalDataset}} & 2023 & Adversarial Loss + Volume Fraction Loss (MSE) & $64^3$ voxels & 3D & $64^3$ voxels & 3D & Multi-phase (3-phase) & SOFC anode (Ni-YSZ) & Conditional vector with volume fraction loss for generating compositions outside training distribution & Volume fraction control
Specific surface area
TPB density
Volume fraction distributions & GPU: NVIDIA GeForce RTX 3090
Dataset: 3,000 structures total
Training time: 100 epochs
Batch size: 32 \\
        \hline
        
        Chi P. et al.\textsuperscript{\citep{Chi2023ReconstructionCVAE-GAN}} & 2023 & Combined Loss (VAE Reconstruction + KL Divergence + Porosity) with WGAN-GP & $128^3$ voxels & 3D & $128^3$ voxels & 3D & Binary & Fontainebleau and Bentheimer sandstones & CVAE-GAN hybrid with integrated porosity control, extrapolation capability, real and random label training & Two-point correlation function
Pore radius distribution
Formation factor
Wave velocity (VP and VS)
Porosity control accuracy & Dataset: Fontainebleau: 1,512 samples, Bentheimer: 1,000 samples
Hardware: Not specified \\
        \hline
        
        Nguyen P.C.H. et al.\textsuperscript{\citep{Nguyen2022SynthesizingLearning}} & 2022 & Standard GAN adversarial loss & $128^3$ voxels & 3D & $128^3$ voxels & 3D & Binary (2-phase) & Bentheim sandstone & GAN combined with actor-critic reinforcement learning for targeted property control via multi-stage training & Kullback-Leibler divergence (KLD)
Maximum mean discrepancy (MMD)
Earth mover's distance (EMD)
Porosity
Specific surface area
Permeability
Degree assortativity
Graph transitivity
Graph density
Average clustering
Efficiency & Dataset: Bentheim sandstone $\mu$CT scans
Batch size: 64
Training: 1,000 episodes (AC stage)
Hardware: Not specified \\
        \hline
        
        Yang Y. et al.\textsuperscript{\citep{Yang2022Multi-scaleNetworks}} & 2022 & cGAN adversarial loss + L1 distance loss & $256^3$ voxels & 3D & $256^3$ voxels & 3D & Binary (2-phase) & Carbonate rock & Multi-scale from single low-resolution input without superposition algorithms or dual-resolution requirement & Two-point correlation function
Porosity
Pore radius
Throat radius
Throat length
Shape factor
Tortuosity
Coordination number
Euler characteristic
Permeability & GPU: NVIDIA K80
Dataset: 2,744 training images \\
        \hline
        
        Matsuda Y. et al.\textsuperscript{\citep{Matsuda2022FrameworkNetwork}} & 2022 & WGAN-GP& $256 \times 128 \times 128$ voxels & 3D & $256 \times 128 \times 128$ voxels & 3D & Binary & Random, packed-sphere, and sponge structured porous materials & CGAN structural hybridization with Bayesian optimization using material type intensity as design variable & Porosity
Specific surface area
Euler characteristics
Minkowski functionals
Pore size distributions
Filtration efficiency
Pressure drop
Hypervolume & GPU: NVIDIA RTX A6000 (48 GB)
CPU: Intel Core i9-10940X
Dataset: 15,000 samples
Training time: ~8 days
Batch size: 20 per material type
Total optimization time: ~30 hours \\
        \hline
        
        Zheng Q. et al.\textsuperscript{\citep{Zheng2022DigitalNetworks}} & 2022 & WGAN-GP & $64^3$ voxels & 3D & Up to $160^3$ voxels & 3D & Binary (2-phase) & Sandstones (Berea, Doddington), carbonates (Estaillade, Ketton), and multiscale sandy medium & Multi-property simultaneous conditioning, scalable outputs, multiple materials, high/low-order statistics control & Two-point correlation function
Specific surface area
Porosity distribution
Correlation length
Absolute permeability
Anisotropic permeability
Sliced Wasserstein Distance (SWD) & GPU: 4× Tesla V100
Dataset: 61,440 samples (5 rock types)
Training time: ~23 hours
Batch size: 32 \\
        \hline
        
        Shams R. et al.\textsuperscript{\citep{Shams2021AST-CGAN}} & 2021 & Adversarial Loss + L1 Analogy Loss & $64^2$ for Berea, $128^2$ for Estaillades & 2D & Up to $400^3$ voxels & 3D & Binary (2-phase) & Berea Sandstone and Estaillades Carbonate & Statistical-CGAN hybrid for 3D from single 2D image, handles heterogeneous media, 1000x faster than MPS & Auto-correlation function
Network parameters (throats, pores, radii, lengths)
Porosity
Permeability & GPU: NVIDIA GeForce GTX 1080 Ti (11 GB)
CPU: Intel i7-5820K
RAM: 32 GB
Dataset: Berea: 44,100 slices, Estaillades: 8,640 slices
Training time: 64³: 1.58h, 128³: 4.05h
Batch size: 32 \\
        \hline
        
        Tang J. et al.\textsuperscript{\citep{Tang2021MachineAlumina}} & 2021 & WGAN-GP & $128^2$ pixels & 2D & $128^2$ pixels & 2D & Grayscale SEM & Laser-sintered alumina & Regression-based continuous conditioning for unexplored parameters with Wasserstein loss and gradient penalty & Phase fraction & GPU: NVIDIA P100
CPU: 8 CPUs
Dataset: 2,080 images per condition
Training time: 5 hours
Batch size: 128 \\
        \hline
        
        Feng J. et al.\textsuperscript{\citep{Feng2019ReconstructionNetworks}} & 2019 & Four component loss: CGAN adversarial loss, L1 loss, pattern loss, porosity loss & $128^2$ pixels & 2D & $128^2$ pixels & 2D & Binary (2-phase) & Silica in rubber matrix, Battery material, Sandstone, Anisotropic porous media & Reconstruction from extremely limited input, novel pattern and porosity losses, multiple candidates, 0.08s speed & Porosity
Two-point correlation function
Lineal-path function
Chord-length distribution function
Two-point cluster function
Local porosity distribution & CPU: Intel i7-4790K
Dataset: 600-1,000 images per category
Batch size: 2
Reconstruction speed: ~0.08s per image \\
        \hline
        
        Feng J. et al.\textsuperscript{\citep{Feng2018AcceleratingLearning}} & 2018 & Adversarial Loss + Perceptual Loss + L1 Loss & $128^2$ pixels & 2D & $128^2$ pixels, $128^3$ voxels & Both 2D and 3D & Binary (2-phase) & Sandstone samples & First deep learning MPS acceleration: direct sampling-to-output translation, 760x (2D) and 25x (3D) speedup & Two-point correlation function (S2)
Lineal-path function (L)
Two-point cluster function (C2)
Local porosity distribution (LPD)
Pore radius
Throat radius
Volumes
Permeability & GPU: NVIDIA GTX 1080 Ti
Dataset: 1,000 training pairs
Training epochs: 200 \\
        \hline
    \end{longtable}
\end{landscape}

\begin{landscape}

    \tiny
    \raggedright  
    \setlength{\LTleft}{0pt}
    \setlength{\LTright}{0pt}
    \begin{longtable}{|>{\raggedright\arraybackslash}p{1.4cm}|>{\raggedright\arraybackslash}p{0.4cm}|>{\raggedright\arraybackslash}p{1.2cm}|>{\raggedright\arraybackslash}p{1.2cm}|>{\raggedright\arraybackslash}p{0.5cm}|>{\raggedright\arraybackslash}p{1.2cm}|>{\raggedright\arraybackslash}p{0.5cm}|>{\raggedright\arraybackslash}p{1cm}|>{\raggedright\arraybackslash}p{1.8cm}|>{\raggedright\arraybackslash}p{3cm}|>{\raggedright\arraybackslash}p{3cm}|>{\raggedright\arraybackslash}p{3cm}|}
    
        \caption{Summary of Attention-Enhanced GAN Architectures for Digital Porous Material Reconstruction}\label{tab:attention_gan_overview}\\
        
        \hline
        \textbf{Reference} & \textbf{Year} & \textbf{Loss Function} & \multicolumn{2}{c|}{\textbf{Training Images}} & \multicolumn{2}{c|}{\textbf{Generated Images}} & \textbf{Image Type} & \textbf{Material} & \textbf{Novelty} & \textbf{Evaluation Metrics} & \textbf{Computational Requirements} \tabularnewline
        \hline
        & & & \textbf{Size} & \textbf{Dim} & \textbf{Size} & \textbf{Dim} & & & & & \tabularnewline
        \hline
        \endfirsthead
        
        \multicolumn{12}{c}{{\tablename\ \thetable{} -- continued from previous page}} \\
        \hline
        \textbf{Reference} & \textbf{Year} & \textbf{Loss Function} & \multicolumn{2}{c|}{\textbf{Training Images}} & \multicolumn{2}{c|}{\textbf{Generated Images}} & \textbf{Image Type} & \textbf{Material} & \textbf{Novelty} & \textbf{Evaluation Metrics} & \textbf{Computational Requirements} \tabularnewline
        \hline
        & & & \textbf{Size} & \textbf{Dim} & \textbf{Size} & \textbf{Dim} & & & & & \tabularnewline
        \hline
        \endhead
        
        \hline
        \multicolumn{12}{r}{{Continued on next page}} \\
        \endfoot
        
        \hline
        \endlastfoot
        
        Chi P. et al.\textsuperscript{\citep{Chi2024MultiscaleNetwork}} & 2024 & GAN loss + cycle-consistency loss + identity preserving loss & Micro-CT: $64^2$ pixels, SEM: $1024^2$ pixels & 2D & $16000^2$pixels (stacked to 3D) & 3D & Grayscale & Tight Sandstone & Integration of attention mechanisms with GAN for multiscale fusion. Combines micro-CT and SEM images. Focus on clay morphology and micropore reconstruction & Two-point correlation functions, Representative volume element analysis, Pore radius distributions, Mineral content analysis, Quantitative segmentation analysis, Permeability measurements & GPU: NVIDIA GeForce RTX 3090
Dataset: 3,776 SEM images (1024×1024)
Training time: ~12 hours, 40 epochs
Learning rate: 0.0002 with decay \\
        \hline
        
        Zhang T. et al.\textsuperscript{\citep{Zhang20233DCBAMs}} & 2023 & WGAN-GP + L2 reconstruction loss & HR: $80^3$, LR: $40^3$ voxels & 3D & $80^3$ voxels & 3D & Binary & Sandstone and Shale & Combines attention mechanisms (CBAM) and residual blocks with SinGAN framework. Achieves 3D super-resolution from a single LR training image. Validated on both sandstone and shale samples & Total, connected, and isolated porosities, Absolute permeability (X, Y, Z directions), Multi-point connectivity curves, Autocorrelation functions, Pore/throat size distributions, Pore/throat numbers, Coordination number distributions, Diversity metrics (MS-SWD-MDS) & GPU: NVIDIA RTX 3080 (10GB)
CPU: AMD Ryzen 9 3900X
Dataset: 6 scales (16³ to 80³ voxels)
Training time: ~6h initial, 1s subsequent
Single-image training approach \\
        \hline
        
        Wang B. et al.\textsuperscript{\citep{Wang20233DStructure}} & 2023 & WGAN loss & $64^3$ voxels & 3D & $64^3$ voxels & 3D & Grayscale & Carbonate (Estaillades carbonate) & Integration of self-attention mechanism for carbonate rock reconstruction. Combines traditional and fully convolutional self-attention mechanisms. Specific focus on carbonate rock features. Enhanced ability to capture heterogeneity in pore distribution & Porosity accuracy, Grayscale value distribution, Mean value comparison & Dataset: 4,000 samples (augmented 4×)
Batch size: 48
Learning rate: 2×10$^{-4}$ \\
        \hline
        
        Wang P. et al.\textsuperscript{\citep{Pingquan2023ShaleEstimation}} & 2023 & Standard GAN logarithmic loss & $64^3$ voxels & 3D & $64^3$ voxels evaluated on $400^3$ voxel reconstructions  & 3D & Binary & Shale & Integration of context-aware modules. High-resolution optical flow estimation. Enhanced feature extraction for nano-micron pores & Two-point correlation function, Two-point cluster function, Linear path function, Chord length distribution function, Porosity metrics, Permeability metrics, Pore structure parameters, REV analysis & GPU: GTX 1080Ti
RAM: 32GB
Dataset: 10,648 hdf5 files
Training time: ~12 hours, 400 epochs
Batch size: 64 \\
        \hline
        
        He L. et al.\textsuperscript{\citep{He2023DigitalNetworks}} & 2023 & Wasserstein loss with gradient penalty (WGAN-GP) & $128^2$ pixels & 2D & $128^2$ pixels & 2D & Grayscale segmented & Shale & Introduction of residual self-attention block (RSAB). Combination of residual structure and self-attention mechanism. Enhanced feature correlation learning & Fréchet Inception Distance, Kernel Inception Distance, Two-Point Covariance Function, Minkowski Functions, Pore analysis metrics (number, diameter, distribution) & GPU: NVIDIA RTX 3060 (12GB)
Dataset: 100,000 samples (augmented 5×)
Training: 100,000 iterations
Batch size: 128 \\
        \hline
        
        Zhang T. et al.\textsuperscript{\citep{Zhang2023StochasticMechanisms}} & 2023 & Hinge-GAN with L2 reconstruction and diversity loss & $64^3$ voxels (5 scales: $4^3$ to $64^3$) & 3D & $64^3$ voxels & 3D & Binary (2-phase) & Shale & Combines multi-scale GAN with CBAM for 3D shale reconstruction. Single-image training approach. One-stage training with simultaneous multi-scale processing (vs ConSinGAN's progressive multi-stage). Hinge-GAN loss function to address gradient vanishing. Diversity loss function to prevent mode collapse & Multi-point connectivity curves (X/Y/Z), Permeability (X/Y/Z), PNM analysis: pore/throat numbers, radius distributions, coordination numbers, Difference Degrees (DD), Porosity (total, connected, isolated), MS-SWD-MDS diversity metrics & GPU: NVIDIA RTX 3080 (10GB)
Dataset: Single 64³ voxel image
Training time: ~1.6h initial, 6s subsequent
One-stage multi-scale training \\
        \hline
        
        Shan L. et al.\textsuperscript{\citep{Shan2022RockNetwork}} & 2022 & Adversarial loss + perceptual loss + content loss & $96^2$ pixels & 2D & $96^2$ pixels & 2D & Grayscale & Sandstone, Coal, Carbonate rock samples & Introduces residual dual-channel attention blocks in both generator and discriminator. Achieves improved reconstruction of high-frequency details and sharp edges & Peak Signal-to-Noise Ratio (PSNR), Structural Similarity Index (SSIM), Edge and texture detail analysis & GPU: 2× NVIDIA GTX 1060
CPU: Intel i9-9900KF
Dataset: 12,000 images (3 rock types)
RAM: 32GB \\
        \hline
    \end{longtable}
\end{landscape}

\begin{landscape}

    \tiny
    \raggedright  
    \setlength{\LTleft}{0pt}
    \setlength{\LTright}{0pt}
    \begin{longtable}{|>{\raggedright\arraybackslash}p{1.4cm}|>{\raggedright\arraybackslash}p{0.4cm}|>{\raggedright\arraybackslash}p{1.2cm}|>{\raggedright\arraybackslash}p{1.2cm}|>{\raggedright\arraybackslash}p{0.5cm}|>{\raggedright\arraybackslash}p{1.2cm}|>{\raggedright\arraybackslash}p{0.5cm}|>{\raggedright\arraybackslash}p{1cm}|>{\raggedright\arraybackslash}p{1.8cm}|>{\raggedright\arraybackslash}p{3cm}|>{\raggedright\arraybackslash}p{3cm}|>{\raggedright\arraybackslash}p{3cm}|}
    
        \caption{Summary of Style-based GAN Architectures for Digital Porous Material Reconstruction}\label{tab:style_gan_overview}\\
        
        \hline
        \textbf{Reference} & \textbf{Year} & \textbf{Loss Function} & \multicolumn{2}{c|}{\textbf{Training Images}} & \multicolumn{2}{c|}{\textbf{Generated Images}} & \textbf{Image Type} & \textbf{Material} & \textbf{Novelty} & \textbf{Evaluation Metrics} & \textbf{Computational Requirements} \tabularnewline
        \hline
        & & & \textbf{Size} & \textbf{Dim} & \textbf{Size} & \textbf{Dim} & & & & & \tabularnewline
        \hline
        \endfirsthead
        
        \multicolumn{12}{c}{{\tablename\ \thetable{} -- continued from previous page}} \\
        \hline
        \textbf{Reference} & \textbf{Year} & \textbf{Loss Function} & \multicolumn{2}{c|}{\textbf{Training Images}} & \multicolumn{2}{c|}{\textbf{Generated Images}} & \textbf{Image Type} & \textbf{Material} & \textbf{Novelty} & \textbf{Evaluation Metrics} & \textbf{Computational Requirements} \tabularnewline
        \hline
        & & & \textbf{Size} & \textbf{Dim} & \textbf{Size} & \textbf{Dim} & & & & & \tabularnewline
        \hline
        \endhead
        
        \hline
        \multicolumn{12}{r}{{Continued on next page}} \\
        \endfoot
        
        \hline
        \endlastfoot
        
        Li J. et al.\textsuperscript{\citep{Li2024Three-dimensionalNetwork}} & 2024 & WGAN-GP loss + pattern density loss & $128^2$ pixels & 2D & $128^3$ voxels & 3D & Binary (2-phase) & Sandstone, limestone, silica particles & End-to-end 3D reconstruction from single 2D image using EFPN-StyleGAN architecture
Multi-scale feature extraction addressing small pore localization
Three discriminators for orthogonal evaluation with pattern density constraint
Feature fusion architecture handling varying sensitivities at different scales & Connectivity parameters (porosity, effective porosity, connected pore volume ratio)
Morphological descriptors (two-point correlation S2, lineal-path function L, two-point cluster C2, local porosity distribution)
Microstructure parameters (pore/throat analysis: AP, AT, ASF, ASPR, ASTR, AVP, AVT, ARRSPT, ACN)
Effective permeability (EP) & Dataset: 16,000 2D images for each material type
Hardware: NVIDIA GeForce RTX 3090Ti (24GB VRAM), Intel Xeon Silver 4114, 128GB RAM
Batch size: 16 \\
        \hline
        
        Thakre S. et al.\textsuperscript{\citep{Thakre2023QuantificationModels}} & 2023 & Traditional GAN loss with binary cross-entropy & $256^2$ pixels & 2D & $256^2$ pixels & 2D & Binary (2-phase) & Dual Phase (DP) steels & First quantification of similarity and physical awareness of GAN-generated microstructures
Implementation of StyleGAN2 with ADA for microstructure generation
Novel evaluation framework combining statistical and physical metrics & Image similarity metrics (SSIM, MSE, PSNR, SNR)
Two-point correlation analysis
Volume fraction analysis
Phase volume ratios
ROM prediction accuracy (R² scores) & Dataset: 300 images (50 per class, 6 classes)
Training time: 6 hours
Hardware: Single Tesla T4 (12GB RAM) \\
        \hline
        
        Haribabu G. et al.\textsuperscript{\citep{Haribabu2023AData}} & 2023 & WGAN-GP loss & $512^2$, $256^2$, $128^2$ pixels & 2D & $512^2$, $256^2$, $128^2$ pixels & 2D & Grayscale & Ti-6Al-4V (titanium alloy) & First demonstration of StyleGAN2-ADA for generating synthetic titanium alloy microstructures using limited dataset (1225 images) & FID (Frechet Inception Distance)
KID (Kernel Inception Distance)
IS (Inception Score)
Phase fraction analysis
Grain type distribution
Two-point correlation analysis
Local boundary thickness measurements & Dataset: 1,225 images
Hardware: Not specified
Batch size: 64
Training epochs: 1000 \\
        \hline
        
        Cao D. et al.\textsuperscript{\citep{Cao2022ReconstructionGAN}} & 2022 & WGAN-GP + InfoGAN mutual information loss & $128^3$ voxels & 3D & $128^3$ voxels & 3D & Binary (2-phase) & Berea Sandstone and Tight Carbonate & Combined InfoGAN and StyleGAN architectures
Multi-scale information fusion through style transfer
Prior information (porosity) guided generation
Classifier Q ensures porosity constraint & Two-point cluster function
Multi-point connectivity function
Chord length distribution
Pore-throat analysis
Porosity accuracy
Permeability measurements & Dataset: 27 samples (26 training, 1 test)
Hardware: NVIDIA TESLA V100 (32GB)
Optimizer: Adam
Batch size: 2
Learning rate: 0.0002 \\
        \hline
        
        Huang Y. and Xiang Z. (1)\textsuperscript{\citep{Huang2022Slice-to-voxelNetworks}} & 2022 & WGAN-GP + L2 regularization & $64^3$ voxels & 2D slices (extracted from 3D volumes) & $64^3$ voxels (training), $128^3$ voxels (evaluation) & 3D & Binary (2-phase) & Berea Sandstone and Estaillades Carbonate & First application of StyleGAN to slice-to-voxel porous media reconstruction
Integration of 3D generator with 2D discriminator for computational efficiency
Achieves or exceeds voxel-to-voxel method quality with spatial continuity preservation & Two-point correlation function with radial averaging
Minkowski functionals (volume, surface area, mean breadth, Euler characteristic)
Visual comparison of reconstructed samples
Comparison with Mosser et al. method & Dataset: 10,647 samples (64³ voxels each)
Training time: 12 hours
Hardware: Single NVIDIA GeForce RTX 2080Ti
Batch size: 128
Progressive growing from 4³ \\
        \hline
        
        Huang Y. et al. (2)\textsuperscript{\citep{Huang2022Deep-learning-basedMethod}} & 2022 & WGAN-GP including regular term & $96^3$ pixels (Set A), $128^3$ pixels (Sets B and C) & 3D & $96^3$ pixels (Set A), $128^3$ pixels (Sets B and C) & 3D & Binary (2-phase) & Bead pack, Berea, Doddington, Bentheimer, Estillades, Ketton & First efficient volumetric version of StyleGAN for 3D microstructures
Quantitative control of morphology generation
Improved understanding of dual-channel control mechanisms & Minkowski functionals (volume, surface area, mean breadth, Euler characteristic)
Two-point correlation function & Dataset: 8,000-9,261 samples
Training time: <48 hours
Hardware: Two NVIDIA GeForce RTX 2080Ti GPUs
Optimizer: Adam
Batch sizes: 64/128 \\
        \hline
        
        Liu M. et al.\textsuperscript{\citep{Liu2022MultiscaleNetworks}} & 2022 & GAN adversarial loss + cycle-consistent loss & SEM: $1024^2$ pixels, Micro-CT: $64^2$ pixels & 2D for SEM, 3D for micro-CT & $1024^2$ pixels (stacked to 3D: 1792×1792×2688 voxels) & 2D and 3D & Binary (2-phase) & Carbonate rock & Combined StyleGAN2-ADA and CycleGAN for multiscale fusion
Handles limited high-resolution training data
Works with unpaired training samples & Porosity distribution
Specific perimeter
Two-point correlation function
Correlation length
Morphological characteristics
Pore structure evaluation
Flow simulation results & Dataset: 371 high-resolution SEM samples, 200,000 low-resolution micro-CT samples
Training time: ~54 hours total (30h StyleGAN2-ADA, 24h CycleGAN)
Hardware: One Nvidia A100 (40GB) \\
        \hline
        
        Fokina D. et al.\textsuperscript{\citep{Fokina2020MicrostructureNetworks}} & 2020 & WGAN-GP & $128^2$ pixels & 2D & $128^2$ (can be merged for larger samples) & 2D & Binary and grayscale & Alporas aluminum foam, Berea sandstone, Ketton limestone & First application of StyleGAN for microstructure synthesis
Image quilting technique for generating larger samples
Demonstrates effectiveness for various material types & Elastic properties (Young's modulus, Poisson's ratio)
Minkowski functionals (area density, perimeter density, Euler characteristic)
Two-point correlation function & Dataset: 16,000 128×128 samples
Hardware: Not specified
Batch size: 16 \\
        \hline
    \end{longtable}
\end{landscape}

\begin{landscape}

    \tiny
    \raggedright  
    \setlength{\LTleft}{0pt}
    \setlength{\LTright}{0pt}
    \begin{longtable}{|>{\raggedright\arraybackslash}p{1.4cm}|>{\raggedright\arraybackslash}p{0.4cm}|>{\raggedright\arraybackslash}p{1.2cm}|>{\raggedright\arraybackslash}p{1.2cm}|>{\raggedright\arraybackslash}p{0.5cm}|>{\raggedright\arraybackslash}p{1.2cm}|>{\raggedright\arraybackslash}p{0.5cm}|>{\raggedright\arraybackslash}p{1cm}|>{\raggedright\arraybackslash}p{1.8cm}|>{\raggedright\arraybackslash}p{3cm}|>{\raggedright\arraybackslash}p{3cm}|>{\raggedright\arraybackslash}p{3cm}|}
    
        \caption{Summary of Hybrid Architecture GAN for Digital Porous Material Reconstruction}\label{tab:hybrid_gan_overview}\\
        
        \hline
        \textbf{Reference} & \textbf{Year} & \textbf{Loss Function} & \multicolumn{2}{c|}{\textbf{Training Images}} & \multicolumn{2}{c|}{\textbf{Generated Images}} & \textbf{Image Type} & \textbf{Material} & \textbf{Novelty} & \textbf{Evaluation Metrics} & \textbf{Computational Requirements} \tabularnewline
        \hline
        & & & \textbf{Size} & \textbf{Dim} & \textbf{Size} & \textbf{Dim} & & & & & \tabularnewline
        \hline
        \endfirsthead
        
        \multicolumn{12}{c}{{\tablename\ \thetable{} -- continued from previous page}} \\
        \hline
        \textbf{Reference} & \textbf{Year} & \textbf{Loss Function} & \multicolumn{2}{c|}{\textbf{Training Images}} & \multicolumn{2}{c|}{\textbf{Generated Images}} & \textbf{Image Type} & \textbf{Material} & \textbf{Novelty} & \textbf{Evaluation Metrics} & \textbf{Computational Requirements} \tabularnewline
        \hline
        & & & \textbf{Size} & \textbf{Dim} & \textbf{Size} & \textbf{Dim} & & & & & \tabularnewline
        \hline
        \endhead
        
        \hline
        \multicolumn{12}{r}{{Continued on next page}} \\
        \endfoot
        
        \hline
        \endlastfoot
        
        Zhang T. \& Zhang W.\textsuperscript{\citep{Zhang20253DTransformer}} & 2025 & WGAN-GP + Reconstruction loss (L2 norm)& $64^3$ voxels (general pores), $128^3$ voxels (fractures) & 3D & $64^3$ voxels (general pores), $128^3$ voxels (fractures) & 3D & Binary & Shale (with significant fractures and general pores) & Novel STGAN combining StyleGAN and Transformer for 3D digital rock reconstruction. Multi-scale feature extraction using Transformer blocks (coarse, intermediate, fine scales). Porosity distribution sequence as conditional/prior information. Classifier Q to constrain porosity distribution during reconstruction & Porosity statistics
Pore network parameters (pore/throat numbers and diameters)
Absolute permeability (X, Y, Z directions)
Multiple-point connectivity (MPC) curves
Minkowski functionals (surface area, mean breadth, Euler characteristic)
MS-SWD-MDS for diversity evaluation & Dataset: 729 sub-images ($64^3$) for general pores; 301 sub-images ($128^3$) for fractures
Training Time: ~50,077s for $64^3$; ~85,340s for $128^3$ (10 reconstructions, 8000 epochs)
GPU: NVIDIA Tesla V100 (16GB)
CPU: Intel Xeon (2.30 GHz)
RAM: 16GB \\

\hline
Zhang Y. et al.\textsuperscript{\citep{Zhang2025ForADA-PGGAN}} & 2025 & WGAN-GP + ResNet50: LogCosh loss + VGG16: MSE + L1 ($\lambda=512$) & $1024^2$ pixels & 2D & $1024^2$ pixels (2D slices), $1090 \times 1024 \times 1024$ voxels (3D reconstruction through stacking) & 2D $\rightarrow$ 3D & Binary & Mt. Simon sandstone (heterogeneous) & RVION (ResNet-VGG Inversion Optimization Network) for inferring latent codes of real rock images. ADA-PGGAN with adaptive data augmentation mechanism to handle limited/scarce training data. Latent space interpolation (Slerp) for 3D reconstruction generating 10 interpolated slices between each pair of scanned slices. Cost efficiency: Up to 10$\times$ reduction in imaging costs & Sliced Wasserstein Distance (SWD)
Minkowski functionals (Euler values, porosity)
Two-point correlation function S$_2$(r)
Absolute permeability (x, y, z directions)
Microstructure parameters (Amount of throats/pores, Average shape factors, Average size of pore radius, Average volume of pore, Average coordination number)
Fluid flow simulation/velocity field comparison & Dataset: 500 training slices ($1024^2$) for ADA-PGGAN, 700 test slices (20\% validation), 80,000 generated images for ResNet50 training
Training Time: 98 hours
GPU: NVIDIA GeForce RTX 3090 \\
\hline

Liu F. et al.\textsuperscript{\citep{LIU2025PoreAuto-segmentation}} & 2025 & StyleGAN2-ADA: Adversarial + ADA mechanism; SRCycleGAN: Adversarial + Cycle-consistency+ Identity; SegGAN: Conditional adversarial + Identity & StyleGAN2-ADA: $512^2$ pixels; SRCycleGAN: LR $64^2$ pixels, HR $512^2$ pixels; SegGAN: $512^2$ pixels & 2D & StyleGAN2-ADA: $512^2$ pixels; SRCycleGAN: $512^2$ pixels (8$\times$ upscaling); SegGAN: $512^2$ pixels; 3D case: 5.12 $\mu$m$^3$ & 2D (extendable to 3D) & Grayscale (multi-phase segmentation: pores, organic matter, clay minerals, pyrite) & Shale & Three-stage GAN framework: image augmentation (StyleGAN2-ADA), 8× super-resolution without paired training (SRCycleGAN), and multi-mineral auto-segmentation (SegGAN with U-Net). Reconstructs fine-scale pores unresolved in LR images. Slice-by-slice 3D processing reduces memory requirements. Addresses resolution vs field of view trade-off & Two-point correlation function S$_2$(r)
Linear-path correlation function L(r)
Entropy (E)
Porosity
Organic Matter Proportion (OMP)
Pore/throat size distribution
Coordination number
Permeability
PSNR, SSIM, IoU
Gray histogram comparison
Mean shape factor & Dataset: StyleGAN2-ADA: 100 HR sub-images ($512^2$); SRCycleGAN: S1: 14,626 LR sub-images ($64^2$), S2: 7,904 LR sub-images ($64^2$); SegGAN: Paired images
Training Time: Not specified
GPU: Single NVIDIA V100
RAM: 32GB \\

\hline

Zhang Y. et al.\textsuperscript{\citep{Zhang2024DA-VEGAN:Sets}} & 2024 & Encoder loss with $\beta$-KLD + Generator loss + Discriminator loss & $64^2$ pixels & 2D & $64^2$ pixels & 2D & Binary & Alloy, Carbonate, Ceramics, Copolymer, Rubber, Sandstone, Composite, Synthetic (10 materials total) & Incorporation of $\beta$-VAE into hybrid GAN. Custom differentiable data augmentation scheme. Ability to train on extremely small datasets. Controllable smoothness of latent space & Two-point spatial correlation function
RMSE of correlation functions
Reconstruction error ($\lambda_{rec}$)
Generation error ($\lambda_{gen}$)
Volume fraction & Dataset: 16 samples
GPU: RTX 3080Ti
Training: ~30 min, 20,000 epochs
RAM: 64GB
CPU: AMD Ryzen 9 5900X
Batch size: 4 \\

        \hline
        Zhang T. et al.\textsuperscript{\citep{Zhang2024StochasticVAE-GAN}} & 2024 & WGAN-GP with code discriminator (two-part loss) & $64^3$ voxels & 3D & $64^3$ voxels & 3D & Binary & Shale and Berea sandstone & Added code discriminator to VAE-GAN to handle distribution error. Uses WGAN-GP loss function for better stability. Two adversarial learning stages in each epoch & Multi-point connectivity curves
Absolute permeability simulation
Pore space analysis (Total/Connected/Isolated porosity, Pore number, Average radius)
Difference degree for MPC curves
Fréchet inception distance & Dataset: 124 training samples (each material)
GPU: GeForce RTX-3080
Training: 4000-5000 epochs (~68,000s first training)
RAM: 16GB
CPU: AMD-R9 3900X \\
\hline
        Zhang T. et al.\textsuperscript{\citep{Zhang20243D-FGAN:Cores}} & 2024 & Hinge loss formulation with two reconstruction losses (L2 norm) & $64^3$ voxels & 3D & $64^3$ voxels & 3D & Binary & Shale & Added 4 Skip-Layer Excitation (SLE) blocks to generator for faster generation and feature calibration. Combined auto-encoders (two SimpleDecoders) with discriminator to prevent overfitting. Parameters saved for reuse (26,304s first vs 41s subsequent) & Statistical Properties (Total/Connected/Isolated pore space)
Pore Network Statistics (Pore/Throat number and diameter)
Distribution Analysis (diameter and coordination ratios)
MPC Analysis
Absolute Permeability
Diversity Analysis (MS-SWD-MDS) & Dataset: 216 samples - 215 train, 1 test
GPU: NVIDIA RTX 3090 (24GB VRAM)
Training: 20,000 epochs (7.3h first)
RAM: 32GB
CPU: Intel Xeon E5-2680 \\

        \hline
        Li Y. et al.\textsuperscript{\citep{Li20233DReconstruction}} & 2023 & BicycleGAN losses (implicit cVAE-GAN + cLR-GAN) + 3D co-occurrence matrix loss function (CCF3D) & $128^3$ voxels & 2D/3D (uses both) & Up to $1000^3$ demonstrated & 3D & Binary & Berea Sandstone & First unlimited-size reconstruction capability. Novel multimodal dictionary approach combining BicycleGAN with super-dimension. New 3D co-occurrence matrix loss function. Diversity preservation strategy & Statistical Functions (Two-point correlation function, Linear path function, Two-point cluster function, Autocorrelation function)
Distribution Analysis (Pore diameter distribution, Pore volume distribution)
Morphological Parameters
Statistical Properties & GPU: NVIDIA GeForce GTX 1080Ti (11GB memory)
Dataset: Not explicitly stated (80\%/20\% train/test split)
Training: 4 hours (GANDB-SD model)
GPU memory: 2.6 GB (for $128^3$ elements)
Batch size: 2 \\
\hline

Li J. et al.\textsuperscript{\citep{Li2023DeepImage}} & 2023 & VAE loss (pixel-wise + KL divergence) + WGAN-GP + porosity loss (L2) & $128^2$ pixels (inferred) & 2D & $128^3$ voxels & 3D & Binary (2-phase) & Bentheimer, Fontainebleau, and Berea sandstone, Ketton limestone & 3D reconstruction from single 2D slice using combined VAE encoder + WGAN-GP (named EWGAN-GP). VAE encoder extracts 2D statistical features. 3 discriminators (x,y,z directions). Porosity loss to constrain 3D structure. Wasserstein distance with gradient penalty for stable training & Statistical metrics (Two-point correlation function, Lineal-path function, Two-point cluster function, Local porosity distribution)
3D pore network properties (Pore/throat number, radius, volume, Shape factor, coordination number, Effective permeability) & Dataset: Single 2D slice per material for training
Training Time: Not specified
Hardware: Not specified \\

        \hline

        Yin R. et al.\textsuperscript{\citep{Yin2023Three-dimensionalModels}} & 2023 & VAE loss + GAN losses + pattern loss (MSE of pattern distributions) & $128^2$ pixels & 2D & $128^3$ voxels & 3D & Binary & Bead pack, Fontainebleau sandstone, and Belgian fieldstone & Hybrid VAE-GAN to generate diverse 3D microstructures from single 2D slice. Self-attention module in encoder and generator to capture long-range dependencies. Pattern loss for morphological features. Multiscale discriminator with spectral normalization. Demonstrated on samples with significantly varied grain shapes & Autocorrelation function (ACF)
Multiple-point connectivity (MPC)
Grain morphology statistics
Pore network parameters
Euler characteristic and Betti numbers
Tortuosity and absolute permeability & GPU: NVIDIA GeForce RTX 2080 Ti
CPU: i7-9700
Training time: Not specified \\
        \hline

        Wu Y. et al.\textsuperscript{\citep{Wu2023AnFeatures}} & 2023 & FastGAN: Combined hinge version adversarial + reconstruction loss; CycleGAN: Adversarial + cycle consistency loss & $128^2$ pixels & 2D & S1: $1024^2 \times 1204$, S2: $1696 \times 1200 \times 2400$, S3: $1520^2 \times 4000$ voxels & 3D & Grayscale & Shale samples and carbonate rock & First time using unpaired training images for 3D high-resolution reconstruction. Novel hybrid architecture combining FastGAN for data augmentation with CycleGAN. Ability to handle extremely large output sizes & Statistical Metrics (Two-point correlation function, Lineal-path function, Two-point cluster function, Porosity distributions, Fractal dimension, Multiple-point correlation function)
Pore Network Properties
Physical Properties (Connectivity, Tortuosity, Permeability) & Dataset: ~100 unpaired images used to generate 30,000 synthetic images
GPU: Nvidia RTX 3080ti (12GB)
FastGAN: ~5 hours
CycleGAN: ~12 hours \\

        \hline

Zhang F. et al.\textsuperscript{\citep{Zhang2023PM-ARNN:Network}} & 2023 & Reconstruction loss + KL-divergence loss + Porosity loss + Adversarial loss & $256^2$ pixels & 2D & $256^3$ voxels & 3D & Binary & Bentheimer sandstone (homogeneous), Estaillades carbonate rock (heterogeneous), and a fractured core & Integration of RNN and GAN architectures. Weak guidance and self-connection structure in generating module. Two-stage training strategy. Layer-by-layer generation with adversarial learning & Two-point correlation function S2(r)
Lineal-path function L(r)
Two-point cluster function C2(r)
Local porosity distribution (LPD)
Quantitative structural properties
Material-specific pore network metrics & Dataset: 500 train/100 test (Bentheimer/Estaillades), 376 train/76 test (fractured cores)
GPU: V100 (32G)
RAM: 32GB
CPU: i7-4790K
Batch size: 4 \\
        \hline
Cao D. et al.\textsuperscript{\citep{Cao2022ReconstructionGAN}} & 2022 & WGAN-GP + InfoGAN mutual information loss & $128^3$ voxels & 3D & $128^3$ voxels & 3D & Binary & Sandstone and carbonate rock & Addition of porosity distribution as prior information. Multi-scale information fusion through style transfer. Classifier Q to ensure reasonable porosity range. Combination of InfoGAN and StyleGAN architectures & Two-point cluster function
Multi-point connectivity from X,Y,Z directions
Chord length distribution
Pore and throat diameter distributions
Porosity distribution
Permeability simulation results & Dataset: 27 samples (26 training, 1 test)
GPU: NVIDIA TESLA V100
Batch size: 2 \\
        \hline
Li Y. et al.\textsuperscript{\citep{Li2022CascadedImage}} & 2022 & BicycleGAN loss + grayscale pattern loss + grayscale pixel value loss & $128^2$ pixel & 2D & Up to $1024^3$ voxels & 3D & Grayscale CT images & Berea sandstone (contains quartz and K-feldspar) & Deep gray padding technology for 2D to 3D conversion. Cascade network architecture. New loss functions based on grayscale distribution. Single image training approach. Progressive multi-scale reconstruction & Variogram Analysis
Local Porosity Distribution Function
Gray Level Co-occurrence Matrix Features (Energy, Entropy, Contrast, Inverse difference moment, Correlation)
Pore Network Model Comparison
Material Composition Accuracy
Permeability
Seepage Analysis & GPU: NVIDIA GTX 1080
RAM: 16GB
CPU: Intel i7-6700k
Batch size: 4 \\

        \hline
        Nguyen P.C. et al.\textsuperscript{\citep{Nguyen2022SynthesizingLearning}} & 2022 & GAN adversarial loss + Actor-Critic reinforcement learning reward function & $128^3$ voxels & 3D & $128^3$ voxels & 3D & Binary & Bentheim sandstone & First integration of GAN with Actor-Critic reinforcement learning for microstructure generation. Novel design of GAN-AC framework for controlled generation. Ability to generate structures with specific target properties & Statistical Distance Metrics (Kullback-Leibler divergence, Maximum Mean Discrepancy, Earth Mover's Distance)
Physics-based Metrics (Porosity, Specific surface area, Permeability, Average clustering, Graph density, Degree assortativity coefficient, Local efficiency, Graph transitivity) & Dataset: 1000 training episodes, 32 design iterations per episode
Batch size: 64
Training Time: Not specified
Hardware: Not specified \\

        \hline
Phan J. et al.\textsuperscript{\citep{Phan2022Size-invariantImage}} & 2022 & VQ-VAE: MSE loss + quantization loss; GAN: Adversarial loss & 2D slices (top face) from 3D volumes of varying sizes & 2D & Variable, demonstrated up to $1000^3$ voxels and larger (examples include $1500 \times 1500 \times 3000$) & 3D & Grayscale & Sandstones and Carbonates & Size-invariant multi-step 3D generation workflow. Combination of VQ-VAE, transformer, and GAN. Ability to generate large-scale outputs (>$1000^3$ voxels). Works with multiple rock types & Porosity distributions
Euler characteristic
Permeability measurements & Dataset: 9 porosity images (3 sandstones, 6 carbonates)
GPU: RTX3090 (for generation)
Generation Time: ~9 hours for $1024^3$ volume
Training Time: Not specified \\

        \hline
Volkhonskiy D. et al.\textsuperscript{\citep{Volkhonskiy2022GenerativeSlices}} & 2022 & GAN adversarial loss + autoencoder reconstruction loss (L2) & $128^3$ voxels & 3D & $128^3$ voxels & 3D & Binary & Berea sandstone, Ketton limestone, In-house south Russian sandstone & Generation of 3D porous media from a 2D slice. Combination of autoencoder and 3D convolutional GAN. Generator conditioned on both noise and encoded 2D slice. Allows generating 3D structure around the given 2D slice as the most probable based on learned distribution & Porosity
Permeability
Two-point correlation function
Minkowski functionals (Volume, surface area, mean breadth, Euler characteristic)
Kullback-Leibler divergence & GPU: Tesla V100
Training Time: Around 10 hours
Hardware: Not specified \\

        \hline

        Zhang T. et al.\textsuperscript{\citep{Zhang20223DNetworks}} & 2022 & Modified least-squares GAN + L1 reconstruction + porosity loss & $64^3$ voxels & 3D & $64^3$ standard; $128^3$ super-resolution & 3D & Binary & Sandstone and carbonate samples & Single image training. Scaling transformation for super-resolution. Multi-scale discriminator design. Stochastic super-resolution reconstruction & Porosity, connectivity factor
Permeability (X,Y,Z)
MPC curves
Pore/throat properties (numbers, diameters, coordination)
DD analysis & Dataset: 1 training image (64³ TI)
GPU: RTX 3080 (10GB)
Training: 6000 iterations (~5100s sandstone, ~9350s carbonate SR)
RAM: 16GB, CPU: AMD R9 3900X
Subsequent: ~1s, LR: 0.0002 \\
        \hline
        Zhang H. et al.\textsuperscript{\citep{Zhang2022FastLearning}} & 2022 & WGAN with spectral normalization + custom losses per network & $96^3$ voxels & 3D & $600^3$ voxels demonstrated & 3D & Binary & Ketton limestone (oolitic) & LGCNN: 4 specialized networks (L-WGAN, V-WGAN, S-WGAN, S-CNN). Splicing algorithm with 4 block types. REV-scale (3-4 orders larger). 10min vs 642-33,628min traditional. Batch diversity loss & Porosity, SSA
Average pore/throat radius
Two-point probability
Pore size distribution
Permeability (X,Y,Z)
Connectivity factor
Coordination numbers
Fractal dimension & Dataset: 2197 SPM, 13,182 buffers (10\% validation)
GPU: RTX 3090 (24GB)
Training: Not specified
CPU: Intel i7-8700K \\
\hline
Zheng Y. et al.\textsuperscript{\citep{Zheng2024AEvaluation}} & 2024 & Wasserstein distance loss + L1 image regularization & $256^3$ voxels & 3D & $256^3$ voxels & 3D & Binary/Multi-phase (3 phases for NMC) & Berea sandstone, Ketton rock, lithium battery NMC cathode, hypoeutectic white cast iron, copper-zinc alloy & HFA module (hierarchical fusion), MSCA module (channel attention), CBAM in generator layers. Refined loss function. Comprehensive metric-based evaluation system & SSIM, PSNR
S2(r), L(r), C2(r), CD(p)
Phase Volume Ratio
TPB (Triple Phase Boundary) Density
Relative Diffusion Coefficients & Dataset: Not specified (batch size: 16)
GPU: 4× RTX 3090 (24GB each)
CPU: Intel Xeon Platinum 8175M \\
\hline
        Feri L. et al.\textsuperscript{\citep{Feri2021APenalty}} & 2021 & 3D-Improved WGAN with enhanced gradient penalty & $20^3$ voxels & 3D & $20^3$ voxels & 3D & Binary & Permeable pavement & Implementation of 3D-IWGAN with enhanced GP. Integrated VAE for latent vector generation. Framework for 3D reconstruction from single 2D image. Improved stability vs regular GAN & Porosity values
Surface area
Hydraulic conductivity
Permeability
Euler characteristic & Dataset: 24,389 images (20³ voxels) from 9 samples
GPU: NVIDIA GeForce GTX 1060 (6GB)
RAM: 64GB
CPU: Intel Xeon E5-2609 v4 \\

        \hline
        Zhang T. et al.\textsuperscript{\citep{Zhang20213DAuto-encoders}} & 2021 & VAE loss + GAN loss + GDPP diversity loss & $64^3$ voxels & 3D & $64^3$ voxels & 3D & Three-phase & Shale & Combined VAE-GAN with GDPP (learning diverse generations using determinantal point processes). Addresses fuzzy outputs and model collapse & MPC curves
Porosity measurements
Pore space analysis
Absolute permeability
Facies proportions
Pore diameter distributions
Pore number analysis & Dataset: Single $64^3$ voxel training image
GPU: NVIDIA GeForce RTX 2080 (8GB VRAM)
CPU: Intel Core i5-6200U\\
 
        \hline

        Zhang F. et al.\textsuperscript{\citep{Zhang2021Slice-to-voxelModel}} & 2021 & Combined VAE loss + GAN loss + porosity loss & $128^2$ pixel & 2D & $128^3$ voxels & 3D & Binary & Bentheimer sandstone (isotropic), Stone particle penetrated porous medium (anisotropic) & Combined VAE-GAN architecture for 2D to 3D reconstruction. Encoder extracts features from 2D slice for generator input. Porosity loss function for improved accuracy. Stable training through hybrid architecture & Two-point correlation function
Lineal-path function
Two-point cluster function
Porosity measurements
Pore network analysis
Two-phase flow simulation & Dataset: Not specified (batch size: 4)
GPU: GTX 2080Ti
CPU: i7-9700
Training: 2h for $64^3$, 5h for $128^3$\\

        \hline

        Feng J. et al.\textsuperscript{\citep{Feng2020AnLearning}} & 2020 & BicycleGAN: cVAE-GAN + cLR-GAN losses & $128^2$ pixels & 2D & $128^3$ voxels & 3D & Binary & Sandstone, Silica, and non-stationary porous media & End-to-end 2D to 3D reconstruction framework. Combines BicycleGAN with custom architecture for porous media. Enables diverse 3D reconstructions from single 2D slice. Fast reconstruction after training & Two-point correlation function (S2)
Lineal-path function (L)
Two-point cluster function (C2)
Local Porosity Distribution (LPD)
Microstructure parameters & Dataset: 600-1000 samples per material
CPU: i7-4790K (4.00 GHz)
Batch size: 2
Noise channel: 8
GPU: Not specified
Training time: Not specified \\

        \hline

Chen H. et al.\textsuperscript{\citep{Chen2020Super-resolutionNetworks}} & 2020 & CycleGAN: LSGAN adversarial loss + Cycle consistency loss + Identity loss & Multiple scales: $256^2$ to $1900^2$ pixels & 2D & Multiple scales: $256^2$ to $1900^2$ pixels (upsampling: 1.8$\times$, 2.0$\times$, 4.0$\times$) & 2D & Grayscale & Sandstone (primary), other rock types & First CycleGAN application for real-world rock micro-CT super-resolution
Unpaired training without matched LR-HR pairs addressing practical data challenges
Bidirectional mapping with cycle consistency for structural preservation
Handles real-world CT artifacts and noise
Generates large FOV high-resolution images exceeding hardware limits
Maintains both visual quality and statistical properties & PSNR, SSIM
Porosity
Local porosity distribution
Two-point correlation function
Lineal-path function
Two-point cluster function
Chord-length distribution function
Pore size distribution & Dataset: Unpaired images at different resolutions; Sandstone Real 1.8$\times$: 900 LR ($1900^2$ pixels), 512 HR ($520^2$ pixels)
GPU: Not fully specified
Training time: Not explicitly stated \\

\hline

Shams R. et al.\textsuperscript{\citep{Shams2020CoupledMedia}} & 2020 & GAN: Adversarial loss; AE: MSE reconstruction loss (trained separately) & GAN: $64^3$ (Berea), $128^3$ (Sandy); AE: $256^2$ pixels & GAN: 3D; AE: 2D & $64^3$ to $128^3$ voxels & 3D & Binary & Berea sandstone, Multi-scale sandy microstructure, Carbonate samples & Coupled GAN-AE architecture for multi-scale reconstruction. AE adds intra-grain porosity details to GAN's inter-grain structure. Two-stage approach combining low and high resolution features. First application of coupled GAN-AE for multi-scale porous media & Porosity distribution
Auto-correlation function
Permeability analysis
Visual assessment of structural features & Dataset: 10,647 (Berea), 13,824 (Sandy microstructure), 32,768 (AE 2D samples)
GPU cluster (brand unspecified)
Training Time: Not specified
Batch size: Not mentioned \\

        \hline
    \end{longtable}
\end{landscape}

\bibliographystyle{unsrt}
\bibliography{references}